%% file: main_arxiv.tex
\documentclass{article}

\usepackage[table]{xcolor}

\usepackage[preprint]{neurips_2025}

\usepackage[utf8]{inputenc} %
\usepackage[T1]{fontenc}    %
\usepackage{hyperref}       %
\usepackage{url}            %
\usepackage{booktabs}       %
\usepackage{amsfonts}       %
\usepackage{nicefrac}       %
\usepackage{microtype}      %
\usepackage{graphicx} %
\usepackage{comment}   %
\usepackage{tcolorbox} %
\usepackage{amsmath} %
\usepackage{amsfonts}
\usepackage{xspace}
\usepackage{algorithm}
\usepackage{algorithmic}
\usepackage{caption}

\usepackage{hyperref}
\usepackage{pdflscape}
\usepackage{multirow}
\usepackage{marvosym}
\usepackage{tablefootnote}
\usepackage{url}

\usepackage{wrapfig}

\newcommand{\nextcaption}{}
\newcommand{\setNextCaption}[1]{\renewcommand{\nextcaption}{#1}}

\newcommand{\modelname}{COSMIC\xspace}
\newcommand{\benchall}{Benchmark~I\xspace}
\newcommand{\benchnotarget}{Benchmark~II\xspace}
\newcommand{\benchcaption}{\benchall includes all datasets, \benchnotarget which excludes datasets where other targets are used as covariates (Section~\ref{sec:result-cov-dataset})}

\makeatletter
\AfterEndEnvironment{algorithm}{\let\@algcomment\relax}
\AtEndEnvironment{algorithm}{\kern2pt\hrule\relax\vskip3pt\@algcomment}
\let\@algcomment\relax
\newcommand\algcomment[1]{\def\@algcomment{\footnotesize#1}}

\renewcommand\fs@ruled{\def\@fs@cfont{\bfseries}\let\@fs@capt\floatc@ruled
  \def\@fs@pre{\hrule height.8pt depth0pt \kern2pt}%
  \def\@fs@post{}%
  \def\@fs@mid{\kern2pt\hrule\kern2pt}%
  \let\@fs@iftopcapt\iftrue}
\makeatother

\title{Zero-Shot Time Series Forecasting with Covariates via In-Context Learning}

\author{%
    Andreas Auer\textsuperscript{$1$ \Cross}  \ \ \ \
    Raghul Parthipan\textsuperscript{$2$ *} \ \ \
    Pedro Mercado$^{\ 3}$ \ \ \
    Abdul Fatir Ansari$^{\ 3}$ \ \ \ \vspace{0.2mm}\\
    \textbf{Lorenzo Stella}$^{\ 3}$  \ \ \ \
    \textbf{Bernie Wang}$^{\ 3}$  \ \ \ \
    \textbf{Michael Bohlke-Schneider}$^{\ 3}$  \ \ \ \
    \textbf{Syama Sundar Rangapuram}$^{\ 3}$  \ \ \ \ \vspace{1mm}
\\
    {$^1$}{Johannes Kepler University Linz, Austria}  \vspace{0.1mm}\\
    {$^2$}{University of Cambridge, UK} \vspace{0.1mm}\\
    {$^3$}{AWS AI Labs}    
}

\begin{document}

\maketitle

\begin{abstract}
Pretrained time series models, capable of zero-shot forecasting, have demonstrated significant potential in enhancing both the performance and accessibility of time series forecasting.
However, existing pretrained models either do not support covariates or fail to incorporate them effectively.
We introduce \modelname, a zero-shot forecasting model that utilizes covariates via in-context learning.
To address the challenge of data scarcity, we propose Informative Covariate Augmentation, which enables the training of \modelname without requiring any datasets that include covariates.
\modelname achieves state-of-the-art performance in zero-shot forecasting, both with and without covariates.
Our quantitative and qualitative analysis demonstrates that \modelname effectively leverages covariates in zero-shot forecasting.
\end{abstract}

\section{Introduction}

Machine learning has witnessed a paradigm shift from task-specific models that are trained on a particular dataset for a specific task to pretrained models that are pretrained on a large training corpus and used for unseen datasets and tasks. ~\cite{touvron2023llama, dosovitskiy2021an}.
Recently, pretrained models have been introduced in the time series domain and have been shown to be competitive with established time series forecasting methods~\cite{wooUnifiedTrainingUniversal2024a, ansariChronosLearningLanguage2024b, dasDecoderonlyFoundationModel2024e}. Notably, pretrained models are competitive to task-specific models when used zero-shot, i.e. without any adaptation to the specific target dataset. Zero-shot forecasting makes the dissemination of these models much easier, as practitioners can leverage zero-shot models without specialized domain knowledge or machine learning expertise. Additionally, they might be able to apply these models even when there is insufficient data to train task-specific models. This accessibility improvement might drive adoption in application domains such as energy, retail, or healthcare. 

In certain cases, forecasting practitioners might have access to covariates, additional data that can be used to better predict the target variable.
This data can then be used to improve the forecast, for example to model infrequent events where the behavior of the target variable might be different (peak traffic around events or lack of sales when an item goes out of stock). 
However, existing pretrained models struggle to effectively handle covariates in zero-shot forecasting.
In particular, there are two key challenges:
(1) it remains unclear how a model can leverage the predictive value of covariates in a zero-shot setting, i.e.\ without any prior knowledge of the covariate semantics or their relationships with the target, and
(2) public datasets that include relevant covariates are scarce compared to the vast amounts of data required to train these models effectively.
Note that while covariate data might be scarce in public time series datasets, it might actually be abundant in proprietary applications.
However, it is still unclear how to effectively pretrain time series models with covariates and how to leverage them in a zero-shot setting. 

In this work, we address these limitations and effectively integrate covariates into a pretrained zero-shot forecasting model. The main contributions of this paper are:
\begin{enumerate}
    \item \modelname, a pretrained model that incorporates covariates in a zero-shot setting. In our experiments, \modelname outperforms other pretrained models on 9 out of 11 datasets that include covariates. Additionally, it achieves state-of-the-art results in zero-shot forecasting without covariates and demonstrates performance comparable to task-specific models. 
    \item Informative Covariate Augmentation, a method for training covariate-aware zero-shot models without requiring any training data with covariates.
    \item An analysis of how covariates can enhance performance in a zero-shot forecasting, addressing the limitations and challenges that arise in this setting.
\end{enumerate}

\paragraph{Related Work.}
Classical statistical models, such as ARIMA~\citep{boxRecentAdvancesForecasting1968} and exponential smoothing~\citep{hyndmanForecastingExponentialSmoothing2008}, have long been standard approaches for time series forecasting.
Over the last decade, neural network-based models have become well-performing alternatives in forecasting. These models employ LSTM architectures, transformer architectures, or hybrid architectures combined with distributional output heads or multi-quantile outputs~\cite{salinasDeepARProbabilisticForecasting2020a, limTemporalFusionTransformers2021, nieTimeSeriesWorth2022}. Other notable models include N-Beats \citep{oreshkinNBEATSNeuralBasis2019}, one of the first deep stacked architectures for time series forecasting, later extended with covariate usage \citep{olivaresNeuralBasisExpansion2023}, and N-HiTS \citep{challuNHITSNeuralHierarchical2023}, which adds a hierarchical multi-resolution approach. Simpler one-layer networks also demonstrated competitive performance \cite{zengAreTransformersEffective2023}. However, these models have in common that they require task-specific training on individual datasets. 

Inspired by pretrained models in the vision and language domains, pretrained time series models aim to provide universal models for forecasting. 
These models enable zero-shot forecasting, which removes the need for task-specific training. 
TimesFM \citep{dasDecoderonlyFoundationModel2024e} and  Moirai \citep{wooUnifiedTrainingUniversal2024a} differ in their patching mechanism and their employed transformer architecture. Chronos \citep{ansariChronosLearningLanguage2024b} maps time series to tokens from a fixed vocabulary via quantization and utilizes the T5 transformer architecture. TinyTimeMixer (TTM) \citep{ekambaramTinyTimeMixers2024b} proposes a lightweight pretrained model based on the TSMixer \cite{chenTSMixerAllMLPArchitecture2023a} architecture. Among these, only Moirai supports the input of covariates in zero-shot forecasting. 
Pre-Fitted Networks \citep{mullerTransformersCanBayesian2021,hollmannTabPFNTransformerThat2023} represent a complementary line of research that uses synthetic training data to achieve generalization on real-world tasks. 

Our work draws inspiration from earlier studies that use synthetic data augmentation for pretraining time series models~\cite{dasDecoderonlyFoundationModel2024e, ansariChronosLearningLanguage2024b} and introduce Informative Covariate Augmentation to pretrain models with covariates. We also show that this approach leads to effective use of covariates in zero-shot forecasting. 

The paper structure is as follows: 
Section~\ref{sec:setting} introduces the problem setting and analyzes how covariates can be leveraged via in-context learning.
Section~\ref{sec:ci-chronos} presents \modelname and Section~\ref{sec:augmentation} covers Informative Covariate Augmentation.
In Section~\ref{sec:limitation} we discuss limitations.
Section~\ref{sec:experiments} evaluates \modelname on real-world data and examines the impact of covariates.
Section~\ref{sec:conclusion} concludes the paper.

\section{Zero-Shot Forecasting with Covariates}\label{sec:setting}
Zero-shot forecasting is the task of predicting future values of a target time series without task-specific training~\citep{ansariChronosLearningLanguage2024b}.
Unlike task-specific forecasting methods, which rely on historical and related observations of the target series for fitting a model, zero-shot forecasting exploits generalizable patterns learned from other tasks or domains to make predictions on previously unseen datasets.
Formally, let $\{y_t\}_{t=1}^T$ represent the target time series: the goal is to forecast its future values  $\{y_{T+1}, \ldots, y_{T+h}\}$ for a forecast horizon $h$.
Throughout the paper, we follow a numpy-like array notation and denote a consecutive set of time points, e.g. the target time series, as $\mathbf{y}_{1:T} := \{y_t\}_{t=1}^T$. 
Forecasting inherently involves uncertainty, making probabilistic forecasts more informative than single-point predictions.
Rather than estimating only specific future values, we model the distribution of possible outcomes expressed as $\mathcal{P}(\mathbf{y}_{T+1:T+h} | \mathbf{y}_{1:T})$.

\begin{figure*}[ht]
    \centering
    \includegraphics[width=\textwidth]{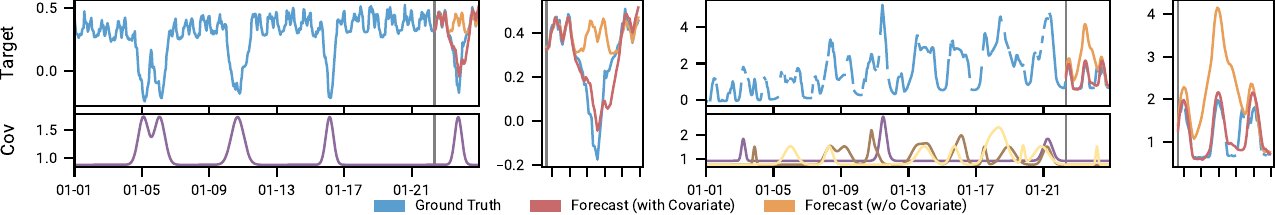}
    \caption{Two synthetic examples demonstrating how \modelname leverages covariate information. Left: The covariate impact, a negative correlation, is inferred from the context and applied in the forecast horizon. Right: The covariate model accounts for covariate impacts in the context to predict the intrinsic target signal. Each example includes three plots: the full target signal (top-left), the covariate signals (bottom-left), and a zoomed-in view of the forecast horizon (right).}
    \label{fig:main-synth-example}
\end{figure*}

In zero-shot forecasting with covariates, auxiliary information is provided as input to the model:
we denote the set of $k$ covariates as $\mathbf{X} = \{\mathbf{x}^1, \ldots \mathbf{x}^k \}$.
These covariates may represent exogenous variables such as weather conditions, economic indicators, or event schedules.
Covariates that are time series can be further categorized into past-only covariates (in $\mathbb{R}^T$), which are observed up to the current time step $T$, and past and future covariates (in $\mathbb{R}^{T+h}$), which are available or can be forecasted for the forecast horizon.
Hence, in covariates-informed zero-shot forecasting, where no sample of the dataset $D$ is known at training time, the goal is to estimate the conditional probability distribution
\begin{equation}
    \mathcal{P}(\mathbf{y}_{T+1:T+h} | \mathbf{y}_{1:T}, \mathbf{X}), \quad \quad (\mathbf{y}, \mathbf{X}) \in D.
\end{equation}

\paragraph{Covariate usage via In-Context Learning.}
Task-specific forecasting models learn \emph{at training time} how covariates from the task relate to the target.
This is not possible in a zero-shot scenario, as the model does not have access to the task-specific data during training.
However, a zero-shot model might infer this relationship by observing the relationship in the context of the given time series.
Thus, the zero-shot model requires the capability of \emph{in-context learning} the covariate-target relationship during inference.
While the in-context information is limited compared to training on global data, and understanding the covariate-target relationship during inference seems challenging, two observations support its potential:
(1) Results in the language modeling domain suggest that transformers are capable of sophisticated in-context learning \citep{brownLanguageModelsAre2020b}, and
(2) the application of local covariate models like ARIMAX indicates that harvesting this local information can be beneficial.
Building on these observations, we developed \modelname.

Figure~\ref{fig:main-synth-example} illustrates two stylized examples where local information can improve the forecasting performance and previews how \modelname utilizes this potential:
In the left example, a single covariate exhibits unpredictable bumps inversely correlated with the target signal.
While the model predicting without covariates fails to anticipate these bumps in the forecast horizon, the model with covariates successfully infers the relationship and adapts its predictions.
In the right example, where multiple covariates influence the target, the covariate model disentangles their impact from the ``intrinsic target signal'' and successfully forecast its pattern.
Even without direct covariate impact in the forecast horizon, the model can better understand the target context as it can ``explain away'' the covariate impact from it.
This capability is particularly valuable when future covariate information is limited.

\section{\modelname}\label{sec:ci-chronos}

\begin{figure*}[ht]
    \centering
\includegraphics[width=\textwidth]{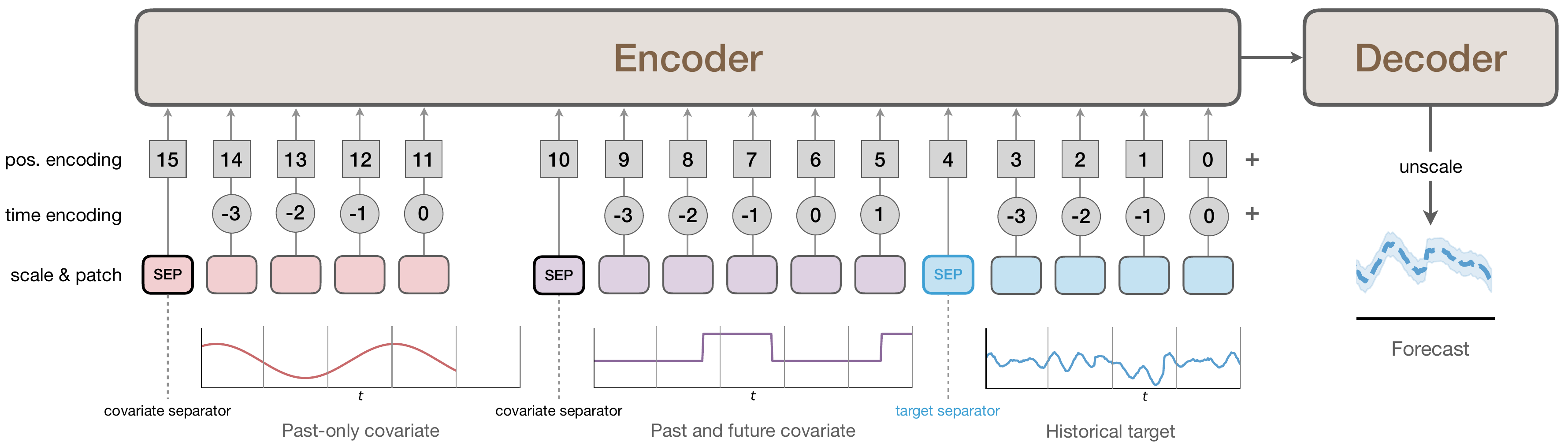}
    \caption{Architecture of \modelname. The illustration shows an example with 4 context tokens and one forecast token, including one past and future covariate (purple), and one past-only covariate (red).}
    \label{fig:main-model-figure}
\end{figure*}

\modelname utilizes a encoder-decoder transformer architecture \citep{vaswani_attention_2017}.
The input layer processes the time series by applying scaling and patching, followed by time and variate encoding to generate input tokens.
The transformer processes these tokens.
The output tokens represent the forecasted patches of the target, and are mapped back to the forecast window.
The individual components are detailed below --- Figure~\ref{fig:main-model-figure} illustrates the architecture.

\paragraph{Scaling \& Patching.}
\modelname should be able to operate with time series across different domains that may exhibit drastically varying scales.
To account for that, \modelname normalizes the target time series and covariates individually, often referred to as instance normalization \citep{kimReversibleInstanceNormalization2021}.
Specifically, we apply $z$-score normalization, where we formally compute, $\mathbf{\tilde{x}^i} = \frac{\mathbf{x}^i - \bar{x}^i}{\sigma_{x^i}} \; \forall \mathbf{x}^i \in \mathbf{X}$ and $\mathbf{\tilde{y}_{0:T}} =  \frac{\mathbf{y}_{0:T} - \bar{y}_{0:T}}{\sigma_{y_{0:T}}}$.
Here, $\bar{x}$ and $\bar{y}$ represent the mean, and $\sigma_{x}$ and $\sigma_{y}$ represent the standard deviation of the specific series.
\modelname maps non-overlapping windows of time steps to tokens via a two-layer residual block~\cite{heDeepResidualLearning2015, srivastavaTrainingVeryDeep2015}.
This is referred to as ``patching'' in previous literature \citep{nieTimeSeriesWorth2022, wooUnifiedTrainingUniversal2024a} and is motivated by a related technique in vision transformers \citep{dosovitskiyImageWorth16x162020}.
To handle missing values \modelname concatenates a binary padding mask, indicating whether a time step is missing, to the residual block input.
Given a window size of $m_{in}$ and a hidden dimension of $d$ for the transformer, the residual block represents a function~$\mathbb{R}^{2m_{in}} \rightarrow  \mathbb{R}^{d}$.
Patching reduces the effective context length of the transformer model by a factor of $m_{in}$.
Hence, the memory and compute complexity are reduced quadratically by this factor.
This enables adding multiple covariate signals directly to the model's context without loss of efficiency compared to models operating on individual observations, such as Chronos.
All variates and time steps are patched with the same residual block weights.

\paragraph{Time and Variate Encoding.} 
\modelname includes patches of multiple variates in the transformer input, hence, the position of a patch in the resulting sequence does not reflect its temporal position within the corresponding variate.
Therefore, the patch embedding is combined with a learned additive embedding vector based on the time position of the specific patch.
To allow the encoder to distinguish between different variates we add a separation token before the start of a new variate.
We use two different separation tokens, $\mathbf{s}_c \in \mathbb{R}^d$ which is used before a new covariate starts, and $\mathbf{s}_t \in \mathbb{R}^d$ which is used before the target variate starts.
To allow the permutation-invariant encoder to make use of these separation tokens we add rotary embeddings \citep{suRoFormerEnhancedTransformer2023} to encode the sequence position. 

\paragraph{Output Layer \& Loss.}
Similar to the input, the decoder's output tokens are first mapped to the forecast window via a residual block and then rescaled to the original target space.
Note that the length of the output windows does not need to be identical to the length of the input windows.
Instead of single-point predictions, the model outputs $\mathbf{Y} \in \mathbb{R}^{h \times |Q|}$, representing $|Q|$ quantile values for each time step of the forecast horizon, enabling probabilistic forecasts.
Specifically, the model outputs forecasts for nine equidistant quantile levels, $Q=\{0.1, 0.2, \dots, 0.9\}$.
The model is trained by minimizing the quantile loss. 
That is, given the true value $y_t$ at time step $t$, and the corresponding quantile forecast $\hat{y}_t^q$ for the quantile level $q$, the loss function is defined as
\begin{equation}
    L = \frac{1}{h*|Q|} \sum_{t=T+1}^{T+h} \sum_{q \in Q}
    \begin{cases}
    q ( y_t - \hat{y}_t^q) & \text{if } \hat{y}_t^q \leq y_t \\
    (1 - q) (\hat{y}_t^q - y_t) & \text{else }.
    \end{cases}
\end{equation}

\section{Informative Covariate Augmentation}\label{sec:augmentation}

To enable \modelname to acquire the necessary in-context learning capability, we identified two critical prerequisites for the training data:
(1) The dataset must be sufficiently large and diverse to support pretraining the model with robust generalization across different domains.
(2) Training samples must include covariates that provide meaningful predictive information for forecasting the target, even when only local context is available.
In other words, the mutual information of the target in the forecast horizon $\mathbf{y}_{T+1:T+h}$ and the covariates $\mathbf{X}$ must be greater than zero.
Otherwise, there is no incentive --- in terms of training loss reduction --- to acquire this in-context learning capability.

Unlike in language and vision domains, public real-world time series datasets are relatively scarce, especially those with covariates.
Even if covaraites are available, their predictive value is still unclear, especially when only the local context is available  (e.g. static covariates do not provide predictive information given the local context).
Another practical problem is that the extensive usage of the available covariate datasets for training limits the evaluation of the zero-shot performance.
While selecting random time series and treating them as covariates could potentially address data scarcity, these signal would lack any predictive value by default.
Given these challenges, we avoid using any covariate-included datasets for training but propose Informative Covariate Augmentation to enhance an existing pre-training dataset without covariates.

Algorithm~\ref{alg:augmentation-main} outlines the procedure of Informative Covariate Augmentation.
First, for a given training target, we sample up to $k$ covariate --- either from the training corpus or synthetic covariate signals.
The number $k$ is sampled from a geometric distribution $G$ and upper bounded by $k_{\mathrm {max}}$.
Then for each covariate, an impact function $f$, that models how the given covariate influences the target signal, is sampled.
A maximum lag $l$ describing the maximum distance between the covariate observation and the impact on the target is selected a priori.
The impact given the function $f$ is computed and added to the original target, resulting in an augmented target.
The augmented target and the sampled covariates represent the augmented training sample, where the covariates now have a predictive value.
\vspace{-0.15cm}
\begin{algorithm}
\caption{Informative Covariate Augmentation}\label{alg:augmentation-main}
\textbf{Input:} Training corpus $\mathcal{C}$, Synthetic covariate generator $\mathcal{G}$, Impact function space $\mathcal{F}$ \\
\textbf{Output:} Augmented training sample %
\begin{algorithmic}[1]
\STATE Sample a time series $\mathbf{y} \in \mathbb{R}^{T+h}$ from $\mathcal{C}$
\STATE $k = \text{min}(\kappa,k_{\mathrm {max}})$ with  $\kappa \sim \mathrm{Geom}(p)$
\STATE Sample $k$ covariates $\{\mathbf{x}^1, \mathbf{x}^2, \ldots, \mathbf{x}^k\}$ from $\mathcal{C} \cup \mathcal{G}$
\STATE Sample $k$ impact functions $\{f^1, f^2, \ldots, f^k\}$ from $ \mathcal{F}$
\STATE \textbf{Augmented target:}\par
$y^{\text{aug}}_t = y_t + \sum_{i=1}^k f_t^i(\mathbf{y}, \mathbf{x}^i) \quad \forall t \in \{1 \dots T+h\}$
\STATE \textbf{Return training sample:} $(\mathbf{y}^{\text{aug}},\{\mathbf{x}^1, \mathbf{x}^2, \ldots, \mathbf{x}^k\})$
\end{algorithmic}
\end{algorithm}
\vspace{-0.25cm}

\paragraph{Synthetic Covariate Sigznals.}
We expand the sampling space of covariates beyond the training corpus by incorporating synthetically generated signals.
These synthetic signals are designed to represent typical covariate patterns that may arise in real-world data but are underrepresented in the training corpus.
Specifically, we generate signals exhibiting step or bell-shaped events and introduce trends with changepoints.
The parameters describing these events and trends are sampled from specific distributions, detailed in Appendix~\ref{sec:app-augment-details}.

\paragraph{Impact Function.} 
The sample space of the impact function $f$ is designed to address multiple key requirements for modeling covariate influences:
First, we bias the impact function to reflect meaningful patterns -- for instance, favoring impact of more recent values, i.e. with smaller temporal lag.
Second, we restrict the space to simple functions, recognizing the limitations of inferring complex relationships from the limited local context.
Additionally, we assume that the relationship between the covariates and the target is stable throughout the context and the forecast horizon.
Despite the focus on simplicity, the function space should still be capable of approximating real-world interactions.
Third, the function must accommodate scenarios where covariates do not impact the target.
Concretely, the chosen sample space $\mathcal{F}$ includes simple piecewise linear functions $f: \mathbb{R}^{T + h} \times \mathbb{R}^{T + h + l} \rightarrow \mathbb{R}^{T + h}$ defined as
\[
f_t(\mathbf{x}, \mathbf{y}) = \begin{cases}
          a_0 + \langle \mathbf{a}, \mathbf{x}_{t-l:t} \rangle + \varepsilon_t & t \in S(\mathbf{x},\mathbf{y}) \\
        0 & \text{else}
    \end{cases}
\]

where $S(\mathbf{x},\mathbf{y})$ is a set of active time steps with non-zero impact, $\varepsilon_t$ represents noise sampled from a Gaussian distribution, and $a_{0}, \mathbf{a}$ are linear coefficients sampled from a zero-inflated Gaussian distribution to enforce sparsity. In particular, the linear coefficient is more likely to be non-zero for recent lags compared to the distant ones.

The set of active time steps, $S(\mathbf{x}, \mathbf{y})$, is defined so that the value of the target or covariate at a time step falls in a certain quantile. We follow this simple procedure to determine $S(\mathbf{x}, \mathbf{y})$: (i) choose either the target or the covariate $\mathbf{z} \sim \{\mathbf{x}, \mathbf{y}\}$, (ii) sample a quantile level $q$ and determine the corresponding empirical quantile value $z_q$ from the given sequence, (iii) sample the inequality type $\diamond$: larger or smaller. Then $S(\mathbf{x},\mathbf{y}) := \{t \mid z_t \diamond z_q\}$.
Our augmentation process accounts for both, past-only and past and future covariates.
More details of the sampling procedure are presented in Appendix~\ref{sec:app-augment-details}.

\begin{figure*}[ht]
    \centering
        \includegraphics[width=\textwidth]{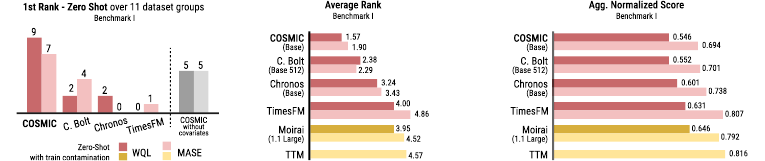}
    \caption{Comparison of pretrained model on all \underline{covariate-included evaluation data} (\benchall).
    Left: Count of first-place rankings within dataset groups, considering only zero-shot models.
    Moirai and TTM never rank first for datasets where they zero-shot forecast.
    \modelname's results without covariates access presented in grey. 
    Middle/Right: Average rank and aggregated score of the MASE and WQL metrics.
    Scores of the individual dataset groups are normalized by naive seasonal scores before aggregation.
    \modelname achieves the best performance.
    Note that Moirai and TTM training data overlap with part of the evaluation data, meaning they cannot be considered zero-shot for these datasets (overlap: Moirai 55\%, TTM 18\%).}
    \label{fig:main-zeroshot-comparison}
\end{figure*}

\begin{figure*}[ht]
    \centering
    \includegraphics[width=\textwidth]{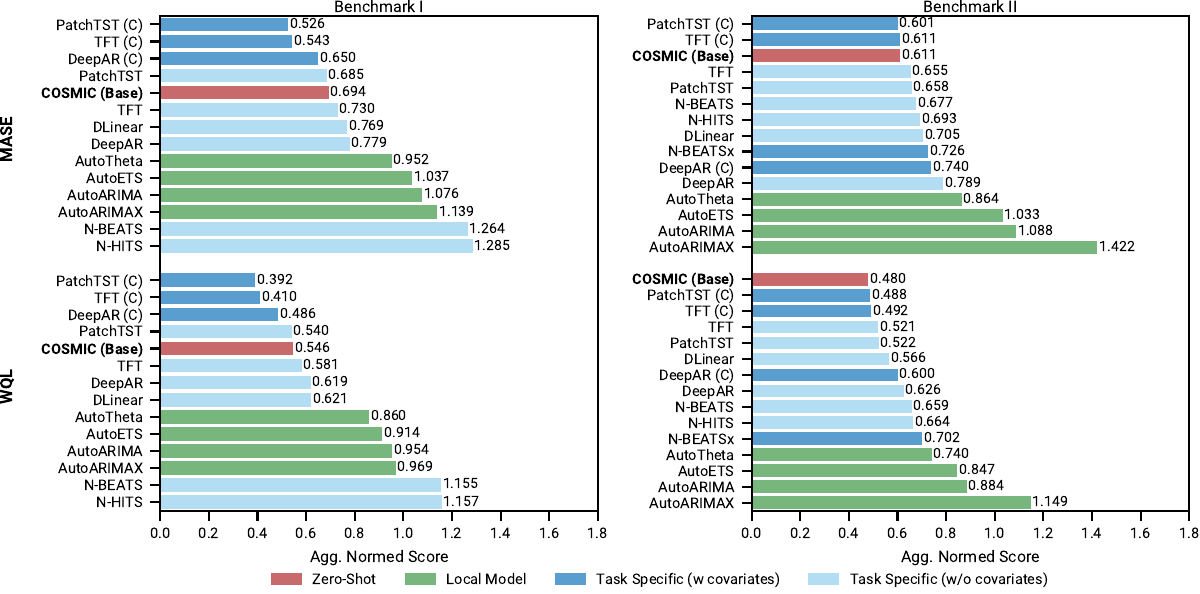}
    \caption{Aggreagted MASE and WQL performances on the \underline{covariate-included evaluation data} of different task-specific and local models compared to \modelname.
    Lower values are better.
    \benchcaption.
    Individual dataset scores are normalized by naive seasonal scores before aggregation.}
    \label{fig:main-taskspecifc-comparison}
\end{figure*}

\section{Experiments}\label{sec:experiments}

We evaluate \modelname in two setups:
covariate-included dataset evaluation in Section~\ref{sec:result-cov-dataset} and no-covariate dataset evaluation in Section~\ref{sec:results-uni-dataset}.
The corresponding procedures are outlined below, with additional details in Appendix~\ref{sec:exp-detail}.

\paragraph{\modelname Training.}
The training data for \modelname is derived from the Chronos corpus, which includes the TSMixup and Kernel Synth datasets \citep{ansariChronosLearningLanguage2024b}.
We enhanced the training data with Informative Covariate Augmentation as described in Section~\ref{sec:augmentation}.
We trained three model sizes — mini (20M), small (40M), and base (200M) — which align with the equivalently named Chronos sizes.
A context length of 512 timesteps is used which ensures the models are directly comparable to the corresponding sizes of Chronos, Chronos Bolt, and TimesFM.
For patching, we use a window size of 32 for the input tokens and 64 for the output tokens.
The training hyperparameters were kept consistent across all model sizes and are detailed in Appendix~\ref{sec:app-hyperparam}.

\paragraph{Metrics.}
We evaluate model performance with two metrics:
mean absolute scaled error (MASE) for point-forecast performance and weighted quantile loss (WQL) for probabilistic forecast performance.
WQL is calculated across nine uniformly spaced quantiles $(0.1, 0.2, \dots, 0.9)$.
For models that provide probabilistic predictions, the 0.5 quantile (median) is used to compute MASE.
For each dataset group, metrics are averaged over time series and forecast horizon to provide a group-level score.
For overall aggregation, each dataset group score is normalized by the score of a naive seasonal baseline, and the geometric mean of these normalized scores is computed.
This allows aggregation across groups with differing error magnitudes.
Overall this follows the evaluation protocol established by \citet{ansariChronosLearningLanguage2024b}.

\paragraph{Compared Models.}
We compare our models against a variety of state-of-the-art methods, including zero-shot, task-specific, and local forecasting approaches.
For zero-shot models, we benchmark the Chronos models \citep{ansariChronosLearningLanguage2024b}, TimesFM, \citep{dasDecoderonlyFoundationModel2024e}, Moirai (all variantes) \citep{wooUnifiedTrainingUniversal2024a}, and Tiny Time Mixer (TTM) \citep{ekambaramTinyTimeMixers2024b}.
We also compare to a new version of Chronos, Chronos Bolt~\citep{ansariabdulfatirFastAccurateZeroshot2024}.
Additionally, we evaluate several task-specific, state-of-the-art models, which are trained individually on each dataset, including PatchTST \citep{nieTimeSeriesWorth2022}, TFT \citep{limTemporalFusionTransformers2021}, DeepAR \citep{salinasDeepARProbabilisticForecasting2020a}, DLinear \citep{zengAreTransformersEffective2023}, N-BEATS \citep{oreshkinNBEATSNeuralBasis2019}, and N-HiTS \citep{challuNHITSNeuralHierarchical2023}.
Although these models are optimized for individual datasets, making a direct comparison to zero-shot approaches not entirely equivalent, such comparisons help highlight the strengths and limitations of current zero-shot models.
When applicable, we also use covariate-supporting variants of these models.
Specifically, we use the implementation of GluonTS \citep{alexandrovGluonTSProbabilisticTime2019, alexandrovGluonTSProbabilisticNeural2020} and NeuralForecast \citep{garza2022statsforecast}.
Finally, we benchmark classical local models, including ARIMA, ETS, and Theta, using the StatsForecast \citep{garza2022statsforecast} implementation with automatic parameter tuning.

\subsection{Covariate-included Data}\label{sec:result-cov-dataset}

In the covariate-included evaluation, we assess the models on 11 dataset groups spanning diverse domains (details in Appendix~\ref{sec:app-datasets}). %
Some of these datasets have a single target variate together with other covariates whereas others have multiple target variables.
In the case of multiple target variables, we individually evaluate each target while treating the remaining targets as covariates. 
The structure of the two types of datasets differs, as target-target correlations are different from covariate-target correlations. 
Therefore, we report the aggregated metrics across two benchmarks: \textbf{\benchall}, containing all datasets, and \textbf{\benchnotarget}, which excludes datasets where other targets are used as covariates.
To ensure robust results, we conduct rolling evaluation over the last 10\% of each series and assess two forecast horizons per dataset --- 1 and 2 periods --- where the period length varies by dataset.
The first 90\% of each series is kept aside to train the task-specific models.
While we treat all covariates as past and future covariates in the main evaluation, an alternative evaluation using covariates as past-only covariates is presented in Appendix~\ref{sec:app-past-cov}.
The raw WQL and MASE scores for individual dataset groups are provided in the Appendix in Tables~\ref{tab:individual-results-taskspecific-mase}-\ref{tab:individual-results-taskspecific-wql}.

\paragraph{Copmarision with Pretrained Models.}
Figure~\ref{fig:main-zeroshot-comparison} presents results comparing \modelname with other zero-shot forecasting models.
\modelname ranks first on 9 of 11 dataset groups in terms of WQL and 7 of 11 in terms of MASE.
The next best-performing model, Chronos-Bolt, ranks first in only 2 of 11 groups for WQL and 4 of 11 for MASE, highlighting \modelname’s strong overall performance.
The aggregate metrics present a similar picture, showing \modelname performs best in average rank and aggregated score for both MASE and WQL on \benchall, which includes all available covariate-included datasets.
The results for all other model sizes and \benchnotarget are presented in \ref{sec:app-extended-results}.
Notably, even the smallest \modelname variant (Mini, 20M parameters) outperforms, Chronos Large (700M parameter), Moirai Large (311M parameter), and TimesFM (200M parameter) in terms of average rank an aggregated score.
Overall \benchnotarget shows similar results to \benchall.
Since Moirai is not fully zero-shot due to training data overlap with evaluation datasets, Appendix~\ref{sec:app-morai-comp} provides a comparison excluding these datasets.

\paragraph{Comparison with Task-Specific and Local Models.}
Figure~\ref{fig:main-taskspecifc-comparison} compares the performance of \modelname in a zero-shot setting, to task-specific models and local models that are specifically trained on the individual evaluation datasets.
The potential advantage of task-specific models is even more pronounced when covariates are available, as task-specific models that utilize covariates can learn the global covariate-target relationship while \modelname can only rely on the context of the given sample.
Despite this limitation, \modelname demonstrates competitive results --- even when they use covariates.
Consistent with prior research \citep{ansariChronosLearningLanguage2024b}, PatchTST and TFT are the best performing task-specific models.
Notably, on \benchnotarget \modelname outperforms all models in terms of WQL.
The average ranking across datasets follows a similar ranking (see Appendix~\ref{sec:app-extended-results}).

\begin{figure*}[t]
    \includegraphics[width=\textwidth]{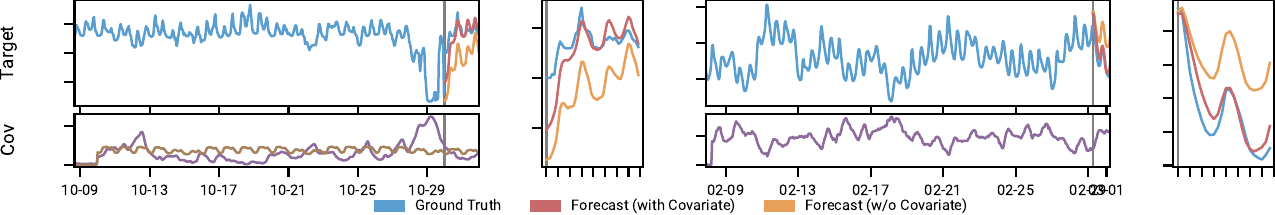}
    \caption{Two COSMIC forecasts with and without the access to covariates (Left: Electricity DE, Right: ProEnfo-GEF12).
    Each example has three plots: the full target signal (top-left), the covariate signals (bottom-left), and a zoomed-in view of the forecast horizon (right).}
    \label{fig:main-qualitative}
\end{figure*}

\begin{figure}
\begin{minipage}{0.4\textwidth}
    \centering
        \centering
        \includegraphics[width=\textwidth]{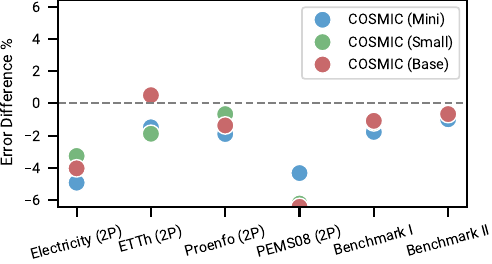}
        \caption{Relative MASE difference for four dataset specific and the aggregate results when covariates are provided to \modelname.}
        \label{fig:main-covariate-advantage}
\end{minipage}
\hfill
\begin{minipage}{0.588\textwidth}
\centering
    \input{tables/augmentation_ablation.tex}
\end{minipage}
\end{figure}
\paragraph{Impact of Covariates \& Augmentation.}
To analyze how \modelname utilizes the covariates, we evaluated it with and without access to the covariate signals.
Figure~\ref{fig:main-zeroshot-comparison} shows that the number of first-rank results increases from 5 to 9 in terms of WQL and 5 to 7 in terms of MASE when covariates are used.
Figure~\ref{fig:main-covariate-advantage} displays the error reduction for selected datasets when covariates are provided to \modelname  --- the complete results are shown Appendix~\ref{sec:app-extended-results}.
We find a pronounced improvement for some datasets; however, the aggregate results only show an improvement of about 1\% due to relatively minor or no improvements on other datasets.
This aligns with our hypothesis that not all datasets provide locally informative covariates.
We conducted the same analysis with Moirai, the only other pretrained model that supports covariates in zero-shot, and observed mixed results (see Appendix~\ref{sec:app-morai-comp}).
This suggests that Informative Covariate Augmentation is superior compared to training with a limited amount real-world covariate-inclued datasets, as done for Moirai.
Notably, we also find that for task-specific models, the performance change shows a considerable variance across datasets, where on some datasets access to covariates even worsens the performance. 

Alongside the quantitative analysis, we qualitatively examined \modelname's use of covariates.
Figure~\ref{fig:main-qualitative} shows two examples where the model advantageously utilizes the covariates for a better forecast.
Both scenarios discussed in Section~\ref{sec:setting} are also present in the real-world data forecasts.
The examples provide further evidence that \modelname in-context learning capabilities generalize to real-world scenarios.

Table~\ref{tab:augmentation-ablation} shows an ablation study examining the importance of the augmentation process.
We trained \modelname without applying Informative Covariate Augmentation --- i.e. we just sampled covariates but did not compute and apply the impact function --- and evaluated the performance of this model both with and without providing covariates.
The results demonstrate that the augmentation enables effective covariate utilization.
As anticipated, the performance gap between augmented and non-augmented models is largest on datasets where we observe the most improvements by utilizing covariates.
The augmentation also slightly improves the performance for some datasets when no covariates are provided to the model.
We hypothesize the augmentation alters the training data to better match the distribution of covariate-included datasets, where targets might be more likely to exhibit sudden jumps.

\subsection{No-Covariate Data}\label{sec:results-uni-dataset}
For the no-covariate evaluation, we use the same benchmark datasets and settings as Chronos \citep{ansariChronosLearningLanguage2024b}.
This includes 27 zero-shot datasets and 15 in-domain datasets.
In the in-domain evaluation datasets, non-overlapping parts of the individual datasets are part of the Chronos training corpus.
Appendix \ref{sec:exp-detail} provides details about the benchmark.

\begin{wrapfigure}{r}{0.6\textwidth}
    \centering
    \vspace{-3.5mm}
    \includegraphics[width=7cm]{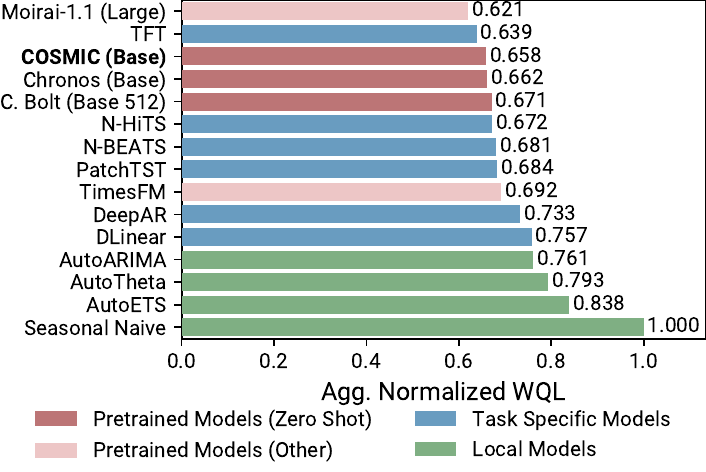}
    \caption{Aggregated WQL scores on the \underline{no-covariate zero-shot} \underline{evaluation benchmark}.
    Lower values are better.
    The individual scores of the dataset groups are normalized by naive seasonal scores before aggregation.
    ``Pretrained Models (Others)'' refers to models where datasets overlap with their training data --- hence, they are not zero-shot (overlap: Moirai 81\%, TimesFM 11\%).}
    \label{fig:agg-univariate-zeroshot}
    \vspace{-1mm}
\end{wrapfigure}

\paragraph{Results.}
Figure \ref{fig:agg-univariate-zeroshot} presents the WQL results of the no-covariate zero-shot benchmark.
\modelname demonstrates comparable performance to Chronos and, in many cases, achieves improved outcomes.
Hence, training with Informative Covariate Augmentation does not harm the performance on no-covariate datasets.
Overall, in zero-shot mode, \modelname achieves performance that is comparable to state-of-the art task-specific models which are explicitly trained on the individual datasets.
The MASE metrics results follow a similar overall ranking, with COSMIC as the best full zero-shot model.
These results are presented, including all model sizes, the in-domain benchmark, individual dataset metrics, and aggregated ranks, in Appendix~\ref{sec:app-extended-results}.

\section{Limitations}\label{sec:limitation}

\modelname represents a significant advancement in leveraging covariates in a zero-shot forecasting setting, yet certain limitations remain.
One constraint is the inability to capture covariate information that is only observable in a global perspective, where one can look at multiple samples from a task/dataset.
Consequently, static covariates, cannot be handled in the current framework.
While Informative Covariate Augmentation enables training models with demonstrated in-context learning capabilities for covariate-target relationships, the use of the simple impact function likely limits the ability to capture highly complex relationships.
However, this intentional choice reflects an inherent constraint of the zero-shot approach:
The limited information in the local context unlikely allows the inference of more intricate relationships in a generalizable manner.
Specialized models could address this by incorporating domain-specific inductive biases in the augmentation process.

\section{Conclusion}\label{sec:conclusion}

In this work, we address the challenge of leveraging covariates in a zero-shot forecasting setting and demonstrate that in-context learning can be an effective approach.
To this end, we propose \modelname, a transformer-based pretrained forecasting model that handles a variable number of covariates and effectively utilizes them to enhance its forecasts.
Additionally, we introduce Informative Covariate Augmentation, to tackle the challenge of data scarcity in the domain of public covariate-included time series data. 
Informative Covariate Augmentation allows us to train \modelname exclusively on data without covariates, while still acquiring the ability to perform covariate-aware zero-shot forecasting.

\modelname achieves the best results in zero-shot forecasting with covariates, outperforming state-of-the-art methods by ranking first on 9 out of 11 datasets for probabilistic forecasting.
Moreover, it delivers competitive performance compared to task-specific models that are specifically trained on the individual datasets, despite operating in zero-shot.
Our experiments show that \modelname leverages covariates in zero-shot forecasting more consistently than existing zero-shot approaches.
Importantly, the capability to handle covariates does not come at the expense of no-covariate forecasting performance --- the model also achieves state-of-the-art results in scenarios without covariates.

\newpage

\bibliography{main_arxiv}
\bibliographystyle{icml2025} %

\newpage
\appendix

\input{appendix_tmp}

\clearpage

\end{document}

%% file: tables/augmentation_ablation.tex
\captionof{table}{Augmentation Ablation: Normalized MASE scores for different datasets (aggregations) for \modelname (Small) and an identical version which was trained without informed covariate augmentation.
For both models the performance with and without providing the covariates is given.}
\label{tab:augmentation-ablation}
\begin{small}
\begin{tabular}{llrrrrr}
\toprule
\rotatebox{90}{\scriptsize Covariates} &
\rotatebox{90}{\scriptsize Augment}  & 
\rotatebox{62}{\scriptsize Electricity(2P)} & 
\rotatebox{62}{\scriptsize ETTh(2P)} & 
\rotatebox{62}{\scriptsize Proenfo(2P)} & 
\rotatebox{62}{\scriptsize PEMS08(2P)} & 
\rotatebox{62}{\scriptsize \benchall} \\
\midrule
no & no & 0.695 & 0.753 & 0.737 & 0.873 & 0.710 \\
\cline{1-7}
yes & no & 0.694 & 0.754 & 0.737 & 0.874 & 0.709 \\
\cline{1-7}
no & yes & 0.683 & 0.743 & 0.732 & 0.822 & 0.708 \\
\cline{1-7}
yes & yes & 0.660 & 0.729 & 0.727 & 0.771 & 0.699 \\
\bottomrule
\end{tabular}
\end{small}

%% file: appendix_tmp.tex
\section{Informative Covariate Augmentation - Procedure Details}\label{sec:app-augment-details}

This section provides more detailed information about Informative Covariate Augmentation that is proposed in Section~\ref{sec:augmentation}.

\paragraph{Impact Function}
First, let's repeat the chosen sample space of the impact function $\mathcal{F}$ $f: \mathbb{R}^{T + h} \times \mathbb{R}^{T + h + l} \rightarrow \mathbb{R}^{T + h}$ defined as
$$ f(\mathbf{x}, \mathbf{y}) = (f_1(\mathbf{x}, \mathbf{y}), ..., f_{T+h}(\mathbf{x}, \mathbf{y}))$$
where the $f_t$ are
\[
f_t(\mathbf{x}, \mathbf{y}) = \begin{cases}
          a_0 + \langle \mathbf{a}, \mathbf{x}_{t-l:t} \rangle + \varepsilon_t & t \in S(\mathbf{x},\mathbf{y}) \\
        0 & \text{else}
    \end{cases}
\]

Note that the sampled parameters (e.g. linear coefficients) are just sampled once per impact function, i.e. the are equal for all time steps.
Algorithm~\ref{alg:augmentation-impact} presents the procedure to sample the parameters for the impact function.
Essentially the procedure needs to sample the linear coefficients $a_i$ and the parameters describing the domain selection of the pice-wise function.
After sampling if an impact exists for a particular covariate (line~2), the linear coefficients are sampled in a 2-step process.
First, the number of active lags is sampled from a geometric distribution (line~3).
Afterward, we sample the lag positions from a geometric distribution (line~4).
The choice of the geometric distribution biases the function towards simplicity, as we favor impact with less active lags, and towards recency, as more recent observations have more likely an impact.
This is motivated by general time series modeling principles.
The values of the active coefficients are sampled from a Gaussian distribution (line~5).
If an impact via pice-wise function is sampled (line~7), the parameters necessary for the domain selection are sampled:
We either choose the covariate or the target as the relevant variable for the condition (line~9).
Then we sample the condition (line~10) and the quantile that defines the condition threshold (line~11).
Hyperparameters and notation details are shown in Table~\ref{tab:hyperparameter-augmentation}.
The noise of the impact $e_t$ is sampled from a Gaussian distribution where the variance is defined by the variance of the impact itself, scaled with by the factor $s_\epsilon$.

\begin{algorithm}
\caption{Sample Impact Function}\label{alg:augmentation-impact}
\textbf{Input:} hyperparameter see Table\ref{tab:hyperparameter-augmentation} \\
\textbf{Output:} FO coefficients $\{a_i\}_0^l$, parameter ($\diamond$, $z$, $q$) for $S$)
\begin{algorithmic}[1]
\STATE set all $a_i$ to 0
\IF{$X \sim U(0,1) > p_{\mathrm{FO}}$}
    \STATE Sample lag count: $c_{\mathrm{lag}} \sim \mathrm{Geom}(p_{\mathrm{lagcount}})$
    \STATE Sample lags: $L = \{\lambda_1, \dots \lambda_{c_\mathrm{lag}}\}$ with $\lambda_i \sim \mathrm{Geom}(p_{\mathrm{lagpos}})$
    \STATE Sample coefficients: $A = \{\alpha_1, \dots \alpha_{c_\mathrm{lag}}\}$ with $\alpha_i \sim N(0,1)$
    \STATE Set $a_i$ depending on $L$ and $A$. L defines which coefficents and $A$ the corresponding values.
    \IF{$X \sim U(0,1) > p_{\mathrm{PW}}$}
    \STATE Sample bias $a_0 \sim N(0,1)$
    \STATE Sample selected variable $z \sim U(\{y, x\})$
    \STATE Sample relalation $\diamond \sim U(\{>, <\})$
    \STATE Sample quantile: $q \sim U(0, 1)$
    \STATE \textbf{Return} $\{a_i\}_0^l$, (, $\oplus, z, q$)
    \ENDIF
\STATE \textbf{Return} $\{a_i\}_0^l$, ($>$, $y$, $0$)
\ENDIF
\end{algorithmic}
\end{algorithm}

\newpage

\paragraph{Synthetic Covariates}
Informative Covariate Augmentation samples covariates from the Chronos training corpus and from a synthetic covariate generator.
This generator generates signals with step or bell-shaped events combined with trends and changepoints.
Algorithm~\ref{alg:synth-covaraite} describes how the generator samples new covariates.
Hyperparameters and notation details are shown in Table~\ref{tab:hyperparameter-augmentation}.

\begin{algorithm}
\caption{Synthetic Covariates Generation}\label{alg:synth-covaraite}
\textbf{Input:} covariate length $T$, hyperparameter see Table\ref{tab:hyperparameter-augmentation} \\
\textbf{Output:} covariate $\mathbf{x\in \mathbb{R}^T}$
\begin{algorithmic}[1]
\STATE Sample event count: $c_{e} \sim U(1, c_e^{\mathrm{max}})$
\STATE Sample positions: $P^e = \{p_0, \dots p_{c_{e}}\}$ with $p_i \sim U(0, T)$
\STATE Sample type: type $\sim$ U(\{step, gauss\})
\STATE Sample event parameters and calculate series  $\mathbf{x}_e \in \mathbb{R}^T$ (\Cross)
\STATE Sample change-point count: $c_{\mathrm{cp}} \sim U(0, c_{\mathrm{cp}}^{\mathrm{max}})$
\STATE Sample change-point positions: $\{\pi_0, \dots \pi_{c_{\mathrm{cp}}}\}$ with $\pi_i \sim U(0, T)$
\STATE Sample change-point amplitudes: $\{a_0, \dots a_{c_{\mathrm{cp}}+2}\}$ with $a_i \sim N(0, \sigma_{\mathrm{cp}})$
\STATE Calculate $\mathbf{x}_{trend} \in \mathbb{R}^T$: Order and Interpolate $(\{0,a_0\}, \{\pi_0, a_1\}, \dots, \{\pi_{c_{\mathrm{cp}}},a_{c_e+1}\}, \{T,a_{c_{\mathrm{cp}}+2}\})$
\STATE \textbf{Return} $\mathbf{x}_e + \mathbf{x}_{trend}$
\end{algorithmic}
\algcomment{\Cross\xspace For type gauss: Amplitude $\alpha_i$ and $\sigma_i$ is sampled --- event $i$ is calculated with $\alpha_i\; G(p_i, \sigma_i)$. For type step: Amplitude is sampled --- step alternates between each event position. Events are summed with $\mathbf{0} \in \mathbb{R}^T$}
\end{algorithm}

\input{tables/hyperparam_augmentation}

\clearpage
\newpage

\section{Experiment Details}\label{sec:exp-detail}

This section provides detailed information about the experiment setup of Section~\ref{sec:experiments}.
Section~\ref{sec:app-datasets} provides information about the datasets used for training and evaluation, Section~\ref{sec:app-hyperparam} presents the hyperparameter used in the experiments, and Section~\ref{sec:app-merics} presents details of the evaluation metrics.

\subsection{Datasets}\label{sec:app-datasets}

\paragraph{\modelname Training Data}

For the training of \modelname, we utilized the same data corpus as \citet{ansariChronosLearningLanguage2024b} for the training of Chronos.
Equivalently to Chronos, we applied ``TsMixup'' and expanded the data pool with ``KernelSynth'' \citep{ansariChronosLearningLanguage2024b}.
The datasets included in the training corpus are:
Brazilian cities temperature, Mexico City Bikes, Solar (5 Min.), Solar (Hourly), Spanish Energy and Weathe, Taxi (Hourly), USHCN, Weatherbench (Hourly), Weatherbench (Daily), Weatherbench (Weekly), Wiki Daily (100k), Wind Farms (Hourly), Wind Farms (Daily), --- see Table~3 in \citet{ansariChronosLearningLanguage2024b}.
The data can be accessed on Hugginface: \url{https://huggingface.co/datasets/autogluon/chronos_datasets}

Unlike \citet{ansariChronosLearningLanguage2024b}, we additionally applied Informative Covariate Augmentation (see Section~\ref{sec:augmentation}).

\paragraph{Covariate-inclusive Evaluation Data}
Table~\ref{tab:dataset-cov} presents the datasets and evaluation parameters for the covariate-included data evaluation.
Datasets that contain multiple target variables, are individually evaluated for each target while the remaining targets are treated as covariates.
Each evaluation is done twice, where the forecast horizon is 1- and 2-period lengths respectively.
We apply a rolling evaluation over utilizing 10\% of each series for a sufficiently long time series, while guaranteeing at least one evaluation with the full horizon per series.
The remainder of the series are reserved for training task-specific models.

\setNextCaption{Overview of all datasets used for the covariate-included data evaluation.
For datasets with multiple targets all targets are evaluated separately.
None of these datasets is part of the Chronos training corpus.
The number of rolls in the evaluation differs between the 1- and 2-period forecast horizon setting..}
\input{tables/dataset_info_cov}

\newpage

\paragraph{No-Covariate Evaluation Data}

For the no-covariate evaluation, we adhere to the benchmark protocol established by \citet{ansariChronosLearningLanguage2024b}.
Since our training data aligns with theirs, the benchmark's division into in-domain and zero-shot settings remains applicable.
Table~\ref{tab:dataset-uni} provides an overview of the datasets used in the benchmarks.
For additional details, please refer to \citet{ansariChronosLearningLanguage2024b}.

\setNextCaption{Overview of all datasets used for the no-covariate data evaluation.
In-Domain data is part of the Chronos training corpus and used to train other task-specific baselines, except for the forecast horizon that is held out for in-domain evaluation.
Zero-Shot data is not part of the Chronos training corpus, but task-specific models are still trained on the data except for the forecast horizon that is held out for the (zero-shot) evaluation.
}
\input{tables/dataset_info_uni}

\clearpage

\subsection{Hyperparameter \& Model sources}\label{sec:app-hyperparam}

\subsubsection{\modelname}
\modelname is built on the encoder-decoder architecture of efficient T5 \citep{tayScaleEfficientlyInsights2021}, adopting the same hyperparameter schema (e.g., layers and dimensions) as T5's mini, small, and base configurations. 
However, we removed the original positional embedding, replacing it with our custom time and variate encodings.
Additionally, due to the use of patching, embedding parameters are not required.
We set the input patch window size to 32 and the output patch window size to 64.
The hidden dimension of the patching network is set to feed-forward dimension of the transformer.
The model is trained using the AdamW optimizer \citep{loshchilovDecoupledWeightDecay2018} with a learning rate of 0.001, a batch size of 256, and a cosine annealing learning rate scheduler.
We employ a 5\% linear warmup phase, a weight decay of 0.01, and train the model for 1 million steps.
The training hyperparameters are kept consistent across all model sizes.

\subsubsection{Zero-Shot Models}

For the compared zero-shot models we used the pre-trained models provided from the following sources:
\begin{enumerate}
    \item Chronos \url{https://huggingface.co/collections/amazon/chronos-models-and-datasets-65f1791d630a8d57cb718444}\\
    Number of Parameters: Mini - 20M, Small - 46M, Base - 200M, Large - 710M
    \item Chronos Bolt: Unpublished models (with 512 and 2048 context length) were provided from the team working on Chrons-Bolt.\\
    Number of Parameters: Small - 48M, Base - 205M,
    \item Moirai: \url{https://huggingface.co/collections/Salesforce/moirai-r-models-65c8d3a94c51428c300e0742}\\
    Number of Parameters: Small 14M, Base 91M, Large 311M
    \item TimesFM: \url{https://huggingface.co/google/timesfm-1.0-200m}\\
    Parameter: 200M
    \item Tiny Time Mixer (TTM): \href{https://github.com/ibm-granite/granite-tsfm/tree/main}{https://github.com/ibm-granite/granite-tsfm/tree/main}. We use TTM in zero-shot mode.\\
    Number of Parameters: 1M
\end{enumerate}

\subsubsection{Task Specific and Local Models}
Table~\ref{tab:hyperparam-baselines} presents the hyperparameters and the implementation sources of the task-specific and local models used in the experiments.
GluonTS refers to \citet{alexandrovGluonTSProbabilisticTime2019, alexandrovGluonTSProbabilisticNeural2020}, NeuralForecast and StatsForecast to \citet{garza2022statsforecast}.

\input{tables/hyperparam_baselines}

\newpage
\subsection{Evaluation Metrics}\label{sec:app-merics}
We employ two primary metrics to evaluate the forecasting performance:
Mean Absolute Scaled Error (MASE) and Weighted Quantile Loss (WQL).
The MASE measures the point forecasting performance.
Given a $H$-step forecast, and a context of length $T$ it is defined as:  
\begin{align*}
   \text{MASE} = \frac{\sum_{t=T+1}^{T+H} |y_t - \hat{y}_t|}{\sum_{t=1}^{T-S} |y_t - y_{t+S}|},
\end{align*}
where $y_t$ is the real value at time $t$, $\hat{y}_t$ is the forecasted value, and $S$ is a seasonality parameter.
The denominator represents the mean absolute error of a naive seasonal baseline model.
This normalization ensures that the MASE is scale-independent.
For models providing only probabilistic forecasts, the $0.5$ quantile (median) prediction is used.

The WQL measures the probabilistic forecasting performance by aggregating the quantile loss over multiple quantiles.
Given a $H$-step forecast, given a context of length $T$ is defined as: 
\begin{align*}
    \mathrm{QL}(q, \hat{y}^q_t, y_t) = & \begin{cases}
    q ( y_t - \hat{y}_t^q) & \text{if } \hat{y}_t^q \leq y_t \\
    (1 - q) (\hat{y}_t^q - y_t) & \text{else}
    \end{cases}, \\
    \text{WQL} =& \frac{1}{|Q| H} \sum_{i \in Q} \sum_{t=T}^{T+H} \frac{\mathrm{QL}(q, \hat{y}^q_t, y_t)}{|y_t|},
\end{align*}
where $y_t$ is the real value at time $t$, $\hat{y}_t^q$ is the forecasted $q$ quantile, and $Q = \{0.1, 0.2, \dots, 0.9\}$ is the set of quantile levels we evaluated.

\paragraph{Metric Aggregation}
To compute a metric for one dataset group and forecast horizion we compute the mean of the metrics over all time series of one dataset group for this forecast horizion.
Aggregating across different datasets for overall aggregations needs more consideration as the metric magnitude of the different datasets may vary.
We follow the protocol established in \citep{ansariChronosLearningLanguage2024b}:
We compute a relative score of each dataset as
the model’s score divided by the score of a Seasonal Naive model, which serves as a baseline.
The relative scores are aggregated across all datasets using the geometric mean.
Additionally, we calculate the average rank across dataset groups for a set of models.

\subsection{Compute and Hardware}\label{sec:app-compute}

We conducted all experiments on A10 GPUs.
For training COSMIC Tiny/Small, a single A10 GPU was sufficient; for COSMIC Base, two A10 GPUs were used.
Specifically, we utilized AWS G5 instances.
Inference requirements are flexible due to an adaptive batch size.

\clearpage
\newpage

\section{Extended Results}\label{sec:app-extended-results}

This section provides additional results from the experiments presented in Section~\ref{sec:experiments}.
The structure within the section is kept the same with first presenting results for the covariate-included data and afterward for no-covariate data.

\subsection{Covariate-included Data}

\begin{figure*}[ht]
    \centering
        \includegraphics[width=\textwidth]{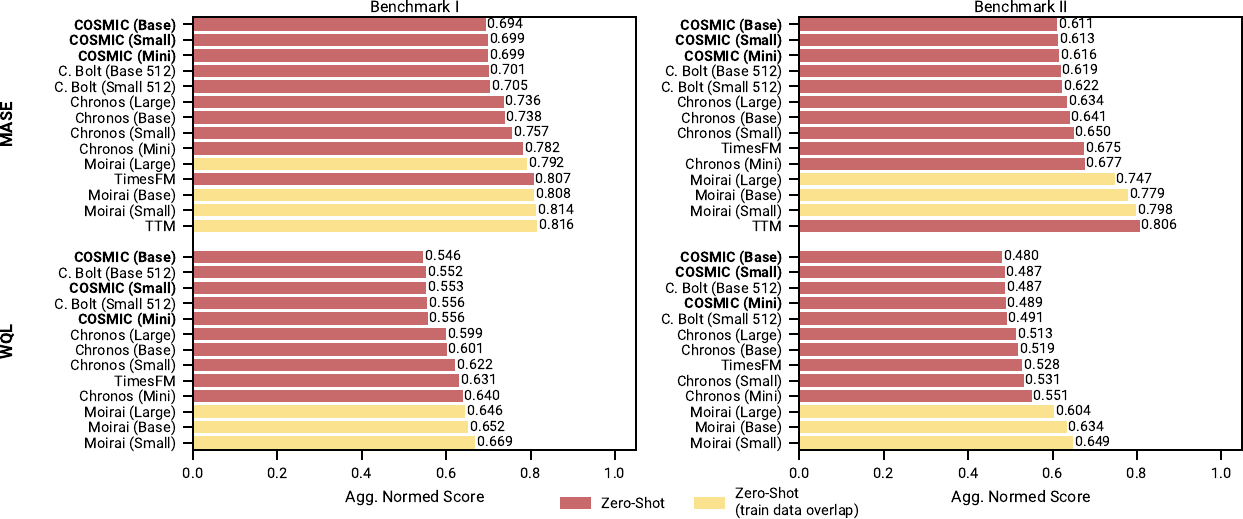}
    \caption{Aggreagted  MASE and WQL performances of different zero-shot models on the \underline{covariate-included evaluation data}.
    Lower values are better.
    \benchcaption.
    Scores of the individual dataset groups are normalized by naive seasonal scores before aggregation.
    Note that Moirai and TTM training data overlaps with part of the evaluation data. Hence the can not be considered zero-shot anymore for these datasets (overlap ratio: Moirai 55\%, TTM 18\%).}
    \label{fig:main-zeroshot-comparison-appendix}
\end{figure*}

\begin{figure*}[h]
    \centering
\includegraphics[width=\textwidth]{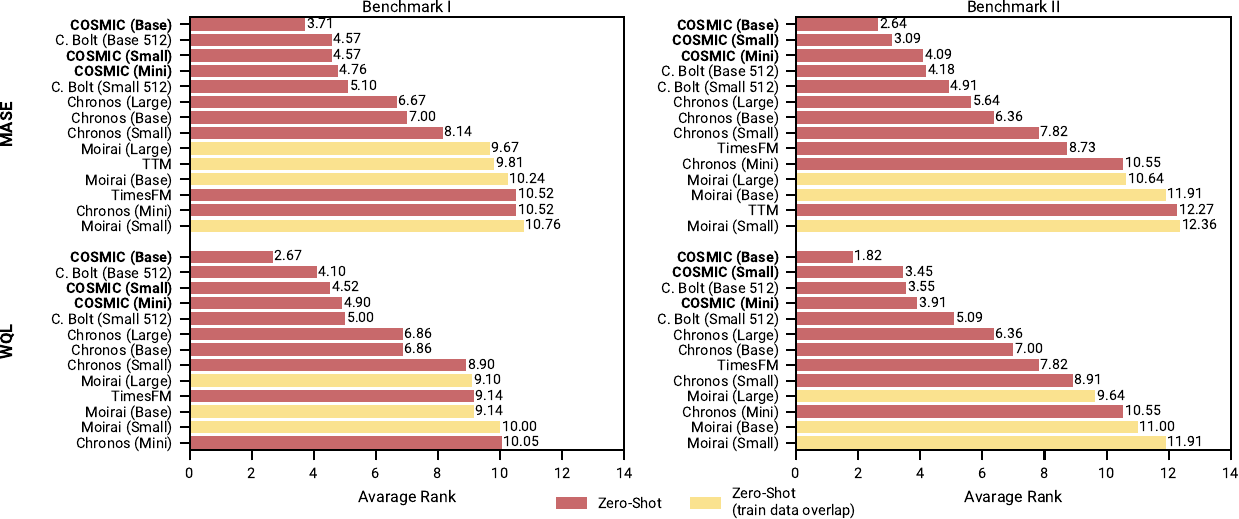}
    \caption{Average rank regarding MASE and WQL scores on the \underline{covariate-included evaluation data}, comparing the different zero-shot models. 
    \benchcaption.
    Note that Moirai and TTM training data overlaps with part of the evaluation data. Hence the can not be considered zero-shot anymore for these datasets (overlap ratio: Moirai 55\%, TTM 18\%).}
    \label{fig:zero-shot-rank}
\end{figure*}

\paragraph{Comparison wit Pretrained Models}
Figure~\ref{fig:main-zeroshot-comparison-appendix} presents the aggregated scores for all evaluated zeros-hot models in all sizes, extending Figure~\ref{fig:main-zeroshot-comparison} of the main paper.
Notably, \modelname does not only perform best when comparing the same model sizes.
Even the smallest \modelname variant (Mini, 20M parameters) outperforms, Chronos Large (700M parameter), Chronos Bolt Base (205M parameter), Moirai Large (311M parameter), and TimesFM (200M parameter) in terms of MASE, and all but Chronos Bolt Base in terms of WQL.
Similar Figure~\ref{fig:zero-shot-rank} presents the average rank over the different dataset groups when comparing different zero-shot models.
The order of the models is very similar to the aggregate metric comparison.
Although some datasets are within the training data of Moirai and TTM its rank average is computed without any adaptations.
Appendix~\ref{sec:app-morai-comp} provides a comparison that specifically considers only datasets that allow a clear-cut zero-shot evaluation of Moirai.

The metric results for the individual datasets and forecast horizons are presented in Table~\ref{tab:individual-results-zs-mase} (MASE) and Table~\ref{tab:individual-results-zs-wql} (WQL).

\paragraph{Comparison with Task-Specific and Local Models}

\begin{figure*}[h]
    \centering
    \includegraphics[width=\textwidth]{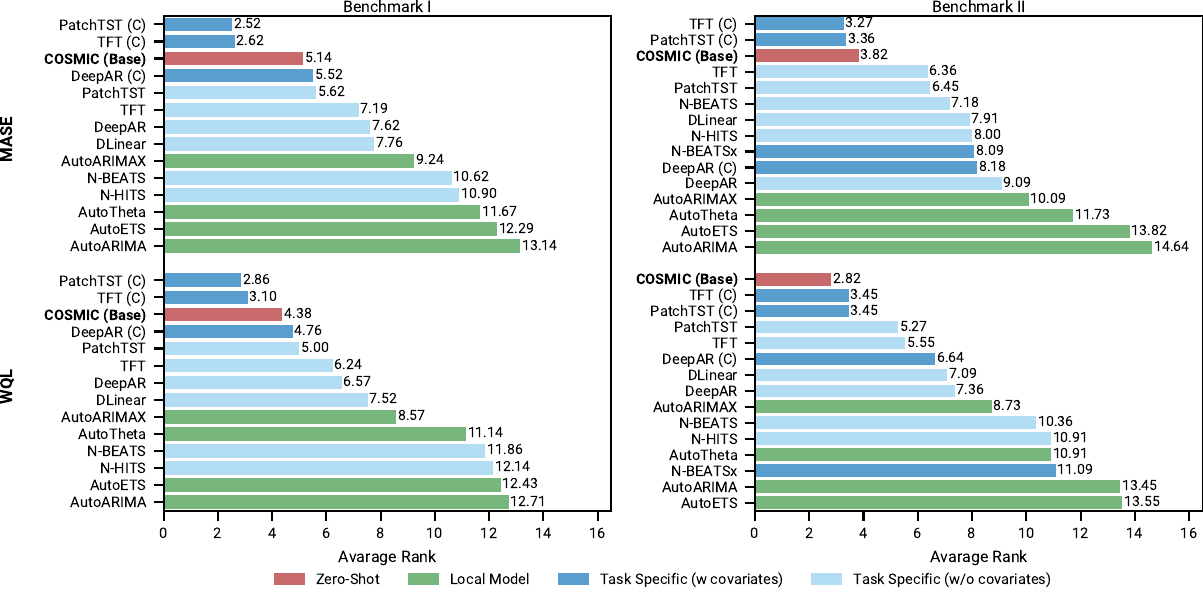}
    \caption{Average rank regarding MASE and WQL scores on the \underline{covariate-included evaluation data}, comparing different task-specific and local models compared to \modelname.
    \benchcaption.
    }
    \label{fig:task-specific-rank}
\end{figure*}

Figure~\ref{fig:task-specific-rank} presents the average rank over the different dataset groups when comparing \modelname (Base) to the state-of-the-art task-specific and local models.
These models are trained individually per dataset while \modelname does zero-shot forecasting.
The order of the model very similar than in the aggregate metric comparison in Figure~\ref{fig:main-taskspecifc-comparison}.

The metric results for the individual datasets and forecast horizons are presented in Table~\ref{tab:individual-results-taskspecific-mase} (MASE) and Table~\ref{tab:individual-results-taskspecific-wql} (WQL).

\begin{figure}[h]
    \centering
    \includegraphics[width=\textwidth]{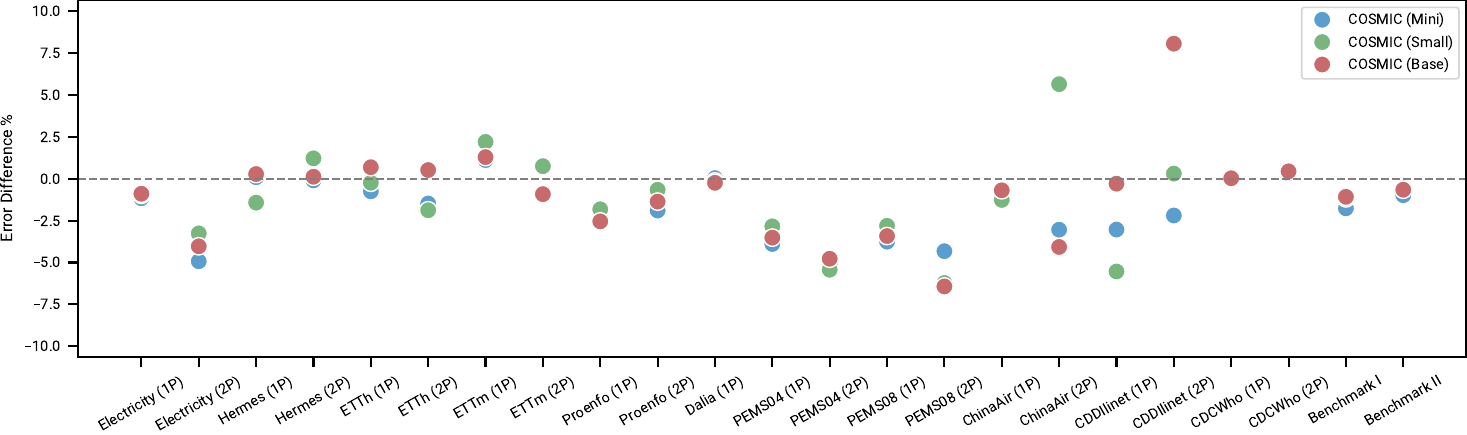}
    \caption{Relative MASE difference for all covariate-included dataset evaluations and the aggregate results when covariates are provided to \modelname.}
    \label{fig:app-covariate-advantage_2}
\end{figure}

\begin{figure}[]
    \centering
    \includegraphics[width=\textwidth]{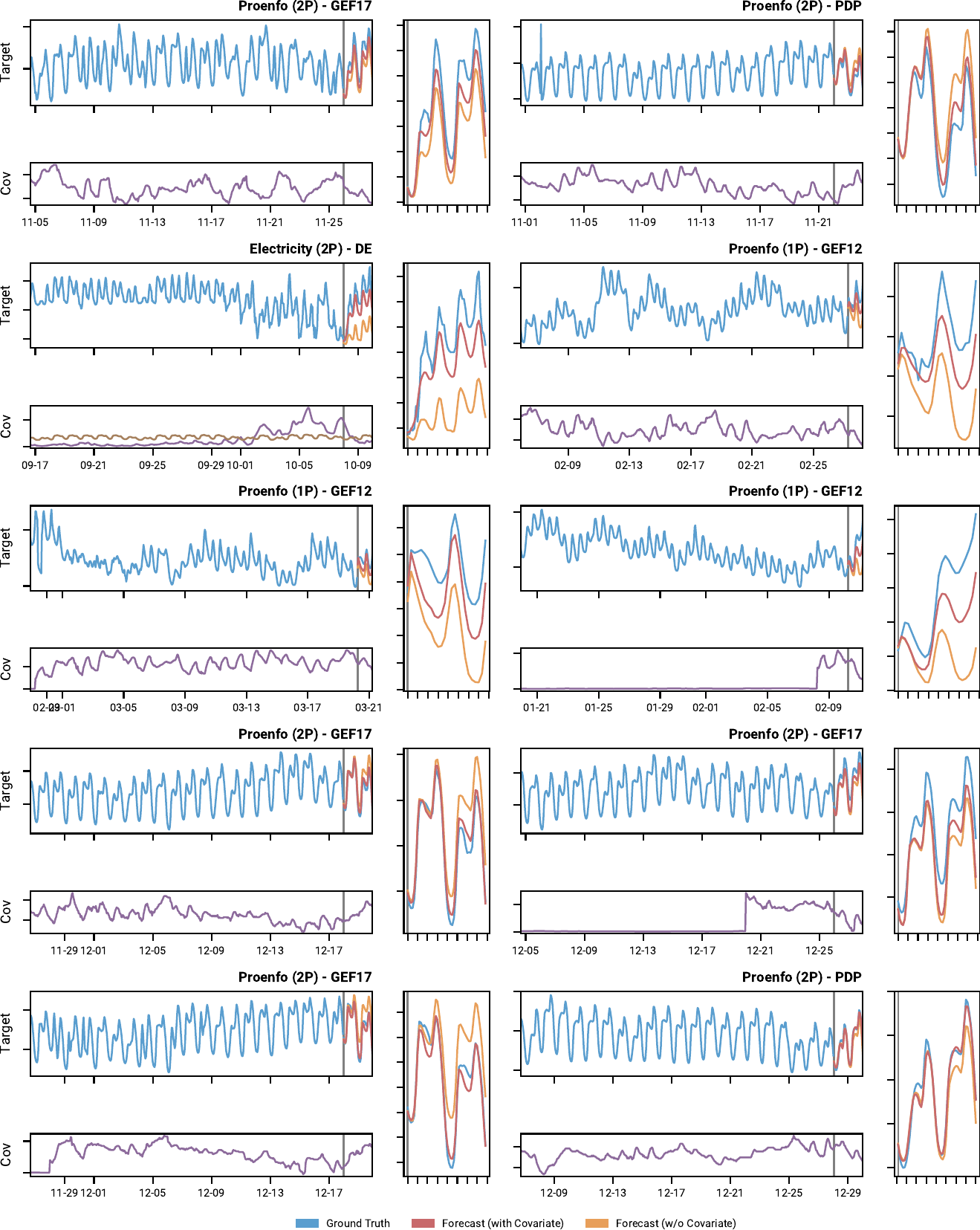}
    \caption{Multiple forecasts of \modelname with and without the access to covariates.
    Each example includes three plots: the full target signal (top-left), the covariate signals (bottom-left), and a zoomed-in view of the forecast horizon (right).
    The dataset of the example as well as the forecast horizon (1 or 2 periods) is shown as the title.}
    \label{fig:appendix-qualitative}
\end{figure}

\begin{figure}[]
    \centering
    \includegraphics[width=\textwidth]{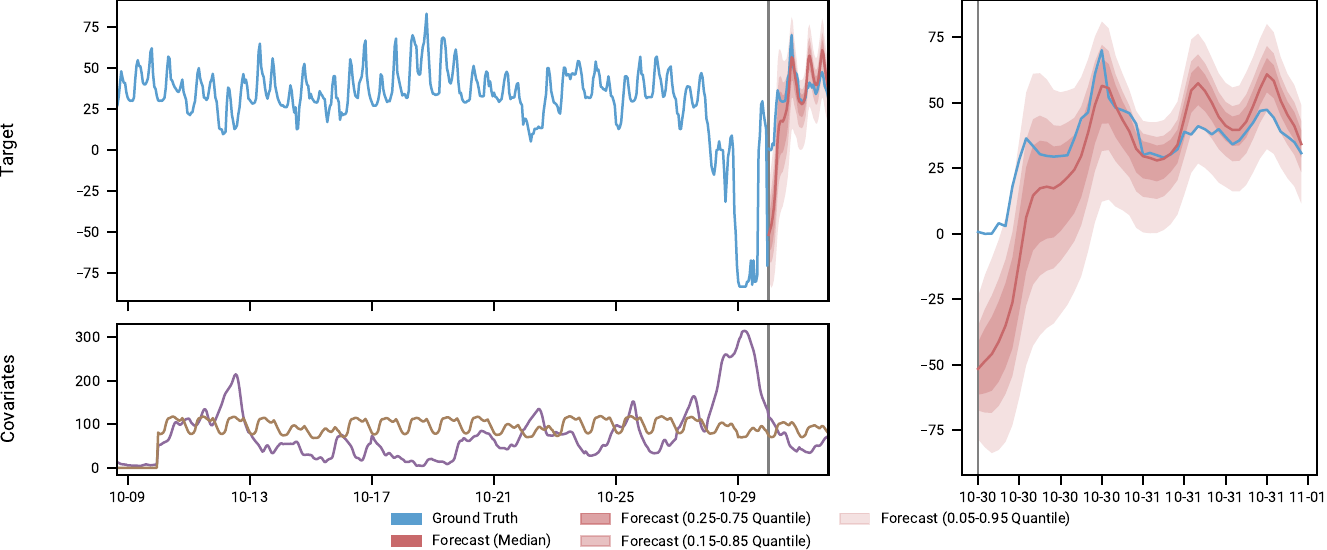}
    \caption{Forecast of \modelname including a visualization of the quantile predictions.
    Three plots are shown: the full target signal (top-left), the covariate signals (bottom-left), and a zoomed-in view of the forecast horizon (right).
    The example corresponds to the forecast also shown in Figure~\ref{fig:main-qualitative} (Electricity DE).}
    \label{fig:appendix-qualitative-ucinterval}
\end{figure}

\paragraph{Impact of Covariates \& augmentation}
Figure~\ref{fig:app-covariate-advantage_2} presents the performance difference when providing covariates to \modelname over all datasets.
Figure~\ref{fig:appendix-qualitative} shows additional forecast examples on real-world data where \modelname is able to utilize covariate information beneficially.
Figure~\ref{fig:appendix-qualitative-ucinterval} shows a forecast that also visualizes the quantile predictions of \modelname.

\subsection{No-Covariate Data}
\begin{figure*}[h]
    \centering
    \includegraphics[width=\textwidth]{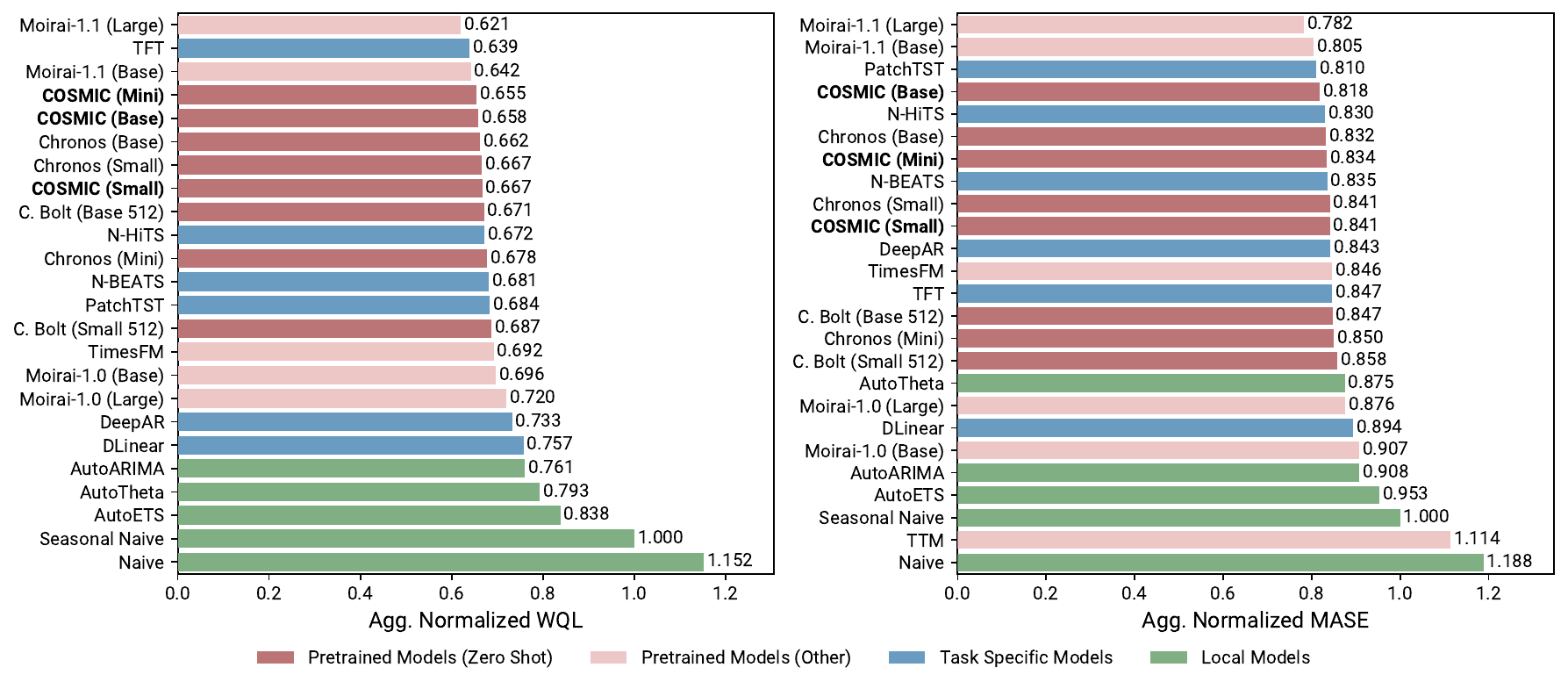}
    \caption{Aggreagted MASE and WQL scores on the \underline{no-covariate zeros-shot evaluation benchmark}.
    Lower values are better.
    The individual scores of the dataset groups are normalized by naive seasonal scores before aggregation.
    The ``Pretrained Models (Others)'' category refers to models where individual datasets overlap with their training data --- hence, they are not fully zero-shot (overlap ratio: Moirai 81\%, TimesFM 11\%, TTM 11\%).}
    \label{fig:agg-univariate-zeroshot-appendix}
\end{figure*}

Extending Figure~\ref{fig:agg-univariate-zeroshot} in the main paper Figure~\ref{fig:agg-univariate-zeroshot-appendix} presents the aggreagted MASE and WQL scores for all evaluated models in all sizes.
Notably, \modelname Mini (20M parameter) outperforms the much larger Chronos Base (200 M parameter) and Chronos Bolt Base (200 M parameter) in terms of WQL, as well as Chronos Small (40M parameter) and Chronos Bolt Base in terms of MASE.
Table~\ref{tab:uni-zeroshot-mase}~and~\ref{tab:uni-zeroshot-wql} show the individual dataset results, the aggregate metrics and the average rank of the comparison of zero-shot models on the zero-shot benchmark.
Similarly, Tables~\ref{tab:uni-zeroshot-taskspecific-mase} and \ref{tab:uni-zeroshot-taskspecific-wql} provide equivalent information for the comparison of \modelname with task-specific and local models.
Table~\ref{tab:uni-indomain-mase}-\ref{tab:uni-indomain-wql} presents the metrics of all models on the in-domain benchmark of \cite{ansariChronosLearningLanguage2024b}.

\clearpage

\section{Additional Analysis \& Experiments}

This section complements Section~\ref{sec:experiments} by presenting additional experiments and analyses. Specifically:
Subsection~\ref{sec:app-past-cov} examines the impact of past-only covariates on model performance.
Subsection~\ref{sec:app-morai-comp} analyzes the performance and covariate usage of Moirai in more detail, considering all available variants.
Subsection~\ref{sec:app-long-ctx} presents results for models with extended context lengths.
Subsection~\ref{sec:app-inference-time} analyzes the inference time of pre-trained models and examines the influence of covariates.
Subsection~\ref{sec:app-in-ctx-model} compares \modelname covariate handling with an alternative approach where a separate regression model is used for covariates.

\subsection{Past-only covariates}\label{sec:app-past-cov}

\begin{figure}[h]
    \centering
    \includegraphics[width=\textwidth]{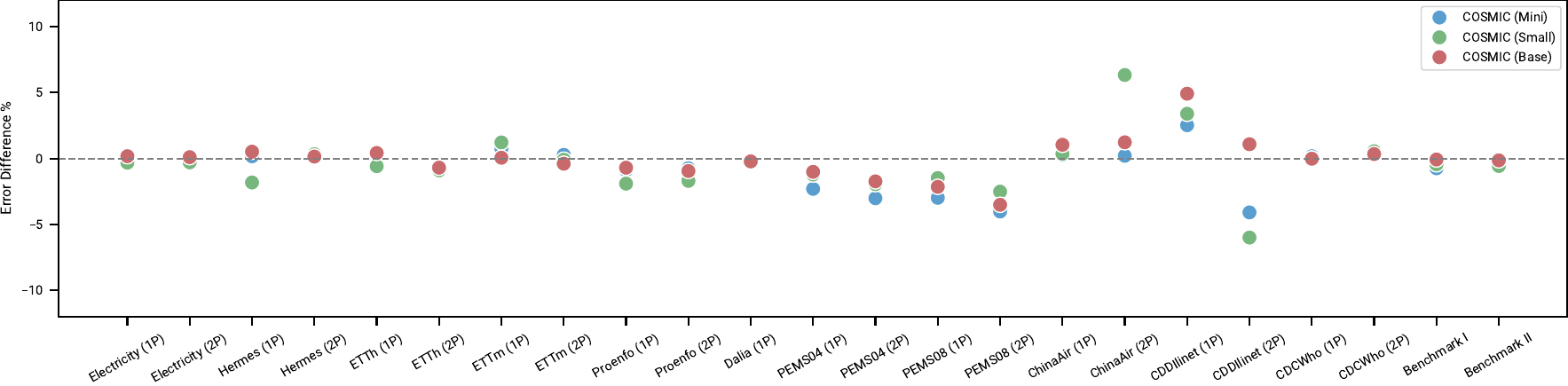}
    \caption{Relative MASE difference for all covariate-included dataset evaluations and the aggregate results when covariates are provided to \modelname \underline{as past-only covariates}.}
    \label{fig:app-past-covariate-advantage}
\end{figure}

In the main experiments, we focused on evaluations where covariates are treated as past and future covariates.
To get a more comprehensive perspective Figure~\ref{fig:app-past-covariate-advantage} presents the impact of incorporating past-only covariates in the input of \modelname.
The effect of past-only covariates is less pronounced compared to past and future covariates.
This is as expected due to their limited potential for additional information.
However, on individual datasets such as ProEnfo, PEMS04, and PEMS08, \modelname demonstrates notable improvements.
Additionally, there is minimal degradation for most datasets.

\begin{figure}[h]
    \centering
    \includegraphics[width=\textwidth]{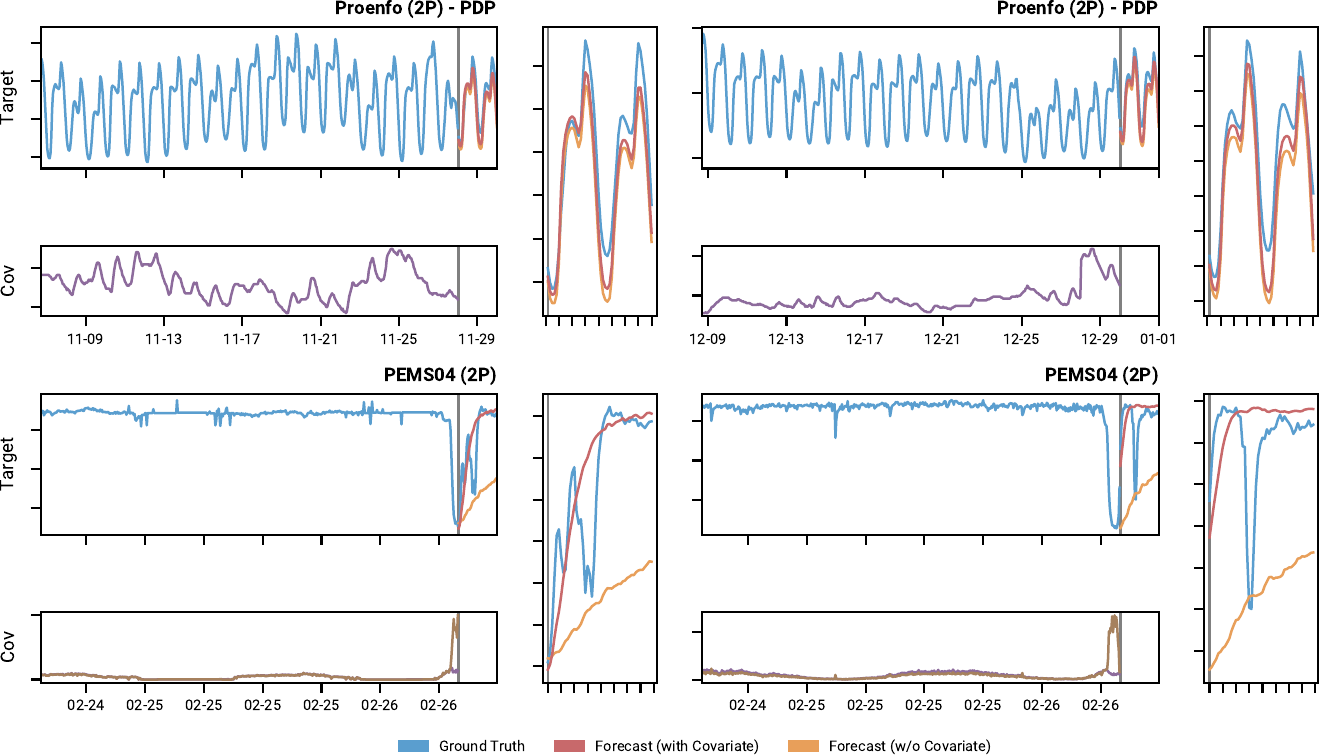}
    \caption{Four forecasts of \modelname with and without the access to covariates \underline{as past-only covarites}.
    Each example includes three plots: the full target signal (top-left), the covariate signals (bottom-left), and a zoomed-in view of the forecast horizon (right).
    The dataset of the example as well as the forecast horizon (1 or 2 periods) is shown as the title.}
    \label{fig:app-past-covariate-quali}
\end{figure}

To explore how past-only covariates enhance \modelname's forecasting capabilities, we qualitatively analyzed its behavior when past-only covariates are included.
Figure~\ref{fig:app-past-covariate-quali} highlights four real-world scenarios where \modelname effectively leverages past-only covariates.

\clearpage

\subsection{Comparison with Moirai}\label{sec:app-morai-comp}

\begin{figure}[]
    \centering
    \includegraphics[width=\textwidth]{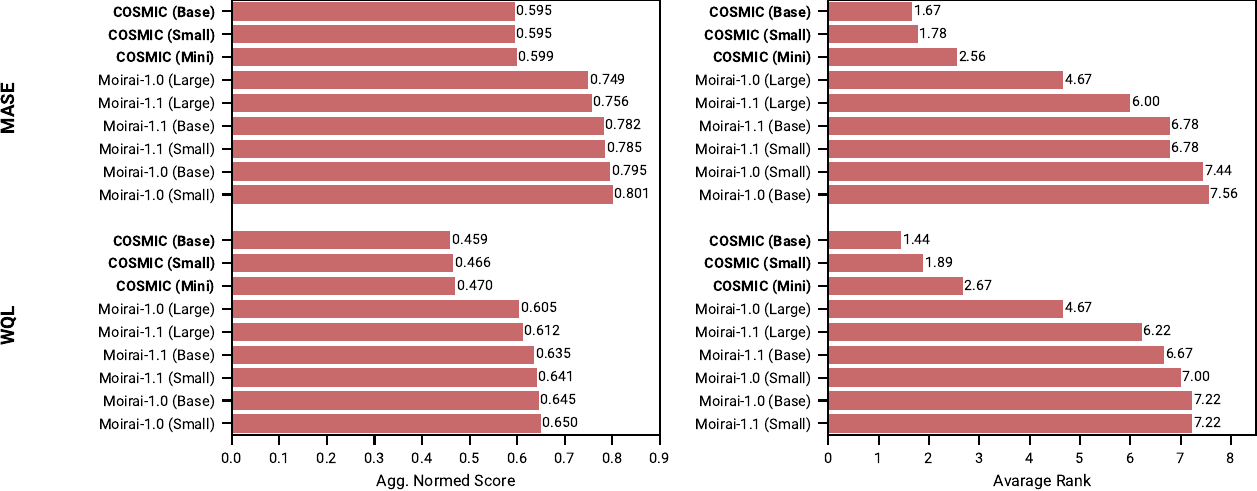}
    \caption{Aggregated MASE and WQL scores and the average ranks regarding all datasets that are not part of the training corpus of Moirai.
    All available Moirai versions are compared with \modelname.
    Consistent with other aggregations, individual scores for dataset groups are normalized using naive seasonal scores prior to aggregation.}
    \label{fig:morai-comparison}
\end{figure}

In our covariate-inclusive evaluation setup (Benchmark I and II), multiple datasets overlap with Moirai's training data.
As a result, Moirai's performance on these datasets cannot be considered entirely zero-shot, which hinders a clear zero-shot comparison.
Given that Moirai is the only other model that, in principle, supports the input of covariate signals in a zero-shot setting, a thorough comparison and analysis are warranted.
To address this, we present in Figure~\ref{fig:morai-comparison} an aggregated comparison focusing exclusively on datasets not included in Moirai's training set.
Our findings indicate that \modelname outperforms Moirai across all model sizes, both for versions 1.0 and 1.1.

To further investigate Moirai’s use of covariates, we conducted an analysis similar to the methodology outlined in Section~\ref{sec:result-cov-dataset}.
We evaluated Moirai's performance with and without covariate inputs across several datasets.
The impact of incorporating covariates varied: some datasets showed notable performance improvements, while others exhibited significant degradation.
On aggregate, only Moirai Small shows an improvement with covariates, whereas Moirai Base and Large show diminished performance.
We hypothesize that this inconsistency stems from Moirai’s training setup, which combines real-world covariate data with arbitrary signals drawn from its training corpus for its training.
While real-world covariate data might be too limited for a robust generalization, the arbitrary signals likely do not have any predictive value for the target.

\begin{figure}[]
    \centering
    \includegraphics[width=\textwidth]{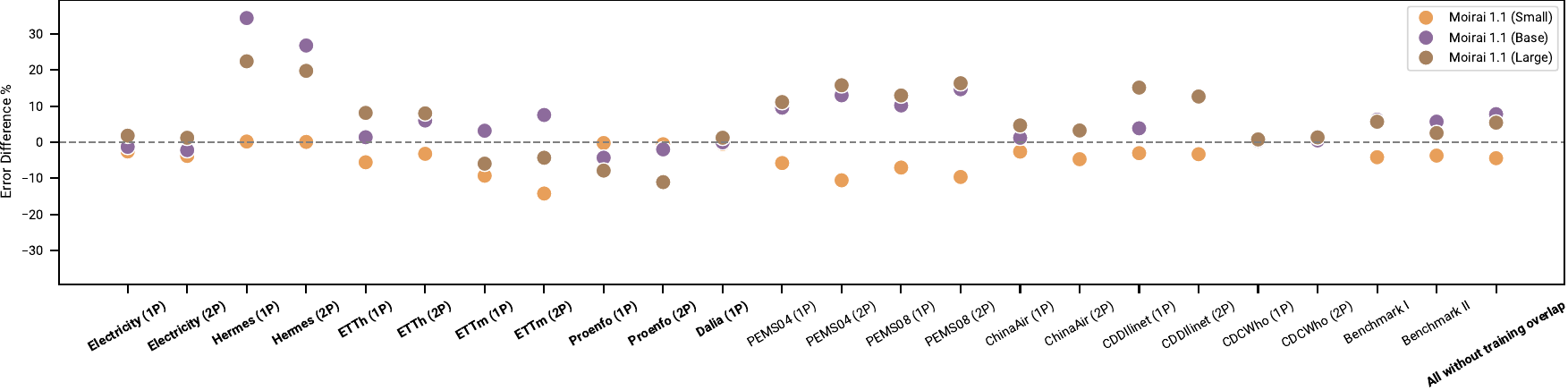}
    \caption{Relative MASE difference for all covariate-included dataset evaluations and the aggregate results when covariates are provided to \underline{Moirai}.
    Datasets that do not overlap with the training data, and therefore are zero-shot, are marked bold.}
    \label{fig:app-morai-covariate-advantage}
\end{figure}

\clearpage

\subsection{Extended Context}\label{sec:app-long-ctx}

\begin{figure}[]
    \centering
    \includegraphics[width=\textwidth]{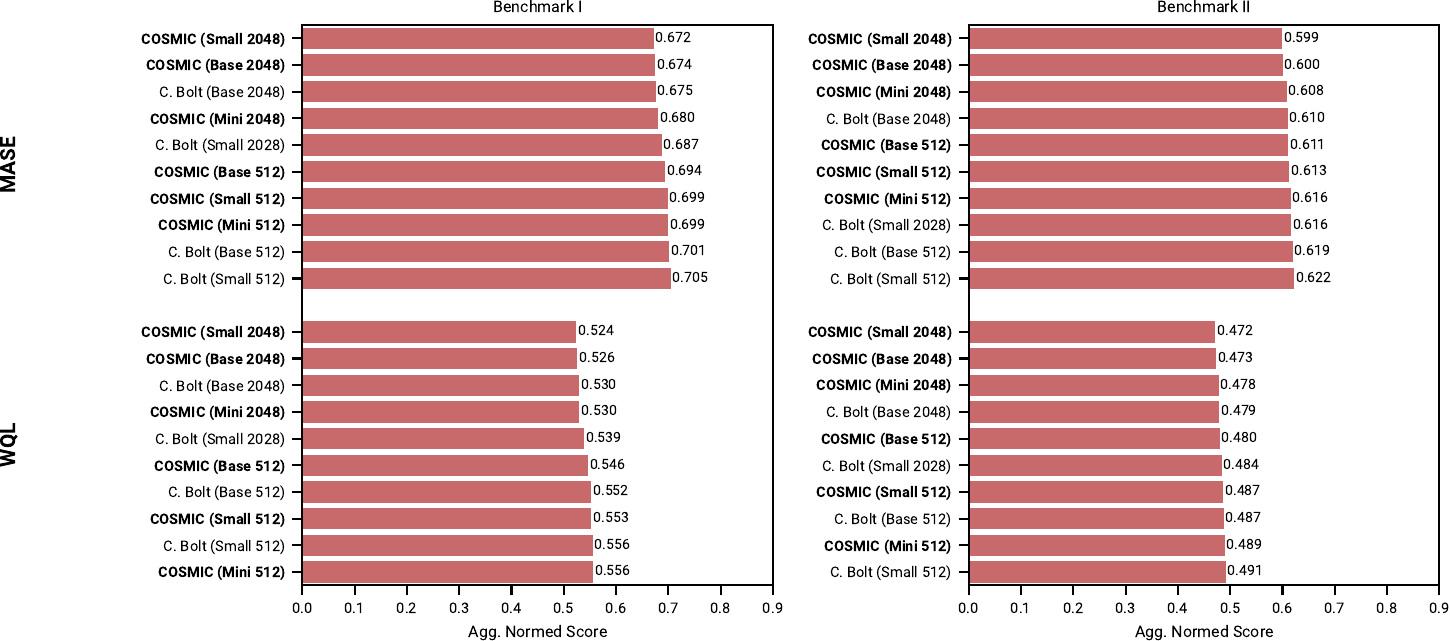}
    \caption{Aggreagted  MASE and WQL scores on the \underline{covariate-included evaluation data}, comparing \modelname and Chronos Bolt with 512 and 2048 context length.
    \benchcaption.
    }
    \label{fig:2k-comparison}
\end{figure}

In our main experiments, we utilized a context length of 512 for COSMIC and all Chronos variants (Chronos and Chronos Bolt), aligning with the default configuration of the original Chronos model.
However, for Chronos Bolt, models with a context length of 2048 were published.
To facilitate a direct comparison, we trained additional \modelname instances with a context length of 2048.
All other hyperparameters were kept constant.
Figure~\ref{fig:2k-comparison} shows the results of the experiment on the covariate-inclusive datasets.
The increased context length improves the performance of all models notably.
The relative difference between Chronos and Chronos Bolt remains steady across models with the same context length, indicating a sustained advantage in covariate utilization.

\subsection{Inference Time}\label{sec:app-inference-time}
\begin{figure}[]
    \centering
    \includegraphics[width=0.9\textwidth]{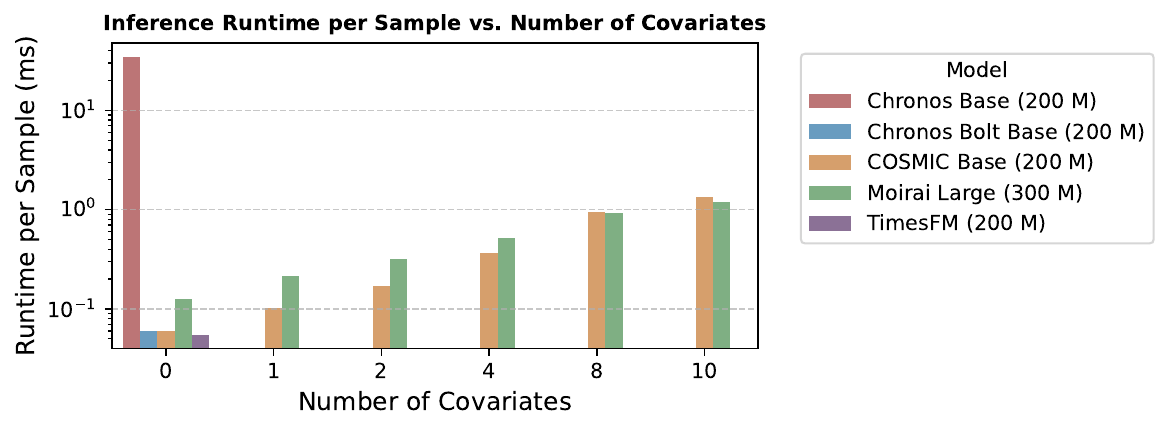}
    \caption{Inference runtime analysis: Runtime per sample depending on the number of covariates of the sample for different pre-trained models. Only Cosmic and Moirai support the utilization of covariates.}
    \label{fig:inference-time}
\end{figure}

Figure~\ref{fig:inference-time} presents an inference time analysis and the impact of the number of covariates.
The experiment evaluates samples with a context length of $512$ and a forecast horizon length of $64$.
As expected, increasing the number of covariates significantly increases the runtime.
However, due to the efficient patching mechanism, inference times remain reasonable --- e.g. the runtime of \modelname Base with $10$ covariates is still over one magnitude below the runtime of Chronos Base.
The runtime of \modelname and Moirai scales similarly, as both append additional covariates to the model context.
Other pre-trained models do not support covariates.

\clearpage

\subsection{Linear In-Context Covariate Model}\label{sec:app-in-ctx-model}

An alternative to incorporating covariates directly within the pretrained model is to leverage an external in-context regression model.
To account for (1) the limited data in-context and (2) the need for model fitting in the inference of every forecast,  linear regression is a viable choice.
The methods works as follows:
First, we regress the $k$ covariates $\{\mathbf{x}^1_{1:T},\dots, \mathbf{x}^k_{1:T}\}$ on the target value $\mathbf{y}_{1:T}$ given the context of time series --- i.e.\ we estimate the coefficients $a_i$ in
\begin{align*}
    \tilde{y}_t =& x^1_t * a_1 + \ldots x^k_t * a_k + \varepsilon_t \quad \text{with} \; \varepsilon_t \sim N(0, \sigma)  \\
\end{align*}

Next, we compute the residuals $\mathbf{r}_{0:T}$ which represent the part of the target signal not captured by the linear regression model --- i.e.\ the signal that is not determined by the covariates in our model.
\begin{align*}
    \mathbf{r}_{0:T} =& \mathbf{y}_{0:T} - \tilde{\mathbf{y}}_{0:T}. \\
\end{align*}

These residuals, which form a time series themselves, are then forecasted using the pretrained model.
Finally, the forecasted residuals $\hat{\mathbf{r}}_{T:T+h}$ are combined with the estimation of the in-context model for the overall forecast $\hat{\textbf{y}}_{T:T+h}$:
\begin{align*}
    \hat{\mathbf{y}}_{T:T+h} =& \hat{\mathbf{r}}_{T:T+h} + \tilde{\mathbf{y}}_{T:T+h}.
\end{align*}

To avoid overfitting we utilize L2-Regularization, i.e. we employ ridge regression.
Note that this approach has several limitations:
(1) The method cannot handle past-only covariates as we need to compute the linear model for the forecast horizon.
(2) To account for lagged covariate impact, we would need to specify beforehand which lag parameters we want to estimate.
This requires careful consideration, as each additional lag increases the number of model parameters, while the number of samples available for fitting the model is limited to the context length, which is relatively small.
Moreover, the usable context length of the pretrained model is reduced by the maximum applied lag.
(3) The model can not leverage any global information to decide which covariate model might be more or less likely for the given covariates.
The latter hinders also any potential improvement with scaling the model and \/or the training data.

\begin{figure}[]
    \centering
    \includegraphics[width=\textwidth]{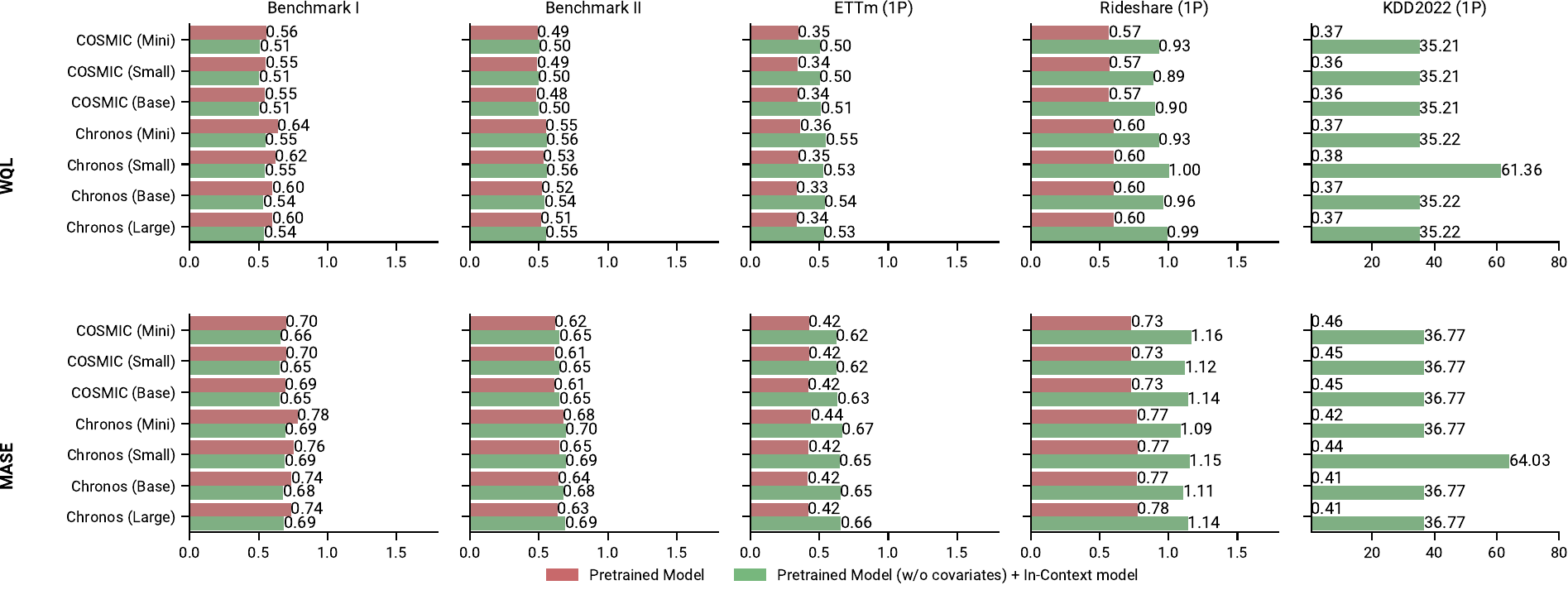}
    \caption{WQL and MASE scores for different sized Chronos and \modelname models compared to Chronos and \modelname without access to covariates in combination with a linear in-context covariate model. The comparison with \modelname highlights the difference between the two variants of utilizing covariates. The comparison with Chronos highlights how the external in-context model changes the no-covariate prediction performance.
    \benchcaption.
    }
    \label{fig:ph-quantative}
\end{figure}

\paragraph{Results}
Despite these limitations, the approach works very well for a lot of datasets.
Figure~\ref{fig:ph-quantative} compares \modelname when covariates are provided with \modelname without any covariates but used with a linear in-context model.
While on the aggregate across all datasets the linear in-context model approach performs best (\benchall) --- \modelname performs best when considering only datasets without target signals as covariates (\benchnotarget).
However, we also observe severe failure cases for some datasets.
For example, in the ETTm \cite{zhouInformerEfficientTransformer2021a} and the Rideshare \cite{godahewa_2021_5122114} dataset we observe approximately 50\% increase in error scores.
For the KDD dataset \cite{wooUnifiedTrainingUniversal2024a, zhouSDWPFDatasetSpatial2022}, the error scores increase drastically, exceeding 3500\%.
Note that the latter two datasets are not in our covariate-inclusive evaluation dataset  -- hence also not considered in \benchall, because the no-covariate versions of it are part of the Chronos training corpus, and therefore also not considered in the aggregate metrics.

\begin{figure}[]
    \centering
    \includegraphics[width=\textwidth]{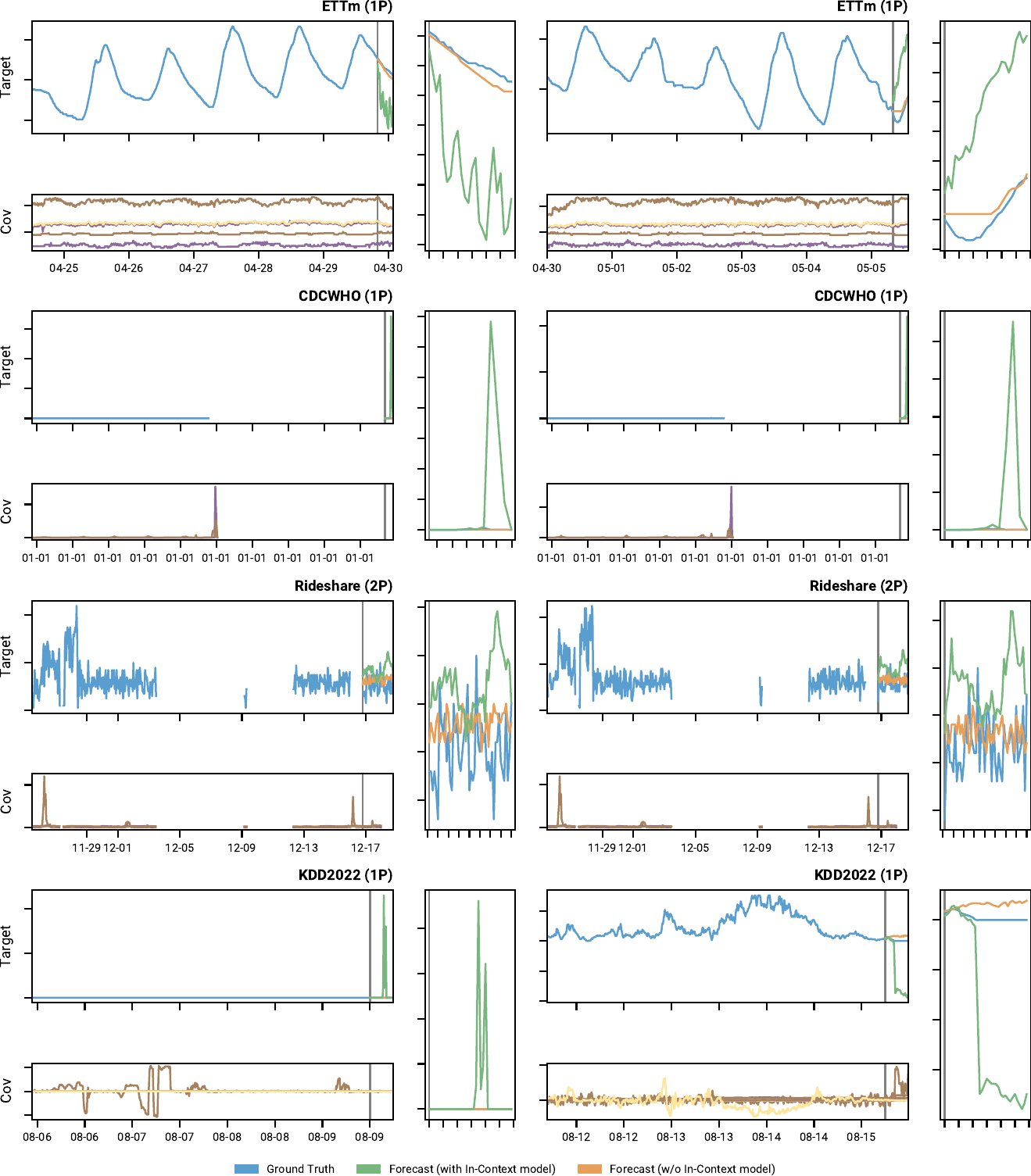}
    \caption{Multiple forecasts for which the application of an external linear in-context covariate model results in a considerably increased error.
    Each example includes three plots: the full target signal (top-left), the covariate signals (bottom-left), and a zoomed-in view of the forecast horizon (right).
    The dataset of the example as well as the forecast horizon (1 or 2 periods) is shown as the title.}
    \label{fig:appendix-ph-qualitiative}
\end{figure}

The qualitative analysis, shown in Figure~\ref{fig:appendix-ph-qualitiative}, illustrates examples of these failure cases.
We find that the primary issue is the distributional shift of covariates in the forecast horizon.
Often this distribution shift is in scenarios where the variance of the covariate in the context is much lower than in the forecast horizon.
\modelname seems to be more robust to such shifts.

\paragraph{Frequency of impact observations}
Additionally, we investigated why the in-context model performs better than \modelname in some instances.
We hypothesize that \modelname is more cautious in utilizing signals that it infers from covariates.
While this might reduce performance on some datasets, it might also help to avoid the failure cases presented before.

To empirically explore this hypothesis we draft an experiment as follows:
We created a dataset with synthetic covariates with Bell-shaped events and deterministic impact on the covariate.
The target value is drawn from the Chronos training corpus.
We deterministically set how much of such events should be present in the context.
The covariates are designed so that \emph{in expectation} one covariate in each sample exhibits an additional event in the forecast horizon.
We generate 200 samples for each number of events/number of covariates combination.
We evaluated the dataset with \modelname, and covariate-excluded \modelname with and without a linear in-context model.

\begin{figure}[]
    \centering
    \includegraphics[width=\textwidth]{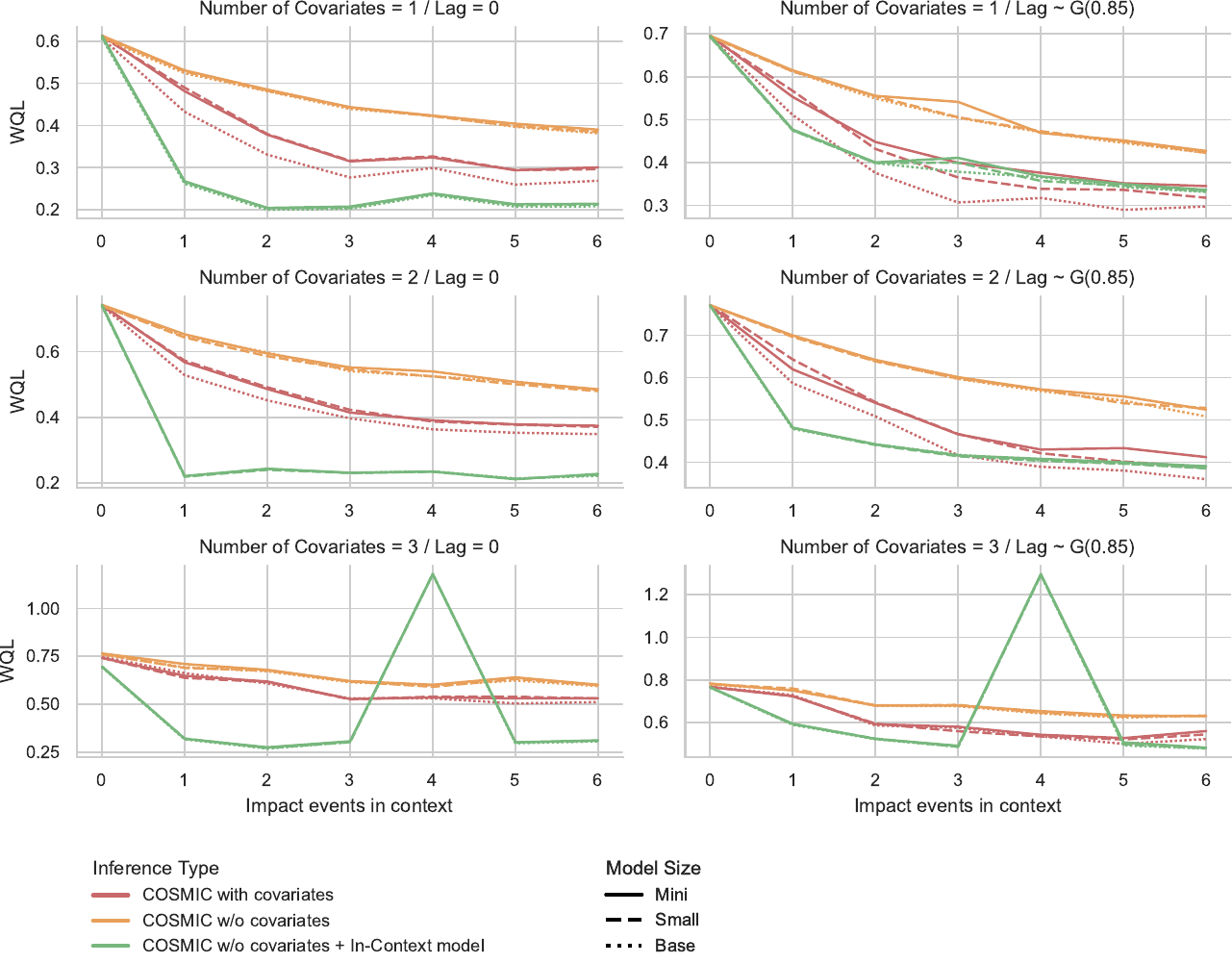}
    \caption{ 
    Analysis of the effect of varying numbers of impact observations in the context, on utilization of the covariate-target relationship in the forecast.
    The y-axis represents the average WQL across all series of a certain setting.
    The x-axis shows the number of bell events of one covariate within the context.
    Left column plots: Results for the data where the covariate impact has no lag (lag=0).
    Right column plots: Results for the data where the impact lags are sampled from a geometric distribution G(0.85).
    The rows represent results for different data depending on the number of covariates in one sample.
    The line color indicates the inference type that is used and the line style the size of the model.
    The model parameters for each size remain consistent across inference types, meaning all methods utilize the same pre-trained models.}
    \label{fig:impact-sensitivity}
\end{figure}

Figure~\ref{fig:impact-sensitivity} presents the results of the experiment. 
In the dataset where the impact is never shifted via a lag, the external linear in-context model can reduce the error score more than \modelname when events can be observed in the context.
Note that this is the best-case scenario for the external in-context model.
As hypothesized this gap is indeed the biggest when a low number of impact events are observed.
Also, the gap between \modelname with and without covariates increases with an increased amount of event observations in the context.
This suggests that \modelname does more likely and strongly incorporate the covariate signal in its forecast when the covariate-target relationship is visible in more observation --- i.e.\ when the model is more confident.
When we sample the impact lag from a geometric distribution \modelname starts to outperform the in-context models when multiple event observations are available.
Comparing the different model sizes we find that the performance of the covariate-included \modelname improves with increasing model size more than in the no-covariate setting.
The outlier result for the external in-context model in the ``3 covariates and 4 impact events'' setting is due to a severe failure cases of the in-context model.

\paragraph{Conclusion}
In summary, a simple external in-context model can be a viable approach for enhancing forecasting models with covariate modeling capabilities.
However, its susceptibility to severe failure cases calls for deliberated decisions from a practical perspective.  
The missing potential for scaling improvements diminishes its appeal from a research perspective.

\clearpage

\section{Societal Impact}\label{sec:societal-impact}

The proposed pre-trained time series models can democratize access to advanced forecasting tools, enabling practitioners without machine learning expertise to generate predictions across diverse domains.
However, overreliance on such models—particularly when their limitations (as outlined in Section~\ref{sec:limitation}) are not adequately considered—may lead to misinformed conclusions. 
The societal consequences of erroneous forecasts depend on the specific application context.

\clearpage

\setNextCaption{MASE scores of different zero-shot models on \underline{covariate-included datasets}. The models achieving the \textbf{best} and \underline{second-best} scores are highlighted. Results for datasets that are part of the training data for the respective models are shaded in grey, and these results are excluded from the calculation of the best score.}
\input{tables/individual-results-zs-mase}

\setNextCaption{WQL scores of different zero-shot models on \underline{covariate-included datasets}. The models achieving the \textbf{best} and \underline{second-best} scores are highlighted. Results for datasets that are part of the training data for the respective models are shaded in grey, and these results are excluded from the calculation of the best score.}
\input{tables/individual-results-zs-wql}

\setNextCaption{MASE scores of \modelname (Base) compared with various task-specific and local models on \underline{covariate-included datasets}. Task-specific model variants that support covariates are indicated with a (C) postfix. Models achieving the \textbf{best} and \underline{second-best} scores are highlighted. N-BEATAS (C) failed for some datasets (indicated with nan) and was therefore not considered for the respective benchmark aggregation.}
\input{tables/individual-results-taskspecific-mase}

\setNextCaption{WQL scores of \modelname (Base) compared with various task-specific and local models on \underline{covariate-included datasets}. Task-specific model variants that support covariates are indicated with a (C) postfix. Models achieving the \textbf{best} and \underline{second-best} scores are highlighted. N-BEATAS (C) failed for some datasets (indicated with nan) and was therefore not considered for the respective benchmark aggregation.}
\input{tables/individual-results-taskspecific-wql}

\setNextCaption{MASE scores of different zero-shot models on the \underline{no-covariate zero-shot benchmark}. 
The models achieving the \textbf{best} and \underline{second-best} scores are highlighted. 
Results for datasets that are part of the training data for the respective models are shaded in grey.
These results are excluded from the calculation of the best score but are included in the aggregated score and average rank.
The aggregated score is calculated as described in Section~\ref{sec:experiments}.
}
\input{tables/uni-zeroshot-mase}

\setNextCaption{WQL scores of different zero-shot models on the \underline{no-covariate zero-shot benchmark}.
The models achieving the \textbf{best} and \underline{second-best} scores are highlighted.
Results for datasets that are part of the training data for the respective models are shaded in grey.
These results are excluded from the calculation of the best score but are included in the aggregated score and average rank.
The aggregated score is calculated as described in Section~\ref{sec:experiments}.
}
\input{tables/uni-zeroshot-wql}

\setNextCaption{MASE scores of \modelname compared with various task-specific and local models on the \underline{no-covariate zero-shot benchmark}.
Models achieving the \textbf{best} and \underline{second-best} scores are highlighted.
The aggregated score is calculated as described in Section~\ref{sec:experiments}.
For models that failed on certain datasets (indicated with nan), we used a relative score of 1, i.e., the seasonal naive relative score, when aggregating the results.
}
\input{tables/uni-zeroshot-taskspecific-mase}

\setNextCaption{WQL scores of \modelname compared with various task-specific and local models on the \underline{no-covariate zero-shot benchmark}.
Models achieving the \textbf{best} and \underline{second-best} scores are highlighted.
The aggregated score is calculated as described in Section~\ref{sec:experiments}.
For models that failed on certain datasets (indicated with nan), we used a relative score of 1, i.e., the seasonal naive relative score, when aggregating the results.
}
\input{tables/uni-zeroshot-taskspecific-wql}

\setNextCaption{MASE scores of different pre-trained, task-specific, and local models on the \underline{no-covariate in-domain benchmark}.
Models achieving the \textbf{best} and \underline{second-best} scores are highlighted.
Results for datasets where the evaluation part overlaps with the training data for the respective models are shaded in grey.
These results are excluded from the calculation of the best score but are included in the aggregated score and average rank.
The aggregated score is calculated as described in Section~\ref{sec:experiments}. For models that failed on certain datasets (indicated with nan), we used a relative score of 1, i.e., the seasonal naive relative score, when aggregating the results.
}
\input{tables/uni-indomain-mase}

\setNextCaption{WQL scores of different pre-trained, task-specific, and local models on the \underline{no-covariate in-domain benchmark}.
Models achieving the \textbf{best} and \underline{second-best} scores are highlighted.
Results for datasets where the evaluation part overlaps with the training data for the respective models are shaded in grey.
These results are excluded from the calculation of the best score but are included in the aggregated score and average rank.
The aggregated score is calculated as described in Section~\ref{sec:experiments}. For models that failed on certain datasets (indicated with nan), we used a relative score of 1, i.e., the seasonal naive relative score, when aggregating the results.}
\input{tables/uni-indomain-wql}

%% file: tables/hyperparam_augmentation.tex
\begin{table}[h]
\centering
\caption{Hyperparameter used for the informed covariate augmentation for training \modelname}
\begin{tabular}{@{}lll@{}}
\toprule
\textbf{Paramter}     & \textbf{Value} & \textbf{Description}                  \\ \midrule
$p$               & 0.25           & Covariates count parameter                   \\
$p_{\mathrm{FO}}$            & 0.2            & Probability for first order           \\
$p_{\mathrm{PW}}$            & 0.15           & Probability for piece-wise            \\
$k_{\mathrm{max}}$           & 10             & Maximum count                         \\
$p_{\mathrm{lagcount}}$      & 0.85           & Impact lag count parameter                       \\
$p_{\mathrm{lagpos}}$        & 0.15           & Impact lag position parameter                  \\
$l$             & 500            & Maximum lag                           \\
$s_\epsilon$        & 0.02           & Noise scale                           \\ 
\midrule
$c^{\mathrm{max}}_{\mathrm{e}}$      & 20           & Maximum synth. covariate events                       \\
$c^{\mathrm{max}}_{\mathrm{cp}}$     & 8           & Maximum synth. covariate change-points  
\\
$\sigma^{\mathrm{cp}}$      & 2           & Change-point variance   \\
\midrule
$\mathrm{Geom}$                  &              & Geometric Distribution                 \\ 
$U$                  &              & Uniform Distribution / Sample                   \\ 
$N$                  &              & Gaussian Distribution \\
\bottomrule
\end{tabular}
\label{tab:hyperparameter-augmentation}
\end{table}

%% file: tables/dataset_info_cov.tex
\begin{table}[h]
\centering
\caption{\nextcaption}
\begin{small}
\begin{tabular}{lllrrrrl}
\toprule
Data group & Dataset & Source & \rotatebox{90}{\# Covariate} & \rotatebox{90}{\# Targets} & \rotatebox{90}{Period} & \rotatebox{90}{\# Series} & \rotatebox{90}{\# Rolls}\\
\midrule
\multirow[t]{5}{*}{Electricity} & Electricity NP & \citep{lagoForecastingDayaheadElectricity2021, olivaresNeuralBasisExpansion2023} & 2 & 1 & 24 & 1 & 216/108 \\
 & Electricity PJM & \citep{lagoForecastingDayaheadElectricity2021, olivaresNeuralBasisExpansion2023} & 2 & 1 & 24 & 1 & 216/108 \\
 & Electricity FR & \citep{lagoForecastingDayaheadElectricity2021, olivaresNeuralBasisExpansion2023} & 2 & 1 & 24 & 1 & 216/108 \\
 & Electricity BE & \citep{lagoForecastingDayaheadElectricity2021, olivaresNeuralBasisExpansion2023} & 2 & 1 & 24 & 1 & 216/108 \\
 & Electricity DE & \citep{lagoForecastingDayaheadElectricity2021, olivaresNeuralBasisExpansion2023} & 2 & 1 & 24 & 1 & 216/108 \\
\cline{1-8}
\multirow[t]{8}{*}{ProEnfo} & BDG2-Bull & \citep{wooUnifiedTrainingUniversal2024a, wangBenchmarksCustomPackage2024} & 3 & 1 & 24 & 41 & 70/35 \\
 & BDG2-Hog & \citep{wooUnifiedTrainingUniversal2024a, wangBenchmarksCustomPackage2024} & 5 & 1 & 24 & 24 & 70/35 \\
 & BDG2-Cockatoo & \citep{wooUnifiedTrainingUniversal2024a, wangBenchmarksCustomPackage2024} & 5 & 1 & 24 & 1 & 70/35 \\
 & Covid19-Energy & \citep{wooUnifiedTrainingUniversal2024a, wangBenchmarksCustomPackage2024} & 6 & 1 & 24 & 1 & 130/65 \\
 & GEF12 & \citep{wooUnifiedTrainingUniversal2024a, wangBenchmarksCustomPackage2024} & 1 & 1 & 24 & 11 & 162/81 \\
 & GEF14 & \citep{wooUnifiedTrainingUniversal2024a, wangBenchmarksCustomPackage2024} & 1 & 1 & 24 & 1 & 70/35 \\
 & GEF17 & \citep{wooUnifiedTrainingUniversal2024a, wangBenchmarksCustomPackage2024} & 1 & 1 & 24 & 8 & 70/35 \\
 & PDP & \citep{wooUnifiedTrainingUniversal2024a, wangBenchmarksCustomPackage2024} & 1 & 1 & 24 & 1 & 70/35 \\
\cline{1-8}
ETTh & ETTh & \citep{zhouInformerEfficientTransformer2021a} & 5 & 1 & 24 & 2 & 70/35 \\
\cline{1-8}
ETTm & ETTm & \citep{zhouInformerEfficientTransformer2021a} & 5 & 1 & 24 & 2 & 288/144 \\
\cline{1-8}
Hermes & Hermes & \citep{davidHERMESHybridErrorcorrector2023} & 1 & 1 & 28 & 10000 & 1/1 \\
\cline{1-8}
PPGDalia & PPGDalia & \citep{ppg-dalia_495} & 7 & 1 & 60 & 15 & 7/0 \\
\cline{1-8}
PEMS04 & PEMS04 & \citep{wooUnifiedTrainingUniversal2024a, jiangLibCityUnifiedLibrary2024a} & 0 & 3 & 24 & 300 & 68/34 \\
\cline{1-8}
PEMS08 & PEMS08 & \citep{wooUnifiedTrainingUniversal2024a, jiangLibCityUnifiedLibrary2024a} & 0 & 3 & 24 & 170 & 72/36 \\
\cline{1-8}
ChinaAir & ChinaAir & \citep{wooUnifiedTrainingUniversal2024a, zhengForecastingFineGrainedAir2015} & 0 & 6 & 24 & 437 & 5/2 \\
\cline{1-8}
CDCIlinet & CDCIlinet & \citep{wooUnifiedTrainingUniversal2024a}\tablefootnote{CDCIlinet and CDCWho datasets in \citep{wooUnifiedTrainingUniversal2024a} are from \label{cdc-footnote}\url{https://gis.cdc.gov/grasp/fluview/fluportaldashboard.html}} & 0 & 5 & 13 & 75 & 1/1 \\
\cline{1-8}
CDCWho & CDCWho & \citep{wooUnifiedTrainingUniversal2024a}$^{\ref{cdc-footnote}}$ & 0 & 4 & 13 & 74 & 1/1 \\
\bottomrule
\end{tabular}

\end{small}
\label{tab:dataset-cov}
\end{table}

%% file: tables/dataset_info_uni.tex
\begin{table}[h]
\centering
\caption{\nextcaption}
\begin{small}
\begin{tabular}{llrrr}
\toprule
  &    & Offset & Horizon & \# Rolls\\
\midrule
\multirow[t]{15}{*}{In-Domain} & Electricity (15 Min.) & -5376 & 24 & 1 \\
 & Electricity (Hourly) & -24 & 24 & 1 \\
 & Electricity (Weekly) & -8 & 8 & 1 \\
 & KDD Cup 2018 & -48 & 48 & 1 \\
 & M4 (Daily) & -14 & 14 & 1 \\
 & M4 (Hourly) & -48 & 48 & 1 \\
 & M4 (Monthly) & -18 & 18 & 1 \\
 & M4 (Weekly) & -13 & 13 & 1 \\
 & Pedestrian Counts & -48 & 48 & 1 \\
 & Taxi (30 Min.) & -48 & 48 & 1 \\
 & Uber TLC (Hourly) & -24 & 24 & 1 \\
 & Uber TLC (Daily) & -7 & 7 & 1 \\
 & Rideshare & -24 & 24 & 1 \\
 & Temperature-Rain & -30 & 30 & 1 \\
 & London Smart Meters & -48 & 48 & 1 \\
\cline{1-5}
\multirow[t]{27}{*}{Zero-Shot} & Australian Electricity & -48 & 48 & 1 \\
 & Car Parts & -12 & 12 & 1 \\
 & CIF 2016 & -12 & 12 & 1 \\
 & Covid Deaths & -30 & 30 & 1 \\
 & Dominick & -8 & 8 & 1 \\
 & ERCOT Load & -24 & 24 & 1 \\
 & ETTm & -96 & 24 & 1 \\
 & ETTh & -24 & 24 & 1 \\
 & Exchange Rate & -30 & 30 & 1 \\
 & FRED-MD & -12 & 12 & 1 \\
 & Hospital & -12 & 12 & 1 \\
 & M1 (Quarterly) & -8 & 8 & 1 \\
 & M1 (Monthly) & -18 & 18 & 1 \\
 & M1 (Yearly) & -6 & 6 & 1 \\
 & M3 (Monthly) & -18 & 18 & 1 \\
 & M3 (Quarterly) & -8 & 8 & 1 \\
 & M3 (Yearly) & -6 & 6 & 1 \\
 & M4 (Quarterly) & -8 & 8 & 1 \\
 & M4 (Yearly) & -6 & 6 & 1 \\
 & M5 & -28 & 28 & 1 \\
 & NN5 (Daily) & -56 & 56 & 1 \\
 & NN5 (Weekly) & -8 & 8 & 1 \\
 & Tourism (Monthly) & -24 & 24 & 1 \\
 & Tourism (Quarterly) & -8 & 8 & 1 \\
 & Tourism (Yearly) & -4 & 4 & 1 \\
 & Traffic & -24 & 24 & 1 \\
 & Weather & -30 & 30 & 1 \\
\bottomrule
\end{tabular}

\end{small}
\label{tab:dataset-uni}
\end{table}

%% file: tables/hyperparam_baselines.tex
\begin{table}[h]
\centering
\caption{Hyperparameter choices of task-specific and local models. Unspecified hyperparameters are set to the defaults of their respective implementations. Symbols: $C$ - context length, $d_h$ - hidden layer dimension, $n_L$ - number of layers, $n_H$ - number of heads.}
\label{tab:hyperparam-baselines}
\begin{tabular}{l l l l l}
\toprule
Model & Model Type & Implementation  & Hyperparameters \\
\midrule
DeepAR         & Task-specific       & GluonTS                 & $d_h = 40, n_{L} = 2$\\
TFT            & Task-specific       & GluonTS                 & $d_h = 32, n_{H} = 4$\\
PatchTST       & Task-specific       & GluonTS                 & Patch length: 16, Stride: 8, $d_h = 32, n_L = 2, n_{H} = 4$\\
DLinear        & Task-specific       & GluonTS                 & Kernel size: 25, $d_h = 20$\\
N-BEATS        & Task-specific       & NeuralForecast          & Input size multiplier: 5, Stack Type: Generalized\\
N-HITS         & Task-specific       & NeuralForecast          & Input size multiplier: 5\\
AutoETS        & Local               & StatsForecast                    & $C = 512$\\
AutoARIMA      & Local               & StatsForecast                    & $C = 512$\\
AutoTheta      & Local               & StatsForecast                    & $C = 512$\\
SeasonalNaive  & Local               & StatsForecast                    & - \\
\bottomrule
\end{tabular}
\end{table}

%% file: tables/individual-results-zs-mase.tex
\begin{table}[h]
\centering
\caption{\nextcaption}
\begin{small}
\begin{tabular}{llllllllllll}
\toprule
Dataset & \multicolumn{2}{l}{Electricity} & \multicolumn{2}{l}{Hermes} & \multicolumn{2}{l}{ETTh} & \multicolumn{2}{l}{ETTm} & \multicolumn{2}{l}{ProEnfo} & Dalia \\
Horizon & 24 & 48 & 28 & 56 & 24 & 48 & 24 & 48 & 24 & 48 & 60 \\
 &  &  &  &  &  &  &  &  &  &  &  \\
\midrule
COSMIC (Mini) & 0.746 & 0.850 & 0.817 & 0.851 & \textbf{0.636} & \underline{0.790} & 0.385 & 0.614 & 0.737 & \underline{0.906} & 0.439 \\
COSMIC (Small) & 0.737 & \underline{0.849} & \textbf{0.802} & 0.856 & \underline{0.641} & \textbf{0.769} & 0.384 & 0.612 & \underline{0.736} & 0.909 & \underline{0.438} \\
COSMIC (Base) & \textbf{0.729} & \textbf{0.835} & 0.808 & \underline{0.837} & 0.658 & 0.802 & \underline{0.380} & \textbf{0.597} & \textbf{0.723} & \textbf{0.898} & 0.439 \\
Chronos (Mini) & 0.799 & 0.941 & 0.902 & 0.958 & 0.712 & 0.915 & 0.399 & 0.650 & 0.803 & 1.00 & 0.487 \\
Chronos (Small) & 0.780 & 0.929 & 0.876 & 0.924 & 0.690 & 0.849 & 0.383 & 0.615 & 0.768 & 0.965 & 0.459 \\
Chronos (Base) & 0.771 & 0.918 & 0.854 & 0.903 & 0.700 & 0.834 & \textbf{0.377} & \underline{0.611} & \underline{0.736} & 0.938 & 0.465 \\
C. Bolt (Small 512) & 0.750 & 0.883 & \textbf{0.802} & \underline{0.837} & 0.647 & 0.794 & 0.388 & 0.631 & 0.753 & 0.930 & \underline{0.438} \\
C. Bolt (Base 512) & \underline{0.731} & 0.869 & \underline{0.804} & \textbf{0.834} & 0.659 & 0.824 & 0.384 & 0.615 & 0.742 & 0.917 & \textbf{0.437} \\
Moirai-1.0 (Base) & 0.876 & 1.00 & 1.42 & 1.45 & 0.727 & 0.954 & 0.548 & 0.942 & \cellcolor{gray!30}{0.851} & \cellcolor{gray!30}{1.03} & 0.457 \\
Moirai-1.0 (Large) & 0.833 & 0.949 & 1.33 & 1.38 & 0.705 & 0.916 & 0.484 & 0.844 & \cellcolor{gray!30}{0.783} & \cellcolor{gray!30}{0.947} & 0.455 \\
Moirai-1.0 (Small) & 0.944 & 1.07 & 1.43 & 1.38 & 0.720 & 0.922 & 0.549 & 0.939 & \cellcolor{gray!30}{0.956} & \cellcolor{gray!30}{1.11} & 0.458 \\
Moirai-1.1 (Base) & 0.854 & 0.995 & 1.51 & 1.38 & 0.716 & 0.975 & 0.512 & 0.868 & \cellcolor{gray!30}{0.810} & \cellcolor{gray!30}{0.982} & 0.459 \\
Moirai-1.1 (Large) & 0.858 & 0.991 & 1.26 & 1.26 & 0.775 & 1.00 & 0.476 & 0.813 & \cellcolor{gray!30}{0.762} & \cellcolor{gray!30}{0.894} & 0.464 \\
Moirai-1.1 (Small) & 0.945 & 1.06 & 1.39 & 1.42 & 0.730 & 0.895 & 0.529 & 0.845 & \cellcolor{gray!30}{0.926} & \cellcolor{gray!30}{1.08} & 0.457 \\
TTM & 0.925 & 1.06 & 1.83 & 1.86 & 0.737 & 0.896 & 0.461 & 0.722 & 0.885 & 1.03 & 0.443 \\
TimesFM & 0.782 & 0.907 & 0.891 & 0.959 & 0.704 & 0.903 & 0.459 & 0.710 & 0.793 & 0.958 & \textbf{0.437} \\
Seasonal Naive & 1.06 & 1.29 & 1.82 & 1.67 & 0.912 & 1.06 & 0.906 & 0.906 & 1.09 & 1.25 & 0.704 \\
\bottomrule
\end{tabular}

\vspace{0.3cm}

\begin{tabular}{lllllllllll}
\toprule
Dataset & \multicolumn{2}{l}{PEMS04} & \multicolumn{2}{l}{PEMS08} & \multicolumn{2}{l}{ChinaAir} & \multicolumn{2}{l}{CDCIlinet} & \multicolumn{2}{l}{CDCWho} \\
Horizon & 24 & 48 & 24 & 48 & 24 & 48 & 13 & 26 & 13 & 26 \\
 &  &  &  &  &  &  &  &  &  &  \\
\midrule
COSMIC (Mini) & \underline{0.720} & 0.874 & 0.676 & 0.862 & 0.610 & \underline{0.679} & 2.06 & 1.80 & 45.2 & \textbf{22.0} \\
COSMIC (Small) & 0.722 & \underline{0.862} & 0.680 & \textbf{0.824} & \textbf{0.607} & 0.738 & 1.94 & 1.92 & 45.3 & 22.1 \\
COSMIC (Base) & \textbf{0.708} & \textbf{0.856} & \textbf{0.668} & \underline{0.825} & \underline{0.608} & \textbf{0.669} & 2.08 & 1.83 & 45.3 & 22.1 \\
Chronos (Mini) & 0.871 & 1.19 & 0.839 & 1.19 & 0.672 & 0.783 & 1.89 & 2.05 & 45.3 & 22.1 \\
Chronos (Small) & 0.855 & 1.15 & 0.833 & 1.14 & 0.668 & 0.773 & 1.83 & 1.86 & \underline{45.2} & 22.1 \\
Chronos (Base) & 0.809 & 1.08 & 0.784 & 1.08 & 0.652 & 0.759 & 1.92 & \textbf{1.63} & 45.2 & 22.1 \\
C. Bolt (Small 512) & 0.728 & 0.898 & 0.681 & 0.861 & 0.643 & 0.738 & 1.87 & \underline{1.74} & \textbf{45.1} & \underline{22.1} \\
C. Bolt (Base 512) & 0.723 & 0.895 & \underline{0.673} & 0.868 & 0.643 & 0.737 & \textbf{1.77} & 1.77 & 45.2 & 22.1 \\
Moirai-1.0 (Base) & \cellcolor{gray!30}{0.713} & \cellcolor{gray!30}{0.832} & \cellcolor{gray!30}{0.672} & \cellcolor{gray!30}{0.801} & \cellcolor{gray!30}{0.694} & \cellcolor{gray!30}{0.781} & \cellcolor{gray!30}{2.29} & \cellcolor{gray!30}{3.05} & \cellcolor{gray!30}{45.5} & \cellcolor{gray!30}{23.6} \\
Moirai-1.0 (Large) & \cellcolor{gray!30}{0.727} & \cellcolor{gray!30}{0.842} & \cellcolor{gray!30}{0.693} & \cellcolor{gray!30}{0.831} & \cellcolor{gray!30}{0.683} & \cellcolor{gray!30}{0.778} & \cellcolor{gray!30}{2.30} & \cellcolor{gray!30}{3.11} & \cellcolor{gray!30}{45.1} & \cellcolor{gray!30}{23.1} \\
Moirai-1.0 (Small) & \cellcolor{gray!30}{0.678} & \cellcolor{gray!30}{0.768} & \cellcolor{gray!30}{0.625} & \cellcolor{gray!30}{0.718} & \cellcolor{gray!30}{0.711} & \cellcolor{gray!30}{0.809} & \cellcolor{gray!30}{2.01} & \cellcolor{gray!30}{3.59} & \cellcolor{gray!30}{45.4} & \cellcolor{gray!30}{23.1} \\
Moirai-1.1 (Base) & \cellcolor{gray!30}{0.709} & \cellcolor{gray!30}{0.856} & \cellcolor{gray!30}{0.667} & \cellcolor{gray!30}{0.827} & \cellcolor{gray!30}{0.718} & \cellcolor{gray!30}{0.809} & \cellcolor{gray!30}{1.96} & \cellcolor{gray!30}{2.29} & \cellcolor{gray!30}{45.3} & \cellcolor{gray!30}{22.4} \\
Moirai-1.1 (Large) & \cellcolor{gray!30}{0.708} & \cellcolor{gray!30}{0.854} & \cellcolor{gray!30}{0.674} & \cellcolor{gray!30}{0.841} & \cellcolor{gray!30}{0.723} & \cellcolor{gray!30}{0.798} & \cellcolor{gray!30}{2.14} & \cellcolor{gray!30}{2.16} & \cellcolor{gray!30}{45.3} & \cellcolor{gray!30}{22.4} \\
Moirai-1.1 (Small) & \cellcolor{gray!30}{0.682} & \cellcolor{gray!30}{0.767} & \cellcolor{gray!30}{0.622} & \cellcolor{gray!30}{0.700} & \cellcolor{gray!30}{0.736} & \cellcolor{gray!30}{0.798} & \cellcolor{gray!30}{2.24} & \cellcolor{gray!30}{2.51} & \cellcolor{gray!30}{45.5} & \cellcolor{gray!30}{23.1} \\
TTM & \cellcolor{gray!30}{0.649} & \cellcolor{gray!30}{0.732} & \cellcolor{gray!30}{0.598} & \cellcolor{gray!30}{0.674} & 0.657 & 0.732 & 2.37 & 3.07 & 45.4 & 24.5 \\
TimesFM & 0.974 & 1.32 & 0.929 & 1.28 & 0.686 & 0.788 & \underline{1.83} & 2.40 & 45.6 & 25.6 \\
Seasonal Naive & 1.06 & 1.06 & 1.07 & 1.07 & 0.837 & 0.885 & 2.22 & 2.29 & 45.4 & 22.6 \\
\bottomrule
\end{tabular}

\end{small}
\label{tab:individual-results-zs-mase}
\end{table}

%% file: tables/individual-results-zs-wql.tex
\begin{table}[h]
\centering
\caption{\nextcaption}
\begin{small}
\begin{tabular}{llllllllllll}
\toprule
Dataset & \multicolumn{2}{l}{Electricity} & \multicolumn{2}{l}{Hermes} & \multicolumn{2}{l}{ETTh} & \multicolumn{2}{l}{ETTm} & \multicolumn{2}{l}{ProEnfo} & Dalia \\
Horizon & 24 & 48 & 28 & 56 & 24 & 48 & 24 & 48 & 24 & 48 & 60 \\
 &  &  &  &  &  &  &  &  &  &  &  \\
\midrule
COSMIC (Mini) & 0.103 & \underline{0.115} & 0.100 & 0.109 & \textbf{0.063} & 0.078 & \underline{0.039} & 0.064 & \underline{0.068} & \textbf{0.085} & 0.840 \\
COSMIC (Small) & \underline{0.102} & \underline{0.115} & 0.098 & 0.108 & \underline{0.064} & \textbf{0.076} & \underline{0.039} & \underline{0.063} & 0.069 & \underline{0.086} & \textbf{0.833} \\
COSMIC (Base) & \textbf{0.101} & \textbf{0.114} & \textbf{0.094} & \textbf{0.102} & \underline{0.064} & \underline{0.077} & \underline{0.039} & \textbf{0.062} & \textbf{0.067} & \underline{0.086} & \underline{0.835} \\
Chronos (Mini) & 0.114 & 0.132 & 0.112 & 0.121 & 0.073 & 0.092 & 0.041 & 0.068 & 0.075 & 0.096 & 1.03 \\
Chronos (Small) & 0.113 & 0.133 & 0.108 & 0.117 & 0.070 & 0.085 & 0.040 & 0.065 & 0.073 & 0.093 & 0.981 \\
Chronos (Base) & 0.110 & 0.129 & 0.105 & 0.113 & 0.070 & 0.082 & \textbf{0.038} & \textbf{0.062} & 0.070 & 0.090 & 1.01 \\
C. Bolt (Small 512) & 0.104 & 0.121 & \underline{0.096} & \underline{0.103} & \underline{0.064} & 0.078 & \underline{0.039} & 0.066 & 0.069 & 0.087 & 0.840 \\
C. Bolt (Base 512) & \underline{0.102} & 0.118 & \underline{0.096} & \textbf{0.102} & 0.065 & 0.081 & \underline{0.039} & \underline{0.063} & \underline{0.068} & \textbf{0.085} & 0.838 \\
Moirai-1.0 (Base) & 0.124 & 0.138 & 0.188 & 0.205 & 0.073 & 0.095 & 0.058 & 0.101 & \cellcolor{gray!30}{0.076} & \cellcolor{gray!30}{0.093} & 0.870 \\
Moirai-1.0 (Large) & 0.118 & 0.129 & 0.177 & 0.194 & 0.072 & 0.091 & 0.050 & 0.088 & \cellcolor{gray!30}{0.072} & \cellcolor{gray!30}{0.087} & 0.869 \\
Moirai-1.0 (Small) & 0.133 & 0.150 & 0.186 & 0.193 & 0.073 & 0.094 & 0.057 & 0.100 & \cellcolor{gray!30}{0.084} & \cellcolor{gray!30}{0.099} & 0.878 \\
Moirai-1.1 (Base) & 0.122 & 0.138 & 0.199 & 0.194 & 0.073 & 0.099 & 0.054 & 0.092 & \cellcolor{gray!30}{0.073} & \cellcolor{gray!30}{0.091} & 0.879 \\
Moirai-1.1 (Large) & 0.122 & 0.138 & 0.164 & 0.171 & 0.079 & 0.101 & 0.050 & 0.088 & \cellcolor{gray!30}{0.068} & \cellcolor{gray!30}{0.080} & 0.884 \\
Moirai-1.1 (Small) & 0.133 & 0.146 & 0.189 & 0.203 & 0.074 & 0.090 & 0.056 & 0.091 & \cellcolor{gray!30}{0.081} & \cellcolor{gray!30}{0.097} & 0.878 \\
TTM & 0.158 & 0.178 & 0.281 & 0.301 & 0.093 & 0.112 & 0.059 & 0.093 & 0.098 & 0.116 & 1.01 \\
TimesFM & 0.107 & 0.123 & 0.103 & 0.118 & 0.070 & 0.087 & 0.047 & 0.073 & 0.073 & 0.089 & 0.836 \\
Seasonal Naive & 0.186 & 0.222 & 0.313 & 0.291 & 0.113 & 0.131 & 0.114 & 0.114 & 0.122 & 0.138 & 1.65 \\
\bottomrule
\end{tabular}

\vspace{0.3cm}

\begin{tabular}{lllllllllll}
\toprule
Dataset & \multicolumn{2}{l}{PEMS04} & \multicolumn{2}{l}{PEMS08} & \multicolumn{2}{l}{ChinaAir} & \multicolumn{2}{l}{CDCIlinet} & \multicolumn{2}{l}{CDCWho} \\
Horizon & 24 & 48 & 24 & 48 & 24 & 48 & 13 & 26 & 13 & 26 \\
 &  &  &  &  &  &  &  &  &  &  \\
\midrule
COSMIC (Mini) & \underline{0.104} & 0.128 & \underline{0.076} & 0.099 & \textbf{0.226} & \underline{0.246} & 0.152 & 0.111 & 0.884 & 0.842 \\
COSMIC (Small) & \underline{0.104} & \underline{0.126} & \underline{0.076} & \textbf{0.094} & \underline{0.227} & 0.259 & 0.128 & 0.122 & 0.884 & 0.859 \\
COSMIC (Base) & \textbf{0.102} & \textbf{0.125} & \textbf{0.074} & \textbf{0.094} & \textbf{0.226} & \textbf{0.242} & 0.153 & 0.110 & 0.877 & 0.841 \\
Chronos (Mini) & 0.133 & 0.185 & 0.098 & 0.141 & 0.256 & 0.294 & 0.137 & 0.141 & 0.880 & 0.837 \\
Chronos (Small) & 0.131 & 0.181 & 0.098 & 0.137 & 0.253 & 0.286 & 0.137 & 0.128 & 0.887 & 0.831 \\
Chronos (Base) & 0.123 & 0.168 & 0.090 & 0.126 & 0.248 & 0.287 & 0.147 & \textbf{0.108} & \underline{0.872} & \textbf{0.822} \\
C. Bolt (Small 512) & 0.105 & 0.132 & \underline{0.076} & \underline{0.097} & 0.241 & 0.272 & 0.125 & 0.110 & \textbf{0.865} & \textbf{0.822} \\
C. Bolt (Base 512) & \underline{0.104} & 0.132 & \underline{0.076} & 0.098 & 0.239 & 0.269 & \underline{0.124} & \underline{0.109} & 0.873 & \underline{0.826} \\
Moirai-1.0 (Base) & \cellcolor{gray!30}{0.102} & \cellcolor{gray!30}{0.121} & \cellcolor{gray!30}{0.074} & \cellcolor{gray!30}{0.089} & \cellcolor{gray!30}{0.256} & \cellcolor{gray!30}{0.279} & \cellcolor{gray!30}{0.197} & \cellcolor{gray!30}{0.251} & \cellcolor{gray!30}{0.895} & \cellcolor{gray!30}{1.13} \\
Moirai-1.0 (Large) & \cellcolor{gray!30}{0.102} & \cellcolor{gray!30}{0.119} & \cellcolor{gray!30}{0.075} & \cellcolor{gray!30}{0.089} & \cellcolor{gray!30}{0.256} & \cellcolor{gray!30}{0.285} & \cellcolor{gray!30}{0.179} & \cellcolor{gray!30}{0.225} & \cellcolor{gray!30}{0.875} & \cellcolor{gray!30}{1.19} \\
Moirai-1.0 (Small) & \cellcolor{gray!30}{0.098} & \cellcolor{gray!30}{0.112} & \cellcolor{gray!30}{0.070} & \cellcolor{gray!30}{0.080} & \cellcolor{gray!30}{0.266} & \cellcolor{gray!30}{0.292} & \cellcolor{gray!30}{0.146} & \cellcolor{gray!30}{0.284} & \cellcolor{gray!30}{0.888} & \cellcolor{gray!30}{0.928} \\
Moirai-1.1 (Base) & \cellcolor{gray!30}{0.102} & \cellcolor{gray!30}{0.126} & \cellcolor{gray!30}{0.075} & \cellcolor{gray!30}{0.095} & \cellcolor{gray!30}{0.268} & \cellcolor{gray!30}{0.294} & \cellcolor{gray!30}{0.142} & \cellcolor{gray!30}{0.152} & \cellcolor{gray!30}{0.880} & \cellcolor{gray!30}{0.850} \\
Moirai-1.1 (Large) & \cellcolor{gray!30}{0.103} & \cellcolor{gray!30}{0.127} & \cellcolor{gray!30}{0.076} & \cellcolor{gray!30}{0.097} & \cellcolor{gray!30}{0.277} & \cellcolor{gray!30}{0.298} & \cellcolor{gray!30}{0.171} & \cellcolor{gray!30}{0.159} & \cellcolor{gray!30}{0.882} & \cellcolor{gray!30}{0.845} \\
Moirai-1.1 (Small) & \cellcolor{gray!30}{0.100} & \cellcolor{gray!30}{0.113} & \cellcolor{gray!30}{0.070} & \cellcolor{gray!30}{0.081} & \cellcolor{gray!30}{0.274} & \cellcolor{gray!30}{0.290} & \cellcolor{gray!30}{0.178} & \cellcolor{gray!30}{0.181} & \cellcolor{gray!30}{0.899} & \cellcolor{gray!30}{1.02} \\
TTM & \cellcolor{gray!30}{0.116} & \cellcolor{gray!30}{0.131} & \cellcolor{gray!30}{0.082} & \cellcolor{gray!30}{0.093} & 0.301 & 0.330 & 0.217 & 0.324 & 1.05 & 1.47 \\
TimesFM & 0.137 & 0.191 & 0.104 & 0.147 & 0.253 & 0.284 & \textbf{0.113} & 0.155 & 0.899 & 0.949 \\
Seasonal Naive & 0.188 & 0.188 & 0.152 & 0.152 & 0.394 & 0.409 & 0.233 & 0.247 & 0.893 & 0.891 \\
\bottomrule
\end{tabular}

\end{small}
\label{tab:individual-results-zs-wql}
\end{table}

%% file: tables/individual-results-taskspecific-mase.tex
\begin{table}[h]
\centering
\caption{\nextcaption}
\begin{small}
\begin{tabular}{llllllllllll}
\toprule
Dataset & \multicolumn{2}{l}{Electricity} & \multicolumn{2}{l}{Hermes} & \multicolumn{2}{l}{ETTh} & \multicolumn{2}{l}{ETTm} & \multicolumn{2}{l}{ProEnfo} & Dalia \\
Horizon & 24 & 48 & 28 & 56 & 24 & 48 & 24 & 48 & 24 & 48 & 60 \\
 &  &  &  &  &  &  &  &  &  &  &  \\
\midrule
COSMIC (Base) & 0.729 & 0.835 & 0.808 & \textbf{0.837} & \textbf{0.658} & \textbf{0.802} & 0.380 & 0.597 & 0.723 & 0.898 & 0.439 \\
DeepAR & 0.960 & 0.957 & \underline{0.791} & 0.951 & 0.693 & 0.900 & 0.826 & 1.54 & 0.904 & 1.07 & 0.439 \\
DeepAR-(C) & 0.899 & 0.857 & \textbf{0.783} & 1.10 & 0.713 & 1.08 & 0.751 & 1.20 & 0.696 & 0.819 & 0.438 \\
PatchTST & 0.768 & 0.980 & 0.824 & 1.02 & 0.692 & 0.924 & 0.362 & 0.610 & 0.804 & 1.02 & 0.437 \\
PatchTST-(C) & \textbf{0.662} & \underline{0.789} & 0.852 & 1.03 & 0.676 & 0.915 & \textbf{0.360} & 0.597 & \textbf{0.617} & \textbf{0.704} & 0.438 \\
TFT & 0.804 & 0.953 & 0.948 & 0.860 & 0.693 & 0.927 & 0.365 & \textbf{0.559} & 0.831 & 1.02 & \underline{0.436} \\
TFT-(C) & 0.711 & 0.871 & 0.938 & \underline{0.838} & 0.708 & \underline{0.895} & \underline{0.361} & \underline{0.593} & \underline{0.679} & \underline{0.720} & \textbf{0.435} \\
N-BEATS & 0.803 & 0.919 & 0.969 & 0.987 & 0.852 & 0.952 & 0.362 & 0.606 & 0.800 & 0.980 & 0.438 \\
N-BEATSx (C) & \underline{0.674} & \textbf{0.773} & 1.32 & 1.34 & 0.750 & 1.02 & 0.398 & 0.609 & 0.830 & 0.957 & 0.689 \\
DLinear & 0.974 & 1.05 & 0.912 & 1.10 & \underline{0.673} & 0.909 & 0.407 & 0.710 & 0.838 & 1.02 & 0.437 \\
N-HITS & 0.816 & 0.935 & 1.19 & 0.998 & 0.771 & 1.02 & 0.367 & 0.603 & 0.818 & 0.999 & 0.438 \\
AutoTheta & 1.06 & 1.21 & 1.69 & 1.60 & 0.733 & 0.916 & 0.525 & 0.954 & 0.997 & 1.21 & 0.447 \\
AutoARIMA & 1.06 & 1.45 & 2.38 & 1.23 & 0.912 & 1.26 & 1.05 & 1.44 & 1.09 & 1.40 & 0.668 \\
AutoARIMAX & 0.762 & 0.853 & 1.64 & 1.69 & 0.831 & 1.03 & 0.490 & 0.774 & 0.826 & 0.982 & 323 \\
AutoETS & 1.46 & 1.64 & 1.95 & 1.82 & 1.02 & 1.12 & 0.551 & 0.790 & 1.48 & 1.64 & \underline{0.436} \\
Seasonal Naive & 1.06 & 1.29 & 1.82 & 1.67 & 0.912 & 1.06 & 0.906 & 0.906 & 1.09 & 1.25 & 0.704 \\
\bottomrule
\end{tabular}

\vspace{0.3cm}

\begin{tabular}{lllllllllll}
\toprule
Dataset & \multicolumn{2}{l}{PEMS04} & \multicolumn{2}{l}{PEMS08} & \multicolumn{2}{l}{ChinaAir} & \multicolumn{2}{l}{CDCIlinet} & \multicolumn{2}{l}{CDCWho} \\
Horizon & 24 & 48 & 24 & 48 & 24 & 48 & 13 & 26 & 13 & 26 \\
 &  &  &  &  &  &  &  &  &  &  \\
\midrule
COSMIC (Base) & 0.708 & 0.856 & 0.668 & 0.825 & 0.608 & 0.669 & 2.08 & 1.83 & 45.3 & 22.1 \\
DeepAR & 0.651 & 0.793 & 0.615 & 0.754 & 0.638 & 0.721 & 1.78 & 1.85 & 45.1 & 22.1 \\
DeepAR-(C) & 0.420 & 0.576 & 0.409 & 0.532 & 0.507 & 0.607 & \underline{0.951} & \textbf{1.11} & 44.8 & \textbf{21.3} \\
PatchTST & 0.648 & 0.624 & 0.552 & 0.553 & 0.630 & 0.690 & 1.58 & 2.06 & 44.8 & 22.3 \\
PatchTST-(C) & \textbf{0.298} & \textbf{0.302} & \textbf{0.270} & \textbf{0.276} & \textbf{0.435} & \textbf{0.487} & 0.958 & 1.44 & \textbf{44.2} & 21.9 \\
TFT & 0.712 & 0.845 & 0.645 & 0.751 & 0.681 & 0.746 & 1.79 & 2.52 & 45.0 & 23.4 \\
TFT-(C) & \underline{0.320} & \underline{0.361} & \underline{0.302} & \underline{0.328} & \underline{0.483} & \underline{0.515} & \textbf{0.828} & \underline{1.34} & \underline{44.7} & \underline{21.6} \\
N-BEATS & 2.00 & 2.69 & 2.98 & 2.39 & 1.16 & 1.65 & 22.7 & 25.7 & 46.2 & 25.6 \\
N-BEATSx (C) & nan & nan & nan & nan & nan & nan & nan & nan & nan & nan \\
DLinear & 0.968 & 0.828 & 0.781 & 0.819 & 0.666 & 0.729 & 1.74 & 2.12 & 45.1 & 22.8 \\
N-HITS & 2.00 & 2.67 & 3.01 & 2.38 & 1.16 & 1.64 & 22.6 & 24.7 & 45.8 & 29.2 \\
AutoTheta & 0.983 & 1.41 & 0.967 & 1.45 & 0.827 & 0.953 & 2.36 & 2.28 & 45.9 & 23.6 \\
AutoARIMA & 0.972 & 1.45 & 0.907 & 1.40 & 0.866 & 1.02 & 2.28 & 2.44 & 45.4 & 23.2 \\
AutoARIMAX & 0.592 & 0.766 & 0.565 & 0.788 & 0.827 & 0.953 & 1.25 & 2.07 & 89.0 & 43.8 \\
AutoETS & 0.955 & 1.43 & 0.895 & 1.38 & 0.827 & 0.953 & 2.32 & 2.29 & 45.6 & 23.2 \\
Seasonal Naive & 1.06 & 1.06 & 1.07 & 1.07 & 0.837 & 0.885 & 2.22 & 2.29 & 45.4 & 22.6 \\
\bottomrule
\end{tabular}

\end{small}
\label{tab:individual-results-taskspecific-mase}
\end{table}

%% file: tables/individual-results-taskspecific-wql.tex
\begin{table}[h]
\centering
\caption{\nextcaption}
\begin{small}
\begin{tabular}{llllllllllll}
\toprule
Dataset & \multicolumn{2}{l}{Electricity} & \multicolumn{2}{l}{Hermes} & \multicolumn{2}{l}{ETTh} & \multicolumn{2}{l}{ETTm} & \multicolumn{2}{l}{ProEnfo} & Dalia \\
Horizon & 24 & 48 & 28 & 56 & 24 & 48 & 24 & 48 & 24 & 48 & 60 \\
 &  &  &  &  &  &  &  &  &  &  &  \\
\midrule
COSMIC (Base) & \underline{0.101} & \underline{0.114} & \textbf{0.094} & \textbf{0.102} & \textbf{0.064} & \textbf{0.077} & 0.039 & 0.062 & 0.067 & 0.086 & \underline{0.835} \\
DeepAR & 0.119 & 0.129 & \underline{0.095} & 0.132 & 0.069 & \underline{0.089} & 0.090 & 0.172 & 0.079 & 0.098 & 0.838 \\
DeepAR-(C) & 0.109 & 0.117 & \textbf{0.094} & 0.164 & 0.071 & 0.109 & 0.079 & 0.133 & 0.067 & 0.082 & 0.838 \\
PatchTST & 0.111 & 0.131 & 0.099 & 0.132 & \underline{0.068} & 0.092 & \textbf{0.036} & 0.061 & 0.075 & 0.096 & 0.842 \\
PatchTST-(C) & \textbf{0.096} & \textbf{0.108} & 0.107 & 0.137 & \underline{0.068} & 0.094 & \textbf{0.036} & \underline{0.060} & \textbf{0.060} & \underline{0.070} & 0.850 \\
TFT & 0.114 & 0.136 & 0.120 & 0.106 & 0.070 & 0.093 & \underline{0.037} & \textbf{0.057} & 0.077 & 0.092 & \underline{0.835} \\
TFT-(C) & 0.102 & 0.124 & 0.116 & \underline{0.103} & 0.074 & \underline{0.089} & \underline{0.037} & \underline{0.060} & \underline{0.064} & \textbf{0.069} & \textbf{0.831} \\
N-BEATS & 0.140 & 0.159 & 0.136 & 0.150 & 0.106 & 0.119 & 0.045 & 0.077 & 0.091 & 0.114 & 1.00 \\
N-BEATSx (C) & 0.115 & 0.131 & 0.179 & 0.192 & 0.092 & 0.124 & 0.050 & 0.077 & 0.100 & 0.116 & 1.56 \\
DLinear & 0.131 & 0.142 & 0.116 & 0.152 & 0.070 & 0.090 & 0.043 & 0.073 & 0.077 & 0.092 & 0.841 \\
N-HITS & 0.142 & 0.161 & 0.144 & 0.148 & 0.096 & 0.128 & 0.046 & 0.076 & 0.093 & 0.116 & 1.00 \\
AutoTheta & 0.174 & 0.217 & 0.225 & 0.244 & 0.073 & 0.090 & 0.058 & 0.109 & 0.087 & 0.109 & 0.902 \\
AutoARIMA & 0.166 & 0.218 & 0.310 & 0.165 & 0.090 & 0.122 & 0.108 & 0.145 & 0.101 & 0.125 & 1.30 \\
AutoARIMAX & 0.110 & 0.122 & 0.223 & 0.242 & 0.085 & 0.110 & 0.053 & 0.082 & 0.066 & 0.077 & 619 \\
AutoETS & 0.210 & 0.260 & 0.263 & 0.297 & 0.102 & 0.113 & 0.061 & 0.085 & 0.112 & 0.131 & 0.867 \\
Seasonal Naive & 0.186 & 0.222 & 0.313 & 0.291 & 0.113 & 0.131 & 0.114 & 0.114 & 0.122 & 0.138 & 1.65 \\
\bottomrule
\end{tabular}

\vspace{0.3cm}

\begin{tabular}{lllllllllll}
\toprule
Dataset & \multicolumn{2}{l}{PEMS04} & \multicolumn{2}{l}{PEMS08} & \multicolumn{2}{l}{ChinaAir} & \multicolumn{2}{l}{CDCIlinet} & \multicolumn{2}{l}{CDCWho} \\
Horizon & 24 & 48 & 24 & 48 & 24 & 48 & 13 & 26 & 13 & 26 \\
 &  &  &  &  &  &  &  &  &  &  \\
\midrule
COSMIC (Base) & 0.102 & 0.125 & 0.074 & 0.094 & 0.226 & 0.242 & 0.153 & 0.110 & 0.877 & 0.841 \\
DeepAR & 0.089 & 0.109 & 0.065 & 0.076 & 0.240 & 0.264 & 0.144 & 0.139 & 0.878 & \underline{0.827} \\
DeepAR-(C) & 0.052 & 0.076 & 0.038 & 0.050 & 0.181 & 0.208 & 0.040 & \underline{0.055} & 0.870 & \textbf{0.800} \\
PatchTST & 0.090 & 0.089 & 0.062 & 0.062 & 0.233 & 0.253 & 0.102 & 0.130 & 0.873 & 0.876 \\
PatchTST-(C) & \textbf{0.039} & \textbf{0.039} & \textbf{0.028} & \textbf{0.029} & \textbf{0.157} & \textbf{0.173} & \underline{0.035} & \textbf{0.052} & 0.919 & 0.844 \\
TFT & 0.102 & 0.121 & 0.070 & 0.085 & 0.260 & 0.272 & 0.135 & 0.162 & \underline{0.867} & 0.908 \\
TFT-(C) & \underline{0.040} & \underline{0.044} & \underline{0.029} & \underline{0.032} & \underline{0.179} & \underline{0.181} & \textbf{0.029} & 0.082 & \textbf{0.860} & 1.03 \\
N-BEATS & 0.474 & 0.562 & 0.469 & 0.403 & 0.540 & 0.654 & 0.777 & 0.870 & 0.997 & 1.02 \\
N-BEATSx (C) & nan & nan & nan & nan & nan & nan & nan & nan & nan & nan \\
DLinear & 0.125 & 0.121 & 0.089 & 0.094 & 0.249 & 0.270 & 0.109 & 0.137 & 0.898 & 1.25 \\
N-HITS & 0.473 & 0.569 & 0.474 & 0.402 & 0.541 & 0.657 & 0.772 & 0.828 & 0.982 & 1.03 \\
AutoTheta & 0.158 & 0.228 & 0.118 & 0.178 & 0.302 & 0.330 & 0.202 & 0.226 & 1.19 & 1.69 \\
AutoARIMA & 0.155 & 0.230 & 0.112 & 0.170 & 0.352 & 0.417 & 0.205 & 0.228 & 1.19 & 1.59 \\
AutoARIMAX & 0.086 & 0.111 & 0.063 & 0.086 & 0.305 & 0.333 & 0.081 & 0.135 & 3.37 & 3.54 \\
AutoETS & 0.148 & 0.233 & 0.109 & 0.178 & 0.302 & 0.330 & 0.203 & 0.231 & 1.16 & 1.59 \\
Seasonal Naive & 0.188 & 0.188 & 0.152 & 0.152 & 0.394 & 0.409 & 0.233 & 0.247 & 0.893 & 0.891 \\
\bottomrule
\end{tabular}

\end{small}
\label{tab:individual-results-taskspecific-wql}
\end{table}

%% file: tables/uni-zeroshot-mase.tex
\begin{table}[h]
\centering
\caption{\nextcaption}
{\scriptsize
\begin{tabular}{llllllllllllllll}
\toprule
  & \rotatebox{90}{ Australian Electricity } & \rotatebox{90}{ Car Parts } & \rotatebox{90}{ CIF 2016 } & \rotatebox{90}{ Covid Deaths } & \rotatebox{90}{ Dominick } & \rotatebox{90}{ ERCOT Load } & \rotatebox{90}{ ETTh } & \rotatebox{90}{ ETTm } & \rotatebox{90}{ Exchange Rate } & \rotatebox{90}{ FRED-MD } & \rotatebox{90}{ Hospital } & \rotatebox{90}{ M1 (Monthly) } & \rotatebox{90}{ M1 (Quarterly) } & \rotatebox{90}{ M1 (Yearly) } & \rotatebox{90}{ M3 (Monthly) }\\
\midrule
COSMIC (Mini) & \underline{1.17} & 0.886 & 0.988 & \underline{39.9} & 0.872 & \textbf{0.532} & \underline{0.759} & \underline{0.714} & 1.94 & 0.640 & 0.815 & 1.14 & 1.81 & 4.66 & 0.885 \\
COSMIC (Small) & 1.55 & 0.879 & 1.01 & 40.4 & 0.879 & 0.575 & 0.766 & 0.741 & 1.87 & 0.597 & 0.812 & 1.14 & 1.74 & 4.68 & 0.893 \\
COSMIC (Base) & 1.55 & 0.876 & 0.984 & \textbf{38.2} & 0.868 & 0.616 & 0.760 & 0.721 & 1.84 & 0.634 & 0.808 & 1.13 & \underline{1.72} & \textbf{3.77} & \textbf{0.865} \\
Chronos (Mini) & \textbf{1.11} & 0.891 & 1.05 & 43.6 & 0.833 & 0.588 & 0.797 & 0.792 & 2.03 & \textbf{0.483} & 0.817 & 1.17 & 1.78 & 4.96 & 0.900 \\
Chronos (Small) & 1.40 & 0.887 & 0.989 & 42.7 & \underline{0.819} & 0.573 & 0.789 & \textbf{0.710} & 2.25 & 0.496 & 0.815 & 1.17 & 1.76 & 4.66 & 0.885 \\
Chronos (Base) & 1.32 & 0.899 & \underline{0.981} & 42.7 & \textbf{0.816} & \underline{0.550} & 0.789 & 0.739 & 2.43 & 0.486 & 0.810 & \underline{1.12} & 1.74 & 4.62 & \underline{0.868} \\
C. Bolt (Small 512) & 1.26 & \underline{0.864} & 1.04 & 42.2 & 0.884 & 0.580 & 0.761 & 0.796 & 1.84 & 0.616 & 0.806 & 1.16 & 1.81 & 5.61 & 0.885 \\
C. Bolt (Base 512) & 1.25 & \textbf{0.862} & 1.04 & 42.4 & 0.877 & 0.699 & \textbf{0.756} & 0.738 & 1.84 & 0.583 & \underline{0.799} & 1.15 & 1.74 & 5.20 & 0.885 \\
Moirai-1.0 (Base) & \cellcolor{gray!30}{1.26} & \cellcolor{gray!30}{1.73} & \cellcolor{gray!30}{1.20} & \cellcolor{gray!30}{33.1} & 0.879 & 0.583 & 0.902 & 0.981 & \textbf{1.51} & \cellcolor{gray!30}{0.607} & \cellcolor{gray!30}{0.821} & \cellcolor{gray!30}{1.27} & \cellcolor{gray!30}{1.90} & \cellcolor{gray!30}{4.62} & \cellcolor{gray!30}{0.946} \\
Moirai-1.0 (Large) & \cellcolor{gray!30}{1.01} & \cellcolor{gray!30}{1.54} & \cellcolor{gray!30}{1.16} & \cellcolor{gray!30}{33.1} & 0.845 & 0.667 & 0.845 & 0.753 & 1.91 & \cellcolor{gray!30}{0.593} & \cellcolor{gray!30}{0.826} & \cellcolor{gray!30}{1.24} & \cellcolor{gray!30}{1.84} & \cellcolor{gray!30}{4.71} & \cellcolor{gray!30}{0.924} \\
Moirai-1.1 (Large) & \cellcolor{gray!30}{0.930} & \cellcolor{gray!30}{0.846} & \cellcolor{gray!30}{1.03} & \cellcolor{gray!30}{32.1} & 0.830 & 0.568 & 0.825 & 0.755 & \underline{1.74} & \cellcolor{gray!30}{0.564} & \cellcolor{gray!30}{0.830} & \cellcolor{gray!30}{1.14} & \cellcolor{gray!30}{1.79} & \cellcolor{gray!30}{3.52} & \cellcolor{gray!30}{0.902} \\
Moirai-1.1 (Base) & \cellcolor{gray!30}{1.25} & \cellcolor{gray!30}{0.863} & \cellcolor{gray!30}{1.06} & \cellcolor{gray!30}{34.3} & 0.828 & 0.569 & 0.902 & 0.826 & 1.81 & \cellcolor{gray!30}{0.572} & \cellcolor{gray!30}{0.820} & \cellcolor{gray!30}{1.14} & \cellcolor{gray!30}{1.64} & \cellcolor{gray!30}{3.61} & \cellcolor{gray!30}{0.897} \\
TimesFM & 1.63 & 0.893 & \textbf{0.925} & 55.6 & 1.22 & 0.590 & 0.890 & 1.04 & 3.31 & \underline{0.484} & \textbf{0.759} & \textbf{1.03} & \textbf{1.63} & \underline{4.00} & 0.870 \\
TTM & \cellcolor{gray!30}{1.40} & 1.57 & 2.02 & 53.5 & 1.27 & 1.04 & 0.985 & 0.915 & 1.79 & 0.659 & 0.912 & 1.58 & 2.18 & 5.70 & 1.51 \\
\bottomrule
\end{tabular}

\vspace{0.3cm}

\begin{tabular}{lllllllllllllll}
\toprule
  & \rotatebox{90}{ M3 (Quarterly) } & \rotatebox{90}{ M3 (Yearly) } & \rotatebox{90}{ M4 (Quarterly) } & \rotatebox{90}{ M4 (Yearly) } & \rotatebox{90}{ M5 } & \rotatebox{90}{ NN5 (Daily) } & \rotatebox{90}{ NN5 (Weekly) } & \rotatebox{90}{ Tourism (Monthly) } & \rotatebox{90}{ Tourism (Quarterly) } & \rotatebox{90}{ Tourism (Yearly) } & \rotatebox{90}{ Traffic } & \rotatebox{90}{ Weather } & \rotatebox{90}{ Agg. Score } & \rotatebox{90}{ Avg. Rank }\\
\midrule
COSMIC (Mini) & 1.29 & 3.18 & 1.25 & 3.77 & 0.921 & 0.614 & 0.948 & 1.76 & \underline{1.68} & 3.93 & 0.913 & \underline{0.787} & 0.834 & 12.0 \\
COSMIC (Small) & 1.27 & 3.23 & \underline{1.23} & 3.72 & 0.920 & 0.618 & 0.951 & 1.73 & \underline{1.68} & 3.86 & 0.914 & \underline{0.787} & 0.841 & 12.4 \\
COSMIC (Base) & 1.24 & \underline{2.91} & \textbf{1.21} & \textbf{3.48} & 0.920 & 0.610 & \textbf{0.925} & 1.64 & \textbf{1.67} & 3.59 & 0.847 & 0.788 & 0.818 & \underline{9.44} \\
Chronos (Mini) & 1.29 & 3.38 & 1.27 & 3.74 & 0.944 & 0.642 & 0.947 & 1.95 & 1.83 & 4.05 & 0.850 & 0.853 & 0.850 & 14.6 \\
Chronos (Small) & 1.26 & 3.28 & 1.25 & \underline{3.65} & 0.940 & 0.615 & 0.944 & 1.90 & 1.73 & 3.90 & \underline{0.837} & 0.836 & 0.841 & 12.4 \\
Chronos (Base) & \underline{1.20} & 3.21 & \underline{1.23} & 3.68 & 0.939 & \underline{0.585} & \underline{0.938} & 1.83 & 1.72 & 3.90 & \textbf{0.828} & 0.824 & 0.832 & 10.6 \\
C. Bolt (Small 512) & 1.35 & 3.78 & 1.26 & 3.81 & \underline{0.916} & \underline{0.585} & 0.942 & 1.67 & 1.97 & 4.19 & 0.893 & 0.790 & 0.858 & 13.6 \\
C. Bolt (Base 512) & 1.32 & 3.56 & \underline{1.23} & 3.77 & \underline{0.916} & \textbf{0.580} & 0.948 & \underline{1.57} & 1.90 & 4.06 & 0.877 & \textbf{0.786} & 0.847 & 11.9 \\
Moirai-1.0 (Base) & \cellcolor{gray!30}{1.43} & \cellcolor{gray!30}{3.66} & \cellcolor{gray!30}{1.29} & \cellcolor{gray!30}{3.60} & \cellcolor{gray!30}{1.44} & \cellcolor{gray!30}{0.698} & \cellcolor{gray!30}{0.980} & \cellcolor{gray!30}{2.04} & \cellcolor{gray!30}{2.72} & \cellcolor{gray!30}{3.05} & \cellcolor{gray!30}{0.726} & \cellcolor{gray!30}{0.831} & 0.907 & 15.7 \\
Moirai-1.0 (Large) & \cellcolor{gray!30}{1.43} & \cellcolor{gray!30}{3.82} & \cellcolor{gray!30}{1.26} & \cellcolor{gray!30}{4.18} & \cellcolor{gray!30}{0.929} & \cellcolor{gray!30}{0.625} & \cellcolor{gray!30}{1.01} & \cellcolor{gray!30}{1.91} & \cellcolor{gray!30}{2.28} & \cellcolor{gray!30}{3.30} & \cellcolor{gray!30}{0.759} & \cellcolor{gray!30}{0.807} & 0.876 & 15.4 \\
Moirai-1.1 (Large) & \cellcolor{gray!30}{1.12} & \cellcolor{gray!30}{2.75} & \cellcolor{gray!30}{1.17} & \cellcolor{gray!30}{3.04} & \cellcolor{gray!30}{0.927} & \cellcolor{gray!30}{0.586} & \cellcolor{gray!30}{0.979} & \cellcolor{gray!30}{1.65} & \cellcolor{gray!30}{1.82} & \cellcolor{gray!30}{3.21} & \cellcolor{gray!30}{0.758} & \cellcolor{gray!30}{0.810} & \textbf{0.782} & \textbf{8.67} \\
Moirai-1.1 (Base) & \cellcolor{gray!30}{1.15} & \cellcolor{gray!30}{2.73} & \cellcolor{gray!30}{1.17} & \cellcolor{gray!30}{3.07} & \cellcolor{gray!30}{0.924} & \cellcolor{gray!30}{0.638} & \cellcolor{gray!30}{0.939} & \cellcolor{gray!30}{1.75} & \cellcolor{gray!30}{1.92} & \cellcolor{gray!30}{3.15} & \cellcolor{gray!30}{0.768} & \cellcolor{gray!30}{0.823} & \underline{0.805} & 9.48 \\
TimesFM & \textbf{1.15} & \textbf{2.70} & \cellcolor{gray!30}{1.16} & \cellcolor{gray!30}{3.34} & \textbf{0.912} & 0.629 & 0.949 & \textbf{1.54} & 1.73 & \textbf{3.23} & \cellcolor{gray!30}{0.638} & 0.912 & 0.846 & 10.4 \\
TTM & 2.11 & 4.24 & 2.03 & 5.14 & 1.09 & \cellcolor{gray!30}{1.03} & 1.05 & 2.49 & 3.22 & \underline{3.27} & 0.841 & \cellcolor{gray!30}{0.876} & 1.11 & 21.0 \\
\bottomrule
\end{tabular}

}
\label{tab:uni-zeroshot-mase}
\end{table}

%% file: tables/uni-zeroshot-wql.tex
\begin{table}[h]
\centering
\caption{\nextcaption}
{\scriptsize
\begin{tabular}{llllllllllllllll}
\toprule
  & \rotatebox{90}{ Australian Electricity } & \rotatebox{90}{ Car Parts } & \rotatebox{90}{ CIF 2016 } & \rotatebox{90}{ Covid Deaths } & \rotatebox{90}{ Dominick } & \rotatebox{90}{ ERCOT Load } & \rotatebox{90}{ ETTh } & \rotatebox{90}{ ETTm } & \rotatebox{90}{ Exchange Rate } & \rotatebox{90}{ FRED-MD } & \rotatebox{90}{ Hospital } & \rotatebox{90}{ M1 (Monthly) } & \rotatebox{90}{ M1 (Quarterly) } & \rotatebox{90}{ M1 (Yearly) } & \rotatebox{90}{ M3 (Monthly) }\\
\midrule
COSMIC (Mini) & \textbf{0.059} & 1.04 & 0.019 & 0.052 & 0.352 & \textbf{0.016} & \textbf{0.070} & \textbf{0.064} & 0.012 & 0.043 & 0.059 & 0.139 & 0.101 & \textbf{0.140} & 0.096 \\
COSMIC (Small) & 0.079 & 1.03 & 0.018 & 0.051 & 0.352 & 0.018 & \underline{0.071} & \underline{0.067} & 0.012 & 0.052 & 0.060 & 0.131 & 0.099 & \underline{0.150} & 0.097 \\
COSMIC (Base) & 0.080 & 1.02 & 0.016 & \textbf{0.039} & 0.351 & 0.020 & \textbf{0.070} & \underline{0.067} & 0.012 & 0.070 & 0.058 & 0.133 & 0.100 & 0.151 & \underline{0.094} \\
Chronos (Mini) & 0.063 & 1.02 & \textbf{0.013} & 0.084 & 0.346 & 0.018 & 0.085 & 0.072 & 0.012 & \textbf{0.017} & 0.058 & 0.138 & 0.103 & 0.179 & 0.099 \\
Chronos (Small) & 0.074 & 1.03 & \underline{0.015} & 0.059 & \underline{0.338} & 0.018 & 0.080 & \textbf{0.064} & 0.013 & \textbf{0.017} & 0.057 & 0.139 & 0.103 & 0.172 & 0.100 \\
Chronos (Base) & 0.075 & 1.06 & \textbf{0.013} & \underline{0.048} & \textbf{0.333} & \textbf{0.016} & 0.081 & 0.069 & 0.014 & \underline{0.022} & \underline{0.056} & \underline{0.128} & 0.105 & 0.181 & 0.097 \\
C. Bolt (Small 512) & \underline{0.061} & \underline{1.01} & 0.019 & 0.083 & 0.354 & 0.018 & \underline{0.071} & 0.069 & 0.012 & 0.066 & 0.058 & 0.131 & 0.100 & 0.157 & 0.095 \\
C. Bolt (Base 512) & 0.063 & \textbf{1.00} & \underline{0.015} & 0.073 & 0.352 & 0.024 & \underline{0.071} & 0.072 & \underline{0.011} & 0.048 & 0.058 & 0.134 & \underline{0.095} & 0.159 & 0.095 \\
Moirai-1.0 (Base) & \cellcolor{gray!30}{0.055} & \cellcolor{gray!30}{1.65} & \cellcolor{gray!30}{0.010} & \cellcolor{gray!30}{0.038} & 0.361 & 0.019 & 0.096 & 0.075 & \textbf{0.010} & \cellcolor{gray!30}{0.045} & \cellcolor{gray!30}{0.060} & \cellcolor{gray!30}{0.155} & \cellcolor{gray!30}{0.111} & \cellcolor{gray!30}{0.194} & \cellcolor{gray!30}{0.102} \\
Moirai-1.0 (Large) & \cellcolor{gray!30}{0.046} & \cellcolor{gray!30}{1.62} & \cellcolor{gray!30}{0.048} & \cellcolor{gray!30}{0.035} & 0.346 & 0.022 & 0.085 & 0.069 & 0.012 & \cellcolor{gray!30}{0.049} & \cellcolor{gray!30}{0.057} & \cellcolor{gray!30}{0.154} & \cellcolor{gray!30}{0.107} & \cellcolor{gray!30}{0.190} & \cellcolor{gray!30}{0.101} \\
Moirai-1.1 (Large) & \cellcolor{gray!30}{0.038} & \cellcolor{gray!30}{0.990} & \cellcolor{gray!30}{0.015} & \cellcolor{gray!30}{0.036} & 0.344 & \underline{0.017} & 0.082 & 0.070 & \underline{0.011} & \cellcolor{gray!30}{0.042} & \cellcolor{gray!30}{0.059} & \cellcolor{gray!30}{0.177} & \cellcolor{gray!30}{0.093} & \cellcolor{gray!30}{0.127} & \cellcolor{gray!30}{0.099} \\
Moirai-1.1 (Base) & \cellcolor{gray!30}{0.054} & \cellcolor{gray!30}{1.02} & \cellcolor{gray!30}{0.016} & \cellcolor{gray!30}{0.045} & 0.343 & 0.018 & 0.091 & 0.076 & \underline{0.011} & \cellcolor{gray!30}{0.050} & \cellcolor{gray!30}{0.058} & \cellcolor{gray!30}{0.169} & \cellcolor{gray!30}{0.076} & \cellcolor{gray!30}{0.123} & \cellcolor{gray!30}{0.101} \\
TimesFM & 0.089 & 1.02 & 0.020 & 0.204 & 0.426 & 0.021 & 0.092 & 0.084 & 0.013 & 0.035 & \textbf{0.051} & \textbf{0.123} & \textbf{0.087} & 0.163 & \textbf{0.093} \\
\bottomrule
\end{tabular}

\vspace{0.3cm}

\begin{tabular}{lllllllllllllll}
\toprule
  & \rotatebox{90}{ M3 (Quarterly) } & \rotatebox{90}{ M3 (Yearly) } & \rotatebox{90}{ M4 (Quarterly) } & \rotatebox{90}{ M4 (Yearly) } & \rotatebox{90}{ M5 } & \rotatebox{90}{ NN5 (Daily) } & \rotatebox{90}{ NN5 (Weekly) } & \rotatebox{90}{ Tourism (Monthly) } & \rotatebox{90}{ Tourism (Quarterly) } & \rotatebox{90}{ Tourism (Yearly) } & \rotatebox{90}{ Traffic } & \rotatebox{90}{ Weather } & \rotatebox{90}{ Agg. Score } & \rotatebox{90}{ Avg. Rank }\\
\midrule
COSMIC (Mini) & 0.076 & 0.148 & 0.079 & 0.128 & 0.568 & 0.159 & \underline{0.086} & 0.099 & \textbf{0.063} & 0.186 & 0.264 & \underline{0.129} & 0.655 & 10.3 \\
COSMIC (Small) & 0.075 & 0.149 & \underline{0.078} & \underline{0.126} & 0.569 & 0.159 & \underline{0.086} & 0.098 & 0.066 & 0.176 & 0.264 & \textbf{0.128} & 0.667 & 10.5 \\
COSMIC (Base) & \underline{0.074} & \underline{0.140} & \textbf{0.077} & \textbf{0.119} & 0.567 & \underline{0.158} & \textbf{0.085} & 0.096 & \underline{0.065} & \underline{0.162} & \textbf{0.245} & \textbf{0.128} & 0.658 & \underline{9.30} \\
Chronos (Mini) & 0.081 & 0.159 & 0.086 & 0.140 & 0.595 & 0.173 & 0.091 & 0.109 & 0.074 & 0.218 & 0.264 & 0.150 & 0.678 & 15.0 \\
Chronos (Small) & 0.079 & 0.155 & 0.084 & 0.136 & 0.590 & 0.169 & 0.090 & 0.113 & 0.069 & 0.200 & 0.263 & 0.143 & 0.667 & 13.0 \\
Chronos (Base) & 0.076 & 0.153 & 0.083 & 0.137 & 0.586 & 0.161 & 0.091 & 0.103 & 0.069 & 0.207 & 0.264 & 0.140 & 0.662 & 12.3 \\
C. Bolt (Small 512) & 0.078 & 0.162 & 0.079 & 0.131 & 0.564 & \textbf{0.151} & \underline{0.086} & 0.099 & 0.070 & 0.200 & 0.253 & 0.130 & 0.687 & 11.1 \\
C. Bolt (Base 512) & 0.077 & 0.151 & \textbf{0.077} & 0.128 & \underline{0.563} & \textbf{0.151} & \underline{0.086} & \underline{0.095} & 0.067 & 0.179 & \underline{0.250} & \textbf{0.128} & 0.671 & 10.1 \\
Moirai-1.0 (Base) & \cellcolor{gray!30}{0.080} & \cellcolor{gray!30}{0.167} & \cellcolor{gray!30}{0.081} & \cellcolor{gray!30}{0.121} & \cellcolor{gray!30}{0.692} & \cellcolor{gray!30}{0.181} & \cellcolor{gray!30}{0.092} & \cellcolor{gray!30}{0.121} & \cellcolor{gray!30}{0.100} & \cellcolor{gray!30}{0.168} & \cellcolor{gray!30}{0.225} & \cellcolor{gray!30}{0.135} & 0.696 & 15.0 \\
Moirai-1.0 (Large) & \cellcolor{gray!30}{0.085} & \cellcolor{gray!30}{0.170} & \cellcolor{gray!30}{0.080} & \cellcolor{gray!30}{0.138} & \cellcolor{gray!30}{0.584} & \cellcolor{gray!30}{0.162} & \cellcolor{gray!30}{0.093} & \cellcolor{gray!30}{0.111} & \cellcolor{gray!30}{0.085} & \cellcolor{gray!30}{0.161} & \cellcolor{gray!30}{0.231} & \cellcolor{gray!30}{0.132} & 0.720 & 14.5 \\
Moirai-1.1 (Large) & \cellcolor{gray!30}{0.070} & \cellcolor{gray!30}{0.130} & \cellcolor{gray!30}{0.076} & \cellcolor{gray!30}{0.108} & \cellcolor{gray!30}{0.594} & \cellcolor{gray!30}{0.152} & \cellcolor{gray!30}{0.091} & \cellcolor{gray!30}{0.098} & \cellcolor{gray!30}{0.067} & \cellcolor{gray!30}{0.137} & \cellcolor{gray!30}{0.236} & \cellcolor{gray!30}{0.133} & \textbf{0.621} & \textbf{8.26} \\
Moirai-1.1 (Base) & \cellcolor{gray!30}{0.070} & \cellcolor{gray!30}{0.131} & \cellcolor{gray!30}{0.076} & \cellcolor{gray!30}{0.109} & \cellcolor{gray!30}{0.583} & \cellcolor{gray!30}{0.165} & \cellcolor{gray!30}{0.089} & \cellcolor{gray!30}{0.099} & \cellcolor{gray!30}{0.070} & \cellcolor{gray!30}{0.138} & \cellcolor{gray!30}{0.238} & \cellcolor{gray!30}{0.135} & \underline{0.642} & 9.48 \\
TimesFM & \textbf{0.072} & \textbf{0.123} & \cellcolor{gray!30}{0.074} & \cellcolor{gray!30}{0.117} & \textbf{0.559} & 0.160 & \underline{0.086} & \textbf{0.088} & 0.069 & \textbf{0.148} & \cellcolor{gray!30}{0.184} & 0.150 & 0.692 & 9.74 \\
\bottomrule
\end{tabular}

}
\label{tab:uni-zeroshot-wql}
\end{table}

%% file: tables/uni-zeroshot-taskspecific-mase.tex
\begin{table}[h]
\centering
\caption{\nextcaption}
{\scriptsize
\begin{tabular}{llllllllllllllll}
\toprule
  & \rotatebox{90}{ Australian Electricity } & \rotatebox{90}{ Car Parts } & \rotatebox{90}{ CIF 2016 } & \rotatebox{90}{ Covid Deaths } & \rotatebox{90}{ Dominick } & \rotatebox{90}{ ERCOT Load } & \rotatebox{90}{ ETTh } & \rotatebox{90}{ ETTm } & \rotatebox{90}{ Exchange Rate } & \rotatebox{90}{ FRED-MD } & \rotatebox{90}{ Hospital } & \rotatebox{90}{ M1 (Monthly) } & \rotatebox{90}{ M1 (Quarterly) } & \rotatebox{90}{ M1 (Yearly) } & \rotatebox{90}{ M3 (Monthly) }\\
\midrule
COSMIC (Base) & 1.55 & 0.876 & \underline{0.984} & 38.2 & 0.868 & 0.616 & 0.760 & 0.721 & 1.84 & 0.634 & 0.808 & 1.13 & 1.72 & 3.77 & \underline{0.865} \\
N-BEATS & 0.828 & 0.803 & 1.44 & \underline{31.7} & \textbf{0.782} & 0.648 & 0.782 & 0.659 & 2.15 & 0.635 & \textbf{0.760} & 1.24 & 2.04 & 6.21 & 0.883 \\
DeepAR & 1.47 & \textbf{0.798} & 1.36 & 38.2 & 0.851 & 1.20 & 0.814 & 0.874 & 1.61 & 0.621 & 0.804 & 1.12 & 1.74 & \textbf{3.68} & 0.943 \\
DLinear & 1.28 & 0.879 & 1.15 & 40.4 & 0.880 & 0.651 & \textbf{0.695} & 0.724 & \textbf{1.46} & 0.713 & 0.940 & 1.37 & 1.94 & 11.6 & 1.16 \\
N-HiTS & \textbf{0.794} & 0.803 & 1.39 & 31.8 & \textbf{0.782} & \underline{0.615} & 0.811 & \underline{0.643} & 2.04 & 0.696 & 0.781 & 1.33 & 2.06 & 5.57 & 0.899 \\
PatchTST & 0.871 & 0.803 & 1.54 & 36.5 & 0.867 & \textbf{0.553} & \underline{0.729} & 0.652 & \underline{1.54} & 0.745 & 0.859 & 1.21 & 1.92 & 4.04 & 1.22 \\
TFT & \underline{0.810} & \underline{0.799} & 1.55 & \textbf{30.6} & \underline{0.800} & 0.690 & 0.875 & 0.962 & 2.36 & 0.929 & 0.799 & 1.33 & 2.14 & 4.32 & 0.916 \\
AutoTheta & 0.897 & 1.23 & 1.00 & 45.4 & 1.02 & 1.31 & 0.900 & \textbf{0.583} & 1.65 & 0.566 & \underline{0.761} & \underline{1.10} & \textbf{1.68} & \underline{3.70} & \textbf{0.861} \\
AutoETS & 2.39 & 1.18 & \textbf{0.957} & 38.1 & 0.885 & 2.83 & 1.14 & 1.18 & 1.64 & \underline{0.544} & \textbf{0.760} & \textbf{1.07} & \underline{1.71} & 4.11 & 0.869 \\
AutoARIMA & 1.39 & nan & 1.01 & \underline{31.7} & nan & 1.28 & 0.977 & 0.879 & 1.88 & \textbf{0.473} & 0.820 & 1.15 & 1.77 & 3.87 & 0.933 \\
Seasonal Naive & 1.25 & 1.20 & 1.29 & 46.9 & 0.871 & 0.761 & 0.932 & 1.17 & 1.74 & 1.10 & 0.921 & 1.31 & 2.08 & 4.89 & 1.15 \\
Naive & 2.36 & nan & 1.26 & 46.9 & 0.871 & 4.23 & 1.65 & 1.16 & 1.87 & 0.622 & 0.968 & 1.47 & 1.95 & 4.89 & 1.17 \\
\bottomrule
\end{tabular}

\vspace{0.3cm}

\begin{tabular}{lllllllllllllll}
\toprule
  & \rotatebox{90}{ M3 (Quarterly) } & \rotatebox{90}{ M3 (Yearly) } & \rotatebox{90}{ M4 (Quarterly) } & \rotatebox{90}{ M4 (Yearly) } & \rotatebox{90}{ M5 } & \rotatebox{90}{ NN5 (Daily) } & \rotatebox{90}{ NN5 (Weekly) } & \rotatebox{90}{ Tourism (Monthly) } & \rotatebox{90}{ Tourism (Quarterly) } & \rotatebox{90}{ Tourism (Yearly) } & \rotatebox{90}{ Traffic } & \rotatebox{90}{ Weather } & \rotatebox{90}{ Agg. Score } & \rotatebox{90}{ Avg. Rank }\\
\midrule
COSMIC (Base) & 1.24 & 2.91 & 1.21 & 3.48 & 0.920 & 0.610 & 0.925 & 1.64 & 1.67 & 3.59 & 0.847 & \textbf{0.788} & \underline{0.818} & \textbf{9.44} \\
N-BEATS & 1.15 & 3.55 & \textbf{1.13} & nan & \underline{0.917} & \underline{0.571} & 1.01 & \textbf{1.49} & \underline{1.62} & 3.56 & 0.968 & 0.888 & 0.835 & 11.3 \\
DeepAR & 1.21 & 2.83 & 1.25 & 3.18 & 0.956 & 0.585 & 0.920 & 1.53 & \textbf{1.59} & 3.70 & \textbf{0.737} & 0.911 & 0.843 & \underline{10.6} \\
DLinear & 1.57 & 3.43 & 1.23 & 3.29 & 1.03 & 0.604 & 0.966 & 1.55 & 1.69 & 3.41 & 0.821 & 0.997 & 0.894 & 14.6 \\
N-HiTS & 1.20 & 3.43 & 1.16 & nan & \underline{0.917} & \underline{0.571} & 0.919 & 1.51 & \textbf{1.59} & 3.45 & 0.927 & 0.910 & 0.830 & 11.1 \\
PatchTST & 1.26 & 2.95 & \underline{1.15} & \textbf{3.07} & 0.919 & 0.575 & \textbf{0.877} & 1.57 & 1.72 & 3.14 & \underline{0.790} & \underline{0.860} & \textbf{0.810} & \textbf{9.44} \\
TFT & 1.16 & 2.86 & 1.25 & \underline{3.12} & \textbf{0.909} & \textbf{0.556} & \underline{0.896} & 1.69 & 1.73 & \textbf{3.05} & 0.880 & 0.913 & 0.847 & 11.6 \\
AutoTheta & \underline{1.13} & \textbf{2.61} & 1.19 & \underline{3.12} & 1.10 & 1.07 & 0.984 & 1.68 & 1.66 & \underline{3.08} & 1.79 & 0.991 & 0.875 & 10.8 \\
AutoETS & \textbf{1.12} & \underline{2.70} & 1.19 & 3.37 & 1.10 & 1.04 & 0.978 & \underline{1.50} & \textbf{1.59} & 3.14 & 1.68 & 1.08 & 0.953 & 11.8 \\
AutoARIMA & 1.42 & 3.16 & 1.28 & 3.73 & 1.06 & 1.21 & 0.995 & 1.57 & 1.66 & 4.04 & nan & 0.907 & 0.908 & 15.7 \\
Seasonal Naive & 1.43 & 3.17 & 1.60 & 3.97 & 1.40 & 1.29 & 1.06 & 1.63 & 1.70 & 3.55 & 1.08 & 1.00 & 1.00 & 19.0 \\
Naive & 1.46 & 3.17 & 1.48 & 3.97 & 1.40 & 1.29 & 1.06 & 3.59 & 3.63 & 3.55 & 2.05 & 1.00 & 1.19 & 21.4 \\
\bottomrule
\end{tabular}

}
\label{tab:uni-zeroshot-taskspecific-mase}
\end{table}

%% file: tables/uni-zeroshot-taskspecific-wql.tex
\begin{table}[h]
\centering
\caption{\nextcaption}
{\scriptsize
\begin{tabular}{llllllllllllllll}
\toprule
  & \rotatebox{90}{ Australian Electricity } & \rotatebox{90}{ Car Parts } & \rotatebox{90}{ CIF 2016 } & \rotatebox{90}{ Covid Deaths } & \rotatebox{90}{ Dominick } & \rotatebox{90}{ ERCOT Load } & \rotatebox{90}{ ETTh } & \rotatebox{90}{ ETTm } & \rotatebox{90}{ Exchange Rate } & \rotatebox{90}{ FRED-MD } & \rotatebox{90}{ Hospital } & \rotatebox{90}{ M1 (Monthly) } & \rotatebox{90}{ M1 (Quarterly) } & \rotatebox{90}{ M1 (Yearly) } & \rotatebox{90}{ M3 (Monthly) }\\
\midrule
COSMIC (Base) & 0.080 & 1.02 & 0.016 & 0.039 & 0.351 & \underline{0.020} & \textbf{0.070} & 0.067 & 0.012 & 0.070 & 0.058 & \textbf{0.133} & 0.100 & 0.151 & \underline{0.094} \\
N-BEATS & 0.038 & \underline{0.877} & 0.039 & 0.056 & \textbf{0.312} & \underline{0.020} & 0.074 & \underline{0.053} & 0.011 & 0.061 & \textbf{0.050} & 0.187 & 0.085 & 0.182 & 0.101 \\
DeepAR & 0.087 & 0.967 & 0.136 & 0.108 & 0.364 & 0.032 & 0.081 & 0.069 & \underline{0.009} & \underline{0.043} & 0.056 & 0.150 & 0.089 & 0.139 & 0.099 \\
DLinear & 0.066 & 1.12 & 0.033 & 0.077 & 0.435 & 0.023 & 0.076 & 0.071 & \textbf{0.008} & 0.069 & 0.089 & 0.189 & \underline{0.079} & 0.245 & 0.121 \\
N-HiTS & \textbf{0.034} & 0.880 & 0.032 & 0.038 & \underline{0.313} & \underline{0.020} & 0.081 & \textbf{0.051} & 0.010 & 0.057 & \underline{0.052} & 0.189 & 0.111 & 0.198 & 0.097 \\
PatchTST & 0.037 & 0.998 & 0.140 & 0.065 & 0.345 & \textbf{0.017} & \underline{0.071} & 0.054 & 0.010 & \textbf{0.042} & 0.070 & 0.165 & \textbf{0.078} & 0.165 & 0.113 \\
TFT & \underline{0.036} & \textbf{0.871} & \underline{0.011} & \underline{0.034} & 0.320 & 0.023 & 0.082 & 0.075 & 0.011 & 0.112 & 0.053 & 0.175 & 0.122 & \textbf{0.124} & 0.096 \\
AutoTheta & 0.055 & 1.34 & 0.027 & 0.094 & 0.485 & 0.041 & 0.133 & 0.079 & 0.010 & 0.057 & 0.055 & 0.159 & 0.082 & \underline{0.137} & 0.095 \\
AutoETS & 0.125 & 1.31 & 0.039 & 0.064 & 0.483 & 0.122 & 0.132 & 0.095 & 0.010 & 0.055 & 0.053 & 0.162 & 0.083 & 0.142 & \textbf{0.093} \\
AutoARIMA & 0.073 & nan & 0.017 & \textbf{0.029} & nan & 0.052 & 0.105 & 0.073 & 0.011 & 0.056 & 0.058 & \underline{0.146} & 0.091 & 0.160 & 0.102 \\
Seasonal Naive & 0.084 & 1.60 & 0.015 & 0.133 & 0.453 & 0.037 & 0.122 & 0.141 & 0.013 & 0.122 & 0.073 & 0.191 & 0.150 & 0.209 & 0.149 \\
Naive & 0.159 & nan & \textbf{0.009} & 0.133 & 0.453 & 0.181 & 0.202 & 0.121 & 0.015 & 0.064 & 0.087 & 0.258 & 0.130 & 0.209 & 0.158 \\
\bottomrule
\end{tabular}

\vspace{0.3cm}

\begin{tabular}{lllllllllllllll}
\toprule
  & \rotatebox{90}{ M3 (Quarterly) } & \rotatebox{90}{ M3 (Yearly) } & \rotatebox{90}{ M4 (Quarterly) } & \rotatebox{90}{ M4 (Yearly) } & \rotatebox{90}{ M5 } & \rotatebox{90}{ NN5 (Daily) } & \rotatebox{90}{ NN5 (Weekly) } & \rotatebox{90}{ Tourism (Monthly) } & \rotatebox{90}{ Tourism (Quarterly) } & \rotatebox{90}{ Tourism (Yearly) } & \rotatebox{90}{ Traffic } & \rotatebox{90}{ Weather } & \rotatebox{90}{ Agg. Score } & \rotatebox{90}{ Avg. Rank }\\
\midrule
COSMIC (Base) & 0.074 & 0.140 & 0.077 & 0.119 & 0.567 & 0.158 & \underline{0.085} & 0.096 & 0.065 & 0.162 & \underline{0.245} & \textbf{0.128} & \underline{0.658} & 9.30 \\
N-BEATS & 0.080 & 0.181 & \textbf{0.073} & nan & \textbf{0.560} & \underline{0.147} & 0.114 & \textbf{0.084} & \underline{0.063} & 0.154 & 0.270 & 0.144 & 0.681 & 10.6 \\
DeepAR & 0.073 & \textbf{0.122} & 0.080 & 0.111 & 0.657 & 0.155 & 0.087 & 0.092 & 0.072 & \underline{0.127} & \textbf{0.233} & 0.147 & 0.733 & 10.6 \\
DLinear & 0.086 & 0.143 & 0.085 & 0.115 & 0.687 & 0.159 & 0.090 & 0.101 & 0.080 & 0.165 & 0.250 & 0.174 & 0.757 & 15.6 \\
N-HiTS & 0.076 & 0.182 & \textbf{0.073} & nan & \underline{0.563} & 0.149 & 0.098 & 0.092 & 0.077 & 0.139 & 0.263 & \underline{0.143} & 0.672 & 11.1 \\
PatchTST & 0.074 & 0.133 & \underline{0.074} & \textbf{0.106} & 0.597 & 0.149 & \textbf{0.081} & 0.092 & 0.074 & 0.136 & 0.246 & \underline{0.143} & 0.684 & \textbf{8.67} \\
TFT & 0.071 & 0.130 & 0.080 & \underline{0.110} & \textbf{0.560} & \textbf{0.145} & 0.086 & 0.096 & 0.074 & \textbf{0.102} & 0.264 & 0.151 & \textbf{0.639} & \underline{9.00} \\
AutoTheta & \underline{0.070} & 0.128 & 0.079 & 0.115 & 0.636 & 0.294 & 0.090 & 0.091 & \textbf{0.061} & 0.176 & 0.905 & 0.217 & 0.793 & 12.9 \\
AutoETS & \textbf{0.069} & \underline{0.127} & 0.080 & 0.118 & 0.628 & 0.264 & 0.088 & \underline{0.090} & 0.070 & 0.159 & 0.557 & 0.214 & 0.838 & 13.4 \\
AutoARIMA & 0.079 & 0.162 & 0.082 & 0.130 & 0.624 & 0.312 & 0.090 & 0.093 & 0.098 & 0.156 & nan & 0.185 & 0.761 & 15.8 \\
Seasonal Naive & 0.101 & 0.167 & 0.119 & 0.161 & 1.02 & 0.425 & 0.123 & 0.104 & 0.119 & 0.209 & 0.362 & 0.217 & 1.00 & 21.5 \\
Naive & 0.103 & 0.167 & 0.110 & 0.161 & 1.02 & 0.425 & 0.123 & 0.297 & 0.166 & 0.209 & 0.643 & 0.217 & 1.15 & 22.1 \\
\bottomrule
\end{tabular}

}
\label{tab:uni-zeroshot-taskspecific-wql}
\end{table}

%% file: tables/uni-indomain-mase.tex
\begin{table}[h]
\centering
\caption{\nextcaption}
{\scriptsize
\begin{tabular}{llllllllll}
\toprule
  & \rotatebox{90}{ Electricity (15 Min.) } & \rotatebox{90}{ Electricity (Hourly) } & \rotatebox{90}{ Electricity (Weekly) } & \rotatebox{90}{ KDD Cup 2018 } & \rotatebox{90}{ London Smart Meters } & \rotatebox{90}{ M4 (Daily) } & \rotatebox{90}{ M4 (Hourly) } & \rotatebox{90}{ M4 (Monthly) } & \rotatebox{90}{ M4 (Weekly) }\\
\midrule
COSMIC (Mini) & 0.462 & 1.60 & 1.81 & 0.663 & 0.736 & \underline{3.09} & 0.957 & 0.976 & 2.30 \\
COSMIC (Small) & 0.455 & 1.52 & 1.78 & 0.646 & 0.731 & 3.10 & 0.915 & 0.963 & 2.13 \\
COSMIC (Base) & 0.435 & 1.36 & 1.79 & \underline{0.638} & \underline{0.724} & \textbf{3.08} & 0.888 & 0.957 & 2.04 \\
Chronos (Mini) & 0.445 & \underline{1.35} & 1.95 & 0.667 & 0.857 & 3.15 & 0.758 & 0.991 & 2.15 \\
Chronos (Small) & \underline{0.418} & 1.48 & 1.94 & 0.687 & 0.846 & 3.15 & \underline{0.721} & 0.982 & 2.11 \\
Chronos (Base) & \textbf{0.394} & 1.59 & 1.80 & 0.646 & 0.838 & 3.16 & \textbf{0.694} & 0.970 & 2.02 \\
C. Bolt (Small 512) & 0.456 & 1.50 & 1.82 & 0.711 & 0.730 & 3.14 & 0.848 & 0.962 & 2.17 \\
C. Bolt (Base 512) & 0.438 & 1.37 & 1.87 & 0.712 & \textbf{0.722} & 3.14 & 0.836 & \underline{0.956} & 2.04 \\
Moirai-1.0 (Base) & 0.707 & 1.71 & 2.87 & \cellcolor{gray!30}{0.662} & \cellcolor{gray!30}{0.770} & 3.45 & 1.21 & 1.03 & 2.48 \\
Moirai-1.0 (Large) & 0.623 & 1.67 & 2.76 & \cellcolor{gray!30}{0.656} & \cellcolor{gray!30}{0.754} & 3.38 & 0.950 & 1.01 & 2.45 \\
Moirai-1.1 (Large) & 0.575 & 1.70 & 2.66 & \cellcolor{gray!30}{0.661} & \cellcolor{gray!30}{0.745} & 4.31 & 0.911 & 0.986 & 2.52 \\
Moirai-1.1 (Base) & 0.629 & 1.71 & 2.73 & \cellcolor{gray!30}{0.653} & \cellcolor{gray!30}{0.758} & 4.84 & 0.999 & 0.986 & 2.53 \\
TimesFM & 0.750 & \textbf{1.20} & \underline{1.77} & 0.687 & 0.822 & \cellcolor{gray!30}{3.27} & \cellcolor{gray!30}{0.767} & \cellcolor{gray!30}{0.886} & 2.26 \\
TTM & 0.677 & 1.74 & 3.88 & \cellcolor{gray!30}{0.675} & \cellcolor{gray!30}{0.886} & 4.40 & 2.78 & 1.53 & 3.48 \\
DeepAR & 0.515 & 1.53 & 2.52 & 0.779 & 0.832 & 3.31 & 1.21 & 1.04 & 2.35 \\
DLinear & 0.452 & 1.37 & 2.61 & 0.695 & 0.799 & 3.46 & 1.87 & 1.02 & 2.43 \\
TFT & 1.11 & 1.79 & 2.80 & 1.02 & 0.788 & 3.29 & 1.83 & 1.01 & 2.75 \\
PatchTST & 0.450 & \underline{1.35} & \textbf{1.63} & \textbf{0.616} & 0.733 & 3.45 & 0.967 & 0.962 & \underline{2.00} \\
N-BEATS & 0.567 & 1.85 & 2.04 & 0.731 & 0.781 & 3.16 & 3.46 & \textbf{0.942} & \textbf{1.98} \\
N-HiTS & 0.579 & 1.88 & 1.98 & 0.674 & 0.777 & 3.14 & 3.23 & 0.994 & 2.09 \\
AutoARIMA & nan & 1.71 & 3.01 & 1.02 & nan & 3.26 & nan & nan & 2.37 \\
AutoTheta & 0.583 & 2.15 & 3.08 & 1.14 & 0.966 & 3.33 & 2.46 & 0.966 & 2.66 \\
AutoETS & nan & 1.77 & 3.09 & 1.01 & nan & 3.27 & 1.60 & 0.970 & 2.55 \\
Seasonal Naive & 0.498 & 1.84 & 3.04 & 0.994 & 0.966 & 3.28 & 1.19 & 1.26 & 2.78 \\
Naive & 1.27 & 4.16 & 3.04 & nan & 1.30 & 3.28 & 11.6 & 1.21 & 2.78 \\
\bottomrule
\end{tabular}

\vspace{0.3cm}

\begin{tabular}{lllllllll}
\toprule
  & \rotatebox{90}{ Pedestrian Counts } & \rotatebox{90}{ Rideshare } & \rotatebox{90}{ Taxi (30 Min.) } & \rotatebox{90}{ Temperature-Rain } & \rotatebox{90}{ Uber TLC (Daily) } & \rotatebox{90}{ Uber TLC (Hourly) } & \rotatebox{90}{ Agg. Score } & \rotatebox{90}{ Avg. Rank }\\
\midrule
COSMIC (Mini) & 0.312 & 0.852 & 0.959 & 0.926 & 0.840 & 0.671 & 0.737 & 7.93 \\
COSMIC (Small) & 0.298 & 0.862 & 0.970 & 0.918 & 0.835 & \underline{0.667} & 0.723 & 5.73 \\
COSMIC (Base) & \underline{0.293} & 1.59 & 0.922 & 0.916 & 0.840 & 0.669 & 0.737 & \underline{5.07} \\
Chronos (Mini) & 0.303 & 0.830 & 0.944 & 1.03 & 0.906 & 0.689 & 0.732 & 9.13 \\
Chronos (Small) & 0.304 & 0.854 & 0.941 & 1.01 & 0.870 & 0.677 & 0.727 & 8.47 \\
Chronos (Base) & \textbf{0.286} & 0.862 & \textbf{0.849} & 0.986 & 0.839 & 0.673 & \textbf{0.706} & 6.07 \\
C. Bolt (Small 512) & 0.295 & 0.863 & 0.915 & \underline{0.904} & 0.835 & 0.670 & 0.722 & 6.47 \\
C. Bolt (Base 512) & \textbf{0.286} & 0.867 & \underline{0.877} & \textbf{0.902} & 0.818 & \textbf{0.665} & \underline{0.707} & \textbf{4.73} \\
Moirai-1.0 (Base) & 0.354 & 0.910 & 1.37 & 0.963 & 0.940 & 0.730 & 0.857 & 15.9 \\
Moirai-1.0 (Large) & 0.330 & 0.900 & 1.09 & 0.988 & 0.871 & 0.716 & 0.806 & 13.1 \\
Moirai-1.1 (Large) & 0.357 & \textbf{0.814} & 1.28 & 1.08 & 0.868 & 0.723 & 0.825 & 13.3 \\
Moirai-1.1 (Base) & 0.332 & 0.896 & 1.27 & 0.992 & 0.874 & 0.703 & 0.839 & 14.3 \\
TimesFM & 0.307 & 0.853 & 1.05 & 1.01 & \textbf{0.803} & 0.677 & 0.745 & 8.40 \\
TTM & 0.363 & 0.932 & 1.02 & 1.42 & 1.22 & 0.812 & 1.03 & 19.7 \\
DeepAR & 0.311 & 0.996 & 1.16 & 1.01 & 0.905 & 0.703 & 0.821 & 14.8 \\
DLinear & 0.327 & 1.45 & 1.02 & 1.37 & 0.855 & 0.778 & 0.864 & 15.0 \\
TFT & 0.364 & 1.07 & 1.11 & 0.994 & 0.916 & 0.746 & 0.939 & 18.1 \\
PatchTST & 0.339 & \underline{0.827} & 1.08 & 1.25 & \underline{0.813} & 0.696 & 0.740 & 7.80 \\
N-BEATS & 0.315 & 0.919 & 0.934 & 1.34 & 0.879 & 0.751 & 0.861 & 12.9 \\
N-HiTS & 0.324 & 0.933 & 0.950 & 1.23 & 0.877 & 0.716 & 0.854 & 13.5 \\
AutoARIMA & 0.383 & 1.03 & nan & 1.52 & 1.11 & 0.982 & 0.941 & 20.9 \\
AutoTheta & 1.28 & 0.970 & 1.19 & 1.95 & 1.31 & 1.04 & 1.13 & 20.4 \\
AutoETS & 0.487 & 0.910 & nan & 1.97 & 1.23 & 1.01 & 0.983 & 20.1 \\
Seasonal Naive & 0.369 & 1.25 & 1.16 & 2.24 & 1.38 & 0.931 & 1.00 & 20.1 \\
Naive & 0.842 & nan & 1.77 & nan & 1.38 & 1.39 & 1.48 & 23.2 \\
\bottomrule
\end{tabular}

}
\label{tab:uni-indomain-mase}
\end{table}

%% file: tables/uni-indomain-wql.tex
\begin{table}[h]
\centering
\caption{\nextcaption}
{\scriptsize
\begin{tabular}{llllllllll}
\toprule
  & \rotatebox{90}{ Electricity (15 Min.) } & \rotatebox{90}{ Electricity (Hourly) } & \rotatebox{90}{ Electricity (Weekly) } & \rotatebox{90}{ KDD Cup 2018 } & \rotatebox{90}{ London Smart Meters } & \rotatebox{90}{ M4 (Daily) } & \rotatebox{90}{ M4 (Hourly) } & \rotatebox{90}{ M4 (Monthly) } & \rotatebox{90}{ M4 (Weekly) }\\
\midrule
COSMIC (Mini) & 0.080 & 0.113 & \textbf{0.060} & 0.273 & 0.340 & \textbf{0.020} & 0.025 & 0.097 & 0.040 \\
COSMIC (Small) & \underline{0.077} & 0.095 & 0.065 & \underline{0.265} & 0.337 & \textbf{0.020} & 0.025 & 0.096 & \underline{0.039} \\
COSMIC (Base) & \textbf{0.076} & 0.096 & 0.064 & 0.268 & \underline{0.334} & \textbf{0.020} & 0.026 & 0.095 & \textbf{0.037} \\
Chronos (Mini) & 0.082 & \underline{0.089} & 0.067 & 0.271 & 0.436 & 0.022 & 0.025 & 0.103 & 0.041 \\
Chronos (Small) & 0.080 & 0.105 & 0.073 & 0.289 & 0.431 & 0.022 & 0.024 & 0.103 & 0.040 \\
Chronos (Base) & 0.078 & 0.114 & \underline{0.062} & 0.268 & 0.428 & 0.022 & 0.024 & 0.103 & \textbf{0.037} \\
C. Bolt (Small 512) & 0.081 & 0.100 & \underline{0.062} & 0.295 & 0.337 & \underline{0.021} & 0.028 & 0.095 & \underline{0.039} \\
C. Bolt (Base 512) & \underline{0.077} & 0.093 & 0.065 & 0.293 & \textbf{0.333} & \underline{0.021} & 0.024 & 0.095 & \textbf{0.037} \\
Moirai-1.0 (Base) & 0.104 & 0.121 & 0.117 & \cellcolor{gray!30}{0.288} & \cellcolor{gray!30}{0.358} & 0.024 & 0.025 & 0.102 & 0.050 \\
Moirai-1.0 (Large) & 0.105 & 0.117 & 0.166 & \cellcolor{gray!30}{0.278} & \cellcolor{gray!30}{0.350} & 0.023 & \underline{0.022} & 0.100 & 0.047 \\
Moirai-1.1 (Large) & 0.095 & 0.121 & 0.097 & \cellcolor{gray!30}{0.275} & \cellcolor{gray!30}{0.346} & 0.033 & \textbf{0.021} & 0.098 & 0.049 \\
Moirai-1.1 (Base) & 0.093 & 0.118 & 0.090 & \cellcolor{gray!30}{0.280} & \cellcolor{gray!30}{0.352} & 0.036 & 0.023 & 0.098 & 0.048 \\
TimesFM & 0.121 & \textbf{0.079} & \underline{0.062} & 0.288 & 0.384 & \cellcolor{gray!30}{0.021} & \cellcolor{gray!30}{0.021} & \cellcolor{gray!30}{0.087} & 0.040 \\
N-HiTS & 0.081 & 0.128 & 0.098 & 0.302 & 0.358 & 0.022 & 0.040 & \underline{0.094} & \underline{0.039} \\
DeepAR & 0.090 & 0.106 & 0.116 & 0.330 & 0.405 & 0.023 & 0.038 & 0.101 & 0.046 \\
DLinear & 0.079 & 0.095 & 0.146 & 0.312 & 0.369 & 0.024 & 0.038 & 0.111 & 0.044 \\
TFT & 0.189 & 0.125 & 0.106 & 0.571 & 0.365 & 0.023 & 0.033 & 0.097 & 0.051 \\
PatchTST & 0.082 & \underline{0.089} & 0.069 & \textbf{0.252} & 0.346 & 0.023 & 0.027 & 0.095 & \underline{0.039} \\
N-BEATS & 0.084 & 0.127 & 0.097 & 0.315 & 0.357 & 0.022 & 0.045 & \textbf{0.093} & 0.040 \\
AutoETS & nan & 0.129 & 0.151 & 2.27 & nan & 0.027 & 0.066 & 0.100 & 0.052 \\
AutoARIMA & nan & 0.126 & 0.138 & 0.528 & nan & 0.023 & nan & nan & 0.050 \\
AutoTheta & 0.229 & 0.198 & 0.146 & 0.521 & 0.660 & 0.024 & 0.041 & 0.098 & 0.053 \\
Seasonal Naive & 0.117 & 0.147 & 0.198 & 0.556 & 0.541 & 0.028 & 0.048 & 0.146 & 0.063 \\
Naive & 0.279 & 0.363 & 0.198 & nan & 0.731 & 0.028 & 0.166 & 0.140 & 0.063 \\
\bottomrule
\end{tabular}

\vspace{0.3cm}

\begin{tabular}{lllllllll}
\toprule
  & \rotatebox{90}{ Pedestrian Counts } & \rotatebox{90}{ Rideshare } & \rotatebox{90}{ Taxi (30 Min.) } & \rotatebox{90}{ Temperature-Rain } & \rotatebox{90}{ Uber TLC (Daily) } & \rotatebox{90}{ Uber TLC (Hourly) } & \rotatebox{90}{ Agg. Score } & \rotatebox{90}{ Avg. Rank }\\
\midrule
COSMIC (Mini) & 0.241 & 0.134 & 0.306 & 0.615 & 0.098 & 0.154 & 0.577 & 6.87 \\
COSMIC (Small) & 0.221 & 0.131 & 0.310 & 0.612 & 0.098 & \underline{0.151} & \underline{0.564} & \underline{4.80} \\
COSMIC (Base) & 0.214 & 20.7 & 0.294 & 0.608 & 0.101 & \underline{0.151} & 0.784 & 5.53 \\
Chronos (Mini) & 0.236 & 0.133 & 0.313 & 0.704 & 0.105 & 0.161 & 0.598 & 9.93 \\
Chronos (Small) & 0.237 & 0.140 & 0.312 & 0.685 & 0.100 & 0.155 & 0.603 & 10.3 \\
Chronos (Base) & \textbf{0.204} & 0.137 & \textbf{0.274} & 0.669 & 0.097 & 0.153 & 0.580 & 7.27 \\
C. Bolt (Small 512) & 0.224 & 0.133 & 0.286 & \underline{0.604} & \underline{0.093} & 0.153 & 0.572 & 6.20 \\
C. Bolt (Base 512) & \underline{0.209} & 0.133 & \underline{0.275} & \textbf{0.602} & \underline{0.093} & \textbf{0.150} & \textbf{0.554} & \textbf{3.60} \\
Moirai-1.0 (Base) & 0.272 & 0.164 & 0.512 & 0.655 & 0.114 & 0.177 & 0.691 & 15.6 \\
Moirai-1.0 (Large) & 0.259 & 0.158 & 0.368 & 0.685 & 0.107 & 0.165 & 0.670 & 13.3 \\
Moirai-1.1 (Large) & 0.276 & \textbf{0.123} & 0.447 & 0.814 & 0.102 & 0.165 & 0.662 & 12.7 \\
Moirai-1.1 (Base) & 0.259 & 0.153 & 0.463 & 0.677 & 0.102 & 0.164 & 0.666 & 12.9 \\
TimesFM & 0.233 & 0.133 & 0.334 & 0.646 & \textbf{0.089} & 0.153 & 0.579 & 7.07 \\
N-HiTS & 0.254 & 0.152 & 0.306 & 0.780 & 0.116 & 0.166 & 0.656 & 12.3 \\
DeepAR & 0.229 & \underline{0.130} & 0.395 & 0.718 & 0.110 & 0.176 & 0.676 & 13.6 \\
DLinear & 0.247 & 0.159 & 0.335 & 0.848 & 0.106 & 0.234 & 0.697 & 14.9 \\
TFT & 0.261 & 0.134 & 0.382 & 0.670 & 0.111 & 0.179 & 0.734 & 15.3 \\
PatchTST & 0.257 & 0.135 & 0.363 & 0.804 & 0.100 & 0.167 & 0.601 & 9.60 \\
N-BEATS & 0.241 & 0.172 & 0.305 & 0.798 & 0.108 & 0.161 & 0.664 & 12.6 \\
AutoETS & 0.619 & 0.154 & nan & 1.18 & 0.167 & 0.462 & 1.08 & 21.2 \\
AutoARIMA & 0.340 & 0.157 & nan & 0.869 & 0.151 & 0.311 & 0.876 & 20.5 \\
AutoTheta & 1.82 & 0.138 & 0.456 & 1.06 & 0.190 & 0.433 & 1.08 & 19.5 \\
Seasonal Naive & 0.319 & 0.186 & 0.471 & 1.42 & 0.231 & 0.299 & 1.00 & 21.5 \\
Naive & 0.814 & nan & 0.741 & nan & 0.231 & 0.625 & 1.43 & 23.1 \\
\bottomrule
\end{tabular}

}
\label{tab:uni-indomain-wql}
\end{table}